\documentclass[twoside,11pt]{article}

\usepackage{subfigure}
\usepackage{paralist,amsmath, amssymb, bm,color}
\usepackage{algorithm, algorithmic}

\usepackage[preprint]{jmlr2e}

\usepackage{booktabs}
\usepackage{makecell}
\usepackage{comment}
\usepackage{caption}
\DeclareMathOperator*{\sgn}{sign}

\makeatletter 
\newcommand\figcaption{\def\@captype{figure}\caption} 
\newcommand\tabcaption{\def\@captype{table}\caption} 
\makeatother

\usepackage{jmlr2e}
\usepackage{hyperref}
\usepackage{cleveref}
\newtheorem{prop}{Proposition}
\newtheorem{myremark}{Remark}

\def \E {\mathrm{E}}
\def \x {\mathbf{x}}
\def \v {\mathbf{v}}
\def \g {\mathbf{g}}
\def \L {\mathcal{L}}

\def \O {\widetilde{O}}

\def \z {\mathbf{z}}
\def \Z {\mathcal{Z}}

\def \u {\mathbf{u}}

\def \w {\mathbf{w}}

\def \R {\mathbb{R}}

\def \P {\mathcal{P}}

\def \W {\mathcal{W}}
\def \N {\mathcal{N}}

\def \q {\mathbf{q}}

\def \M {\mathcal{M}}

\def \p {\mathbf{p}}
\def \q {\mathbf{q}}

\def \I {\mathcal{I}}

\def \B {\mathbf{B}}
\def \SS {\mathcal{S}}

\def \zh {\hat{\z}}

\def \s {\mathbf{s}}
\def \gh {\hat{\g}}
\def \gt {\tilde{\g}}

\def \C {\mathcal{C}}

\def \wt {\widetilde{\w}}

\def \nb {\bar{n}}

\def \wb {\bar{\w}}
\def \qb {\bar{\q}}

\def \wh {\widehat{\w}}
\def \qh {\widehat{\q}}

\def \B {\mathcal{B}}

\def \zn {\mathbb{Z}_+}
\def \Lt  {\widetilde{L}}

\def \Dw {D}

\def \ind {\mathbb{I}}

\newcommand\inner[2]{\langle #1, #2 \rangle}
\def \hats {\hat{s}}
\def \hatsb {\hat{\s}}
\def \tildes {\tilde{s}}
\def \tildesb {\tilde{\s}}
\newcommand \indicator[1]{\mathbb{I}[#1]}

\DeclareMathOperator*{\argmin}{argmin}
\DeclareMathOperator*{\argmax}{argmax}

\newtheorem{ass}{Assumption}

\usepackage{lastpage}
\jmlrheading{23}{2024}{1-\pageref{LastPage}}{10/24; Revised 5/24}{9/24}{21-0000}{Lijun Zhang, Haomin Bai, Peng Zhao, Tianbao Yang and Zhi-Hua Zhou}

\ShortHeadings{GROUP DISTRIBUTIONALLY ROBUST OPTIMIZATION AND BEYOND}{Zhang, Bai, Zhao, Yang, and Zhou}
\firstpageno{1}

\begin{document}

\title{Stochastic Approximation Approaches to \\ Group Distributionally Robust Optimization and Beyond}

\author{\name Lijun Zhang \email zhanglj@lamda.nju.edu.cn
    \AND
    \name Haomin Bai \email baihm@lamda.nju.edu.cn 
    \AND
    \name Peng Zhao \email zhaop@lamda.nju.edu.cn \\
    \addr National Key Laboratory for Novel Software Technology, Nanjing University, China\\
    School of Artificial Intelligence, Nanjing University, China
    \AND
    \name Tianbao Yang \email tianbao-yang@tamu.edu\\
	\addr Department of Computer Science and Engineering, Texas A$\&$M University, College Station, USA
	\AND
	\name Zhi-Hua Zhou \email zhouzh@lamda.nju.edu.cn\\
	\addr National Key Laboratory for Novel Software Technology, Nanjing University, China\\
	School of Artificial Intelligence, Nanjing University, China}

\editor{My editor}

\maketitle

\begin{abstract}%   <- trailing '%' for backward compatibility of .sty file
This paper investigates group distributionally robust optimization (GDRO) with the goal of learning a model that performs well over $m$ different distributions. 
First, we formulate GDRO as a stochastic convex-concave saddle-point problem, which is then solved by stochastic mirror descent (SMD) with $m$ samples in each iteration, and attain a nearly optimal sample complexity. To reduce the number of samples required in each round from $m$ to 1, we cast GDRO as a two-player game, where one player conducts SMD and the other executes an online algorithm for non-oblivious multi-armed bandits, maintaining the same sample complexity. Next, we extend GDRO to address scenarios involving  imbalanced data and heterogeneous distributions. In the first scenario, we introduce a weighted variant of GDRO, enabling \emph{distribution-dependent} convergence rates that rely on the number of samples from each distribution. We design two strategies to meet the sample budget: one integrates non-uniform sampling into SMD, and the other employs the stochastic mirror-prox algorithm with mini-batches, both of which deliver faster rates for distributions with more samples. 
In the second scenario, we propose to optimize the average top-$k$ risk instead of the maximum risk, thereby mitigating the impact of outlier distributions. Similar to the case of vanilla GDRO, we develop two stochastic approaches: one uses $m$ samples per iteration via SMD, and the other consumes $k$ samples per iteration through  an online algorithm for non-oblivious combinatorial semi-bandits.
\end{abstract}


\section{Introduction}
In the classical statistical machine learning, our goal is to minimize the risk with respect to a \emph{fixed} distribution $\P_0$ \citep{Nature:Statistical}, i.e.,
\begin{equation} \label{eqn:supervised}
\min_{\w \in \W} \  \left\{ R_0(\w)=\E_{\z \sim \P_0}\big[\ell(\w;\z)\big] \right\},
\end{equation}
where $\z \in \Z$ is a sample drawn from $\P_0$, $\W$ denotes a hypothesis class, and  $\ell(\w;\z)$ is a loss measuring the prediction error of model $\w$ on $\z$. During the past decades, various algorithms have been developed to optimize (\ref{eqn:supervised}), and can be grouped in two categories: sample average approximation (SAA) and  stochastic approximation (SA) \citep{SA:Springer}. In SAA, we minimize an empirical risk defined as the average loss over a set of samples drawn from $\P_0$, and in SA, we directly solve the original problem by using stochastic observations of the objective $R_0(\cdot)$.

However, a model trained on a single distribution may lack robustness in the sense that (i) it could suffer high error on minority subpopulations, though the average loss is small; (ii) its performance could degenerate dramatically when tested on a different distribution.
Distributionally robust optimization (DRO) provides a principled way to address those limitations by minimizing the worst-case risk in a neighborhood of $\P_0$ \citep{DRO:MS:2013}. Recently, it has attracted great interest in optimization \citep{DRSP:SIAM:2017}, statistics \citep{10.1214/20-AOS2004}, operations research \citep{Statistics:RO}, and machine learning \citep{pmlr-v80-hu18a,ADA-CVAR,NEURIPS2021_164bf317,Regret:RML:DS}.  In this paper, we consider an emerging class of DRO problems, named as Group DRO (GDRO) which optimizes the maximum risk 
\begin{equation} \label{eqn:Max_risk}
\L_{\max}(\w)=\max_{i\in[m]}\left\{R_i(\w)=\E_{\z \sim \P_i}\big[\ell(\w;\z)\big]\right\}
\end{equation}
over a finite number of distributions \citep{oren-etal-2019-distributionally,Gouop_DRO}. Mathematically, GDRO can be formulated as a minimax stochastic problem:
\begin{equation} \label{eqn:group:dro}
\min_{\w \in \W}   \max_{i\in[m]}  \  \left\{ R_i(\w)\right\}
\end{equation}
where $\P_1,\ldots,\P_m$ denote $m$ distributions. A motivating example is federated learning, where a centralized model is deployed at multiple clients, each of which faces a (possibly) different data distribution \citep{pmlr-v97-mohri19a}.

Supposing that samples can be drawn from all distributions freely, we develop efficient SA approaches for (\ref{eqn:group:dro}), in favor of their light computations over SAA methods. As elaborated by \citet[\S~3.2]{nemirovski-2008-robust}, we can cast (\ref{eqn:group:dro}) as a stochastic convex-concave saddle-point problem:
\begin{equation} \label{eqn:convex:concave}
\min_{\w \in \W} \max_{\q \in \Delta_m}  \   \left\{\phi(\w,\q)= \sum_{i=1}^m q_i R_i(\w) \right\}
\end{equation}
where $\Delta_m=\{\q \in \R^m| \q \geq\mathbf{0}, \sum_{i=1}^m q_i=1\}$ is the ($m{-}1$)-dimensional simplex, and then solve (\ref{eqn:convex:concave}) by their mirror descent stochastic approximation method, namely stochastic mirror descent (SMD). In fact, several recent studies have adopted this (or similar) strategy to optimize (\ref{eqn:convex:concave}). But, unfortunately, we found that existing results are unsatisfactory because they either deliver a loose sample complexity \citep{Gouop_DRO}, suffer subtle dependency issues in their analysis~\citep{Online:Multiple:Distribution,DRO:Online:Game}, or hold only in expectation \citep{NEURIPS2022_e90b00ad}.

As a starting point, we first provide a routine application of SMD to (\ref{eqn:convex:concave}), and discuss the theoretical guarantee. In each iteration, we draw $1$ sample from every distribution to construct unbiased estimators of $R_i(\cdot)$ and its gradient, and then update both $\w$ and $\q$ by SMD. The proposed method achieves an $O(\sqrt{(\log m)/T})$ convergence rate in expectation and with high probability, where $T$ is the total number of iterations. As a result, we obtain an $O(m(\log m)/\epsilon^2)$ sample complexity for finding an $\epsilon$-optimal solution of (\ref{eqn:convex:concave}), which matches the $\Omega(m/\epsilon^2)$ lower bound \citep[Theorem~5]{DRO:Online:Game} up to a logarithmic factor, and tighter than the $O(m^2  (\log m)/\epsilon^2)$ bound of \citet{Gouop_DRO} by an $m$ factor. While being straightforward, this result seems \emph{new} for GDRO.  Additionally, we note that the aforementioned method requires setting the number of iterations $T$ in advance, which could be inconvenient in practice. To avoid this limitation, we further propose an \emph{anytime} algorithm by using time-varying step sizes, and obtain an $\O(\sqrt{(\log m)/t})$ \footnote{We use the $\O$ notation to hide constant factors as well as polylogarithmic factors in $t$.}  convergence rate at each iteration $t$.

Then, we proceed to reduce the number of samples used in each iteration from $m$ to $1$. We remark that a naive uniform sampling over $m$ distributions does not work well, and yields a higher sample complexity \citep{Gouop_DRO}. As an alternative, we borrow techniques from online learning with stochastic observations, and explicitly tackle the \emph{non-oblivious} nature of the online process, which distinguishes our method from that of \citet{DRO:Online:Game}. Specifically, we use SMD to update $\w$, and Exp3-IX, a powerful algorithm for non-oblivious multi-armed bandits (MAB) \citep{NIPS2015_e5a4d6bf}, with stochastic rewards to update $\q$. In this way, our algorithm only needs $1$ sample in each round and attains an $O(\sqrt{m (\log m)/T})$ convergence rate, implying the same $O(m (\log m)/\epsilon^2)$ sample complexity. 
Similarly, we also put forward an anytime variant, achieving an $\O(\sqrt{m(\log m)/t})$ convergence rate.

Subsequently, we extend GDRO to address two specific scenarios, as illustrated below.

\subsection{Extension to Imbalanced Data}
In the first extention, we investigate a more practical and challenging scenario in which there are different budgets of samples that can be drawn from each distribution, a natural phenomenon encountered in learning with imbalanced data \citep{pmlr-v48-amodei16}. % Indeed, in real-world applications, the cost of collecting data from different distributions can vary significantly, leading to different sample budgets. As an example, in multilingual multimodal learning, the size of training data for different languages may vary greatly \citep{amini2009learning}.
Let $n_i$ be the sample budget of the $i$-th distribution, and without loss of generality, we assume that $n_1 \geq n_2 \geq \cdots \geq n_m$. 
Now, the goal is not to attain the optimal sample complexity, but to reduce the risk on all distributions as much as possible, under the budget constraint. To achieve this goal, we propose a novel formulation of weighted GDRO, which weights each risk $R_i(\cdot)$ in \eqref{eqn:convex:concave} by a scale factor $p_i$. For GDRO with different budgets, we develop two SA approaches based on non-uniform sampling and mini-batches, respectively. 
%In this setting, the anytime capability is unnecessary because the budgets are known and fixed.

In each iteration of the first approach, we draw $1$ sample from every $\P_i$ with probability $n_i/n_1$, and then construct stochastic gradients to perform mirror descent. Consequently, the budget will be satisfied in expectation after $n_1$ rounds, and our algorithm can be regarded as SMD for an instance of weighted GDRO. With the help of scale factors, we demonstrate  the proposed algorithm enjoys \emph{distribution-dependent} convergence in the sense that it converges faster for distributions with more samples. In particular, the excess risk on distribution $\P_i$ reduces at an $O(\sqrt{n_1 \log m}/n_i)$ rate, and for $\P_1$, it becomes $O(\sqrt{(\log m)/n_1})$, which almost matches the optimal $O(\sqrt{1/n_1})$ rate of learning from a single distribution with $n_1$ samples.

On the other hand, for distribution $\P_i$ with budget $n_i < n_1$, the above $O(\sqrt{n_1 \log m}/n_i)$ rate is worse than the $O(\sqrt{1/n_i})$ rate obtained by learning from $\P_i$ alone. In shape contrast with this limitation, our second approach yields nearly optimal convergence rates for \emph{multiple} distributions across a large range of budgets. To meet the budget constraint, it runs for $\nb \leq n_m$ rounds, and in each iteration, draws a mini-batch of $n_i/\nb$ samples from every distribution $\P_i$. 
As a result, (i) the budget constraint is satisfied \emph{exactly}; (ii) for distributions with a larger budget, the associated risk function can be estimated more accurately, making the variance of the stochastic gradient smaller. To benefit from the small variance, we leverage stochastic mirror-prox algorithm \citep{Nemirovski:SMP}, instead of SMD, to update solutions, and again make use of the weighted GDRO formulation to obtain distribution-wise convergence rates. Theoretical analysis shows that the excess risk converges at an $O((\frac{1}{n_m} + \frac{1}{\sqrt{n_i}}) \log m)$ rate for each $\P_i$. Thus,  we obtain a nearly optimal $O((\log m)/\sqrt{n_i})$ rate for distributions $\P_i$ with $n_i \leq n_m^2$, and an $O((\log m)/n_m)$ rate otherwise. Note that the latter rate is as expected since the algorithm only updates  $O(n_m)$ times.

\subsection{Extension to Heterogeneous Distributions} 
In the second extension, we delve into another scenario where distributions exhibit heterogeneity, indicating significant variations in their risks \citep{federatedlearning2019}.  The widely acknowledged sensitivity of the \emph{max} operation to outliers implies that GDRO could be dominated by  a single outlier distribution, while neglecting others \citep{pmlr-v48-shalev-shwartzb16}.  Inspired by the average top-$k$ loss for supervised learning \citep{ATk_SVM_loss},  we modify our objective from the maximum risk $\L_{\max}(\w)$ in GDRO to  the average top-$k$ risk:
\begin{equation} \label{eqn:ATkR}
\L_{k}(\w)=\max_{\I\in \B_{m,k}}\left\{\frac{1}{k}\sum_{i\in \I} R_i(\w)\right\}
\end{equation}
where $\B_{m,k}$ is the set of subsets of $[m]$ with size $k$, i.e., $\B_{m,k}=\{\I\subseteq [m]||\I|=k\}$. This modification aims to reduce the impact of outliers in heterogeneous distributions while still including GDRO as a special case.

We refer to the minimization of $\L_{k}(\w)$ as average top-$k$ risk optimization (AT$_k$RO), and develop two stochastic algorithms. Similar to GDRO, AT$_k$RO can be formulated as a stochastic convex-concave saddle-point problem, akin to (\ref{eqn:convex:concave}), with the only difference being that the domain of $\q$ is the capped simplex instead of the standard simplex. Therefore, we can employ SMD to update $\w$ and $\q$, which uses $m$ samples in each round.  Theoretical analysis demonstrates that  this approach achieves an $O(\sqrt{(\log (m/k))/T})$ convergence rate, implying an $O((m \log (m/k))/\epsilon^2)$ sample complexity. Furthermore, to circumvent the limitation of predefining the total number of iterations $T$, we  introduce an anytime version that attains an $\O(\sqrt{(\log (m/k))/t})$ convergence rate.

Following the second approach for GDRO, we reduce the number of samples required in each round from $m$ to $k$ by casting AT$_k$RO as a two-player game. In each round, we use
the Dependent Rounding  (DepRound) algorithm \citep{DepRound} to select $k$ distributions based on the current value of $\q$, and then  draw $1$ sample from each selected distribution. Then, we construct unbiased stochastic gradients for $\w$, and apply SMD for updates.  Since the domain of $\q$ is the capped simplex, we model the online problem for $\q$ as an instance of non-oblivious combinatorial semi-bandits, and extend Exp3-IX to develop its update rule. We prove that our algorithm achieves an $O(\sqrt{m(\log m)/(kT)})$ convergence rate, yielding an $O(m (\log m)/\epsilon^2)$ sample complexity. Similarly, we have also designed an anytime approach, which uses $1$ sample per round and achieves an $\O(\sqrt{m(\log m)/t})$ rate.

This paper extends our previous conference version \citep{SA:GDRO:NeurIPS} by developing anytime algorithms, investigating a new scenario, and conducting more experiments, as detailed below.
\begin{compactitem}
\item First, we adapt the two SA algorithms for GDRO to operate in an anytime manner. In the conference paper, our algorithms for GDRO required predefining the total number of iterations $T$ to set step sizes. By adopting time-varying step sizes, we design anytime algorithms and provide the corresponding theoretical analysis.
    
\item Second, we explore the scenario of heterogeneous distributions, which involves outlier distributions with significantly high risks. To mitigate the impact of these outliers, we propose to solve the AT$_k$RO problem and develop two algorithms: one employs SMD with $m$ samples per round, achieving a sample complexity of $O((m \log (m/k))/\epsilon^2)$; the other combines SMD with an algorithm for non-oblivious combinatorial semi-bandits, achieving a sample complexity of $O(m (\log m)/\epsilon^2)$ and using $k$ samples in each iteration. Furthermore, we have also extended these two algorithms into anytime versions.
    
\item Last, we construct a heterogeneous data set and perform experiments to verify the advantages of AT$_k$RO. Additionally, we compare the performance of the anytime algorithms with their non-anytime counterparts, demonstrating the benefits of the anytime capability.
\end{compactitem}

\section{Related Work} \label{sec:related-work}
Distributionally robust optimization (DRO) stems from the pioneering work of \citet{DRO:Scarf:1958}, and has gained a lot of interest with the advancement of robust optimization \citep{Robust:OPT:Book,doi:10.1287/opre.2015.1374}. It has been successfully applied to a variety of machine learning tasks, including adversarial training \citep{Sinha:ICLR:2018},  algorithmic fairness \citep{pmlr-v80-hashimoto18a}, class imbalance \citep{pmlr-v119-xu20b}, long-tail learning  \citep{9710454}, label shift \citep{Label:SHIFT:ICLR:2021}, etc.

In general, DRO is formulated to reflect our uncertainty about the target distribution. To ensure good performance under distribution perturbations,  it minimizes the risk w.r.t.~the worst distribution in an uncertainty set, i.e.,
\begin{equation} \label{eqn:dro}
\min_{\w \in \W}  \   \sup_{\P \in \SS(\P_0)} \left\{\E_{\z \sim \P}\big[\ell(\w;\z)\big]\right\}
\end{equation}
where $\SS(\P_0)$ denotes a set of probability distributions around $\P_0$.  In the literature, there mainly exist three ways to construct $\SS(\P_0)$: (i) enforcing moment constraints \citep{10.2307/40792682}, (ii) defining a neighborhood around $\P_0$ by a distance function such as the $f$-divergence \citep{DRO:MS:2013},  the Wasserstein distance \citep{doi:10.1287/educ.2019.0198}, and the Sinkhorn distance \citep{Sinkhorn:DRO}, and (iii) hypothesis testing of goodness-of-fit  \citep{RSAA:MP:2018}.

By drawing a set of samples from $\P_0$, we can also define an empirical DRO problem, which can be regarded as an SAA approach for solving (\ref{eqn:dro}). When the uncertainty set is defined in terms of the Cressie--Read family of $f$-divergences, \citet{10.1214/20-AOS2004} have studied finite sample and asymptotic properties of the empirical solution. Besides, it has been proved that empirical DRO can also benefit the risk minimization problem in (\ref{eqn:supervised}).  \citet{NIPS2017_5a142a55} show that empirical DRO with the $\chi^2$-divergence has the effect of variance regularization, leading to better generalization w.r.t.~distribution $\P_0$. Later, \citet{Statistics:RO} demonstrate similar behaviors for the $f$-divergence constrained neighborhood, and provide one- and two-sided confidence intervals for the minimum risk in (\ref{eqn:supervised}). Based on the Wasserstein distance, \citet{DRO:MP:2018} establish an upper confidence bound on the risk of the empirical solution.

Since (\ref{eqn:dro}) is more complex than (\ref{eqn:supervised}), considerable research efforts were devoted to develop efficient algorithms for DRO and its empirical version. For  $\P_0$ with finite support, \citet[Corollary 3]{DRO:MS:2013} have demonstrated that (\ref{eqn:dro}) with $f$-divergences is equivalent to a convex optimization problem, provided that the loss $\ell(\w;\z)$ is convex in $\w$. Actually, this conclusion is true even when $\P_0$ is continuous \citep[\S~3.2]{DRSP:SIAM:2017}. Under mild assumptions, \citet{DRO:MP:2018} show that DRO problems over Wasserstein balls can be reformulated as finite convex programs---in some cases even as linear programs. Besides the constrained formulation in (\ref{eqn:dro}), there also exists a penalized (or regularized) form of DRO \citep{Sinha:ICLR:2018}, which makes the optimization problem more tractable. In the past years, a series of SA methods have been proposed for empirical DRO with convex losses \citep{NIPS2016_4588e674}, and DRO with convex loss \citep{NEURIPS2020_64986d86} and non-convex losses \citep{NEURIPS2021_164bf317,NEURIPS2021_533fa796,Tianbao:22}.

The main focus of this paper is the GDRO problem in (\ref{eqn:group:dro})/(\ref{eqn:convex:concave}), instead of the traditional DRO in (\ref{eqn:dro}). \citet{Gouop_DRO} have applied SMD \citep{nemirovski-2008-robust} to (\ref{eqn:convex:concave}), but only obtain a sub-optimal sample complexity of $O(m^2 (\log m)/\epsilon^2)$, because of the large variance in their gradients. In the sequel, \citet{Online:Multiple:Distribution} and \citet{DRO:Online:Game} have tried to improve the sample complexity by reusing samples and applying techniques from MAB respectively, but their analysis suffers dependency issues. \citet[Proposition 2]{NEURIPS2022_e90b00ad} successfully established an $O(m (\log m)/\epsilon^2)$ sample complexity by combining SMD and gradient clipping, but their result holds only in expectation.  To deal with heterogeneous noise in different distributions,  \citet{Regret:RML:DS} propose a variant of GDRO named as minimax regret optimization (MRO),  which replaces the risk $R_i(\w)$ with ``excess risk'' $R_i(\w)-\min_{\w \in \W}R_i(\w)$.  More generally, calibration terms can be introduced to prevent any single distribution from dominating the maximum \citep{pmlr-v151-slowik22a}. Efficient optimization of MRO has been investigated by \citet{MERO}.

In the context of federated learning, \citet{pmlr-v97-mohri19a} have analyzed the generalization error of empricial GDRO when the number of samples from different distributions could be different. However, their convergence rate is unsatisfactory as it depends on the smallest number of samples and is distribution-independent.
Finally, we note that GDRO has a similar spirit with collaborative PAC learning \citep{NIPS2017_186a157b,NEURIPS2018_3569df15,pmlr-v139-rothblum21a} in the sense that both aim to find a single model that performs well on multiple distributions.

\section{SA Approaches to GDRO} \label{sec:sa:grdo}
In this section, we present two efficient SA approaches for GDRO, which achieve the same sample complexity but use a different number of samples in each round ($m$ versus $1$).

\subsection{Preliminaries}
First, we state the general setup of mirror descent \citep{nemirovski-2008-robust}.  We equip the domain $\W$ with a distance-generating function $\nu_w(\cdot)$, which is $1$-strongly convex with respect to certain norm $\|\cdot\|_w$. We define the Bregman distance associated with $\nu_w(\cdot)$ as
\[
B_w(\u,\v)= \nu_w(\u) - \big[\nu_w(\v) + \langle \nabla \nu_w(\v) , \u-\v\rangle\big].
\]
For the simplex $\Delta_m$, we choose the negative entropy (neg-entropy) function $\nu_q(\q)=\sum_{i=1}^m$ $ q_i \ln q_i$, which is $1$-strongly convex with respect to~the vector $\ell_1$-norm $\|\cdot\|_1$, as the distance-generating function. Similarly, $B_q(\cdot,\cdot)$ is the Bregman distance associated with $\nu_q(\cdot)$.

Then, we introduce the standard assumptions about the domain, and the loss function.
\begin{ass}\label{ass:1}
The domain $\W$ is convex and its diameter measured by $\nu_w(\cdot)$ is bounded by $\Dw$, i.e.,
\begin{equation} \label{eqn:domain:W}
 \max_{\w \in \W} \nu_w(\w) -\min_{\w \in \W} \nu_w(\w)  \leq \Dw^2 .
\end{equation}
\end{ass}
For $\Delta_m$, it is easy to verify that its diameter measured by the neg-entropy function is bounded by $\sqrt{\ln m}$.
\begin{ass}\label{ass:2a}
For all $i \in [m]$,  the risk function $R_i(\w)=\E_{\z \sim \P_i}[\ell(\w;\z)]$ is convex.
\end{ass}
To simplify  presentations, we assume the loss  belongs to $[0,1]$, and its gradient is also bounded.
%To simplify  presentations, we assume the loss and its gradient are also bounded.
\begin{ass}\label{ass:2}
For all $i \in [m]$, we have
\begin{equation} \label{eqn:value}
0 \leq \ell(\w;\z) \leq 1, \ \forall \w \in \W,  \ \z \sim \P_i.
\end{equation}
\end{ass}
\begin{ass}\label{ass:3}
For all $i \in [m]$, we have
\begin{equation} \label{eqn:gradient}
\|\nabla \ell(\w;\z) \|_{w,*}\leq G, \ \forall \w \in \W,  \ \z \sim \P_i
\end{equation}
where $\|\cdot\|_{w,*}$ is the dual norm of $\|\cdot\|_w$.
\end{ass}
Note that it is possible to relax the bounded assumptions in (\ref{eqn:value}) and (\ref{eqn:gradient}) to light tail conditions such as the sub-Gaussian Property \citep{HDP:Vershynin}.

Last, we discuss the performance measure. To analyze the convergence property, we measure the quality of an approximate solution $(\wb, \qb)$ to (\ref{eqn:convex:concave})  by the error
\begin{equation} \label{eqn:per:measure}
\epsilon_{\phi}(\wb, \qb) = \max_{\q\in \Delta_m}  \phi(\wb,\q)- \min_{\w\in \W}  \phi(\w,\qb)
\end{equation}
which directly controls the optimality of $\wb$ to the original problem (\ref{eqn:group:dro}), since
\begin{equation} \label{eqn:relation:error}
\begin{split}
&\max_{i\in[m]}  R_i(\wb) -  \min_{\w \in \W}   \max_{i\in[m]}  R_i(\w) = \max_{\q \in \Delta_m} \sum_{i=1}^m q_i R_i(\wb) -  \min_{\w \in \W} \max_{\q\in \Delta_m} \sum_{i=1}^m q_i R_i(\w) \\
\leq & \max_{\q \in \Delta_m} \sum_{i=1}^m q_i R_i(\wb) - \min_{\w\in \W} \sum_{i=1}^m \bar{q}_i R_i(\w) = \epsilon_{\phi}(\wb, \qb).
\end{split}
\end{equation}
%In the theoretical analysis, we will provide upper bounds for (\ref{eqn:per:measure}) that hold in expectation and with high probability.

\subsection{Stochastic Mirror Descent for GDRO} \label{sec:smd:1}
To apply SMD, the key is to construct stochastic gradients of the function $\phi(\w,\q)$ in (\ref{eqn:convex:concave}). We first present its true gradients with respect to~$\w$ and $\q$:
\begin{equation*}\label{eqn:true:grad}
\nabla_\w \phi(\w,\q) = \sum_{i=1}^m q_i \nabla R_i(\w), \textrm{ and } \nabla_\q \phi(\w,\q)= [R_1(\w), \ldots, R_m(\w)]^\top.
\end{equation*}
In each round $t$, denote by $\w_t$ and $\q_t$ the current solutions.  We draw one sample $\z_{t}^{(i)}$ from every distribution $\P_i$, and define stochastic gradients as
\begin{equation}\label{eqn:stoch:grad:1}
\g_{w}(\w_t,\q_t) = \sum_{i=1}^m q_{t,i} \nabla \ell(\w_{t};\z_t^{(i)}), \textrm{ and } \g_{q}(\w_t,\q_t)= [\ell(\w_{t};\z_t^{(1)}), \ldots, \ell(\w_{t};\z_t^{(m)})]^\top .
\end{equation}
Obviously, they are unbiased estimators of the true gradients:
\[
\E_{t-1}[\g_{w}(\w_t,\q_t)] = \nabla_\w \phi(\w_t,\q_t), \textrm{ and } \E_{t-1}[\g_{q}(\w_t,\q_t)]= \nabla_\q \phi(\w_t,\q_t)
 \]
where $\E_{t-1} [\cdot]$ represents the expectation conditioned on the randomness until round $t-1$. It is worth  mentioning that the construction of $\g_{w}(\w_t,\q_t)$ can be further simplified to
\begin{equation}\label{eqn:stoch:grad:2}
\gt_{w}(\w_t,\q_t)=\nabla \ell(\w_{t};\z_t^{(i_t)})
\end{equation}
where $i_t\in[m]$ is drawn randomly according to the probability $\q_t$.

% Change utilize to use because https://www.jmlr.org/format/format.html
Then, we use SMD to update $\w_t$ and $\q_t$:
\begin{align}
\w_{t+1}= &\argmin_{\w \in \W} \big\{ \eta_w \langle \g_{w}(\w_t,\q_t) , \w -\w_t\rangle + B_w(\w,\w_t) \big\} , \label{eqn:upate:wt}  \\
\q_{t+1}=& \argmin_{\q \in \Delta_m} \big\{ \eta_q \langle -\g_{q}(\w_t,\q_t) , \q -\q_t\rangle + B_q(\q,\q_t) \big\} \label{eqn:upate:qt:1}
\end{align}
where $\eta_w>0$ and $\eta_q>0$ are two step sizes that will be determined later.  The updating rule of $\w_t$ depends on the choice of the distance-generating function $\nu_w(\cdot)$. For example, if $\nu_w(\w)= \frac{1}{2} \|\w\|_2^2$, (\ref{eqn:upate:wt}) becomes stochastic gradient descent (SGD), i.e.,
\[
\w_{t+1}= \Pi_{\W}\big[ \w_t -\eta_w \g_{w}(\w_t,\q_t) \big]
\]
where $\Pi_{\W}[\cdot]$ denotes the Euclidean projection onto the nearest point in $\W$. Since $B_q(\q,\q_t)$ is defined in terms of the neg-entropy, (\ref{eqn:upate:qt:1}) is equivalent to
\begin{equation} \label{eqn:upate:qt:2}
q_{t+1, i}  = \frac{q_{t,i} \exp\big(\eta_q \ell(\w_{t};\z_t^{(i)}) \big) }{\sum_{j=1}^m q_{t,j} \exp\big(\eta_q \ell(\w_{t};\z_t^{(j)})\big) }, \ \forall i\in [m]
\end{equation}
which is the Hedge algorithm \citep{FREUND1997119} applied to a maximization problem.  In the beginning, we set $\w_1=\argmin_{\w \in \W}\nu_w(\w)$, and $\q_1=\frac{1}{m} \mathbf{1}_m$, where $\mathbf{1}_m$ is the $m$-dimensional vector consisting of $1$'s. In the last step, we return the averaged iterates $\wb=\frac{1}{T} \sum_{t=1}^T \w_t$  and $\qb=\frac{1}{T} \sum_{t=1}^T \q_t$ as  final solutions. The complete procedure is summarized in Algorithm~\ref{alg:1}.

\begin{algorithm}[t]
\caption{Stochastic Mirror Descent for GDRO}
{\bf Input}: step size $\eta_w$ and $\eta_q$
\begin{algorithmic}[1]
\STATE Initialize $\w_1=\argmin_{\w \in \W}\nu_w(\w)$, and $\q_1=[1/m, \ldots, 1/m]^\top \in \R^m$
\FOR{$t=1$ to $T$}
\STATE For each $i\in[m]$, draw a sample $\z_{t}^{(i)}$ from distribution $\P_i$
\STATE Construct the stochastic gradients defined in (\ref{eqn:stoch:grad:1})
\STATE Update $\w_t$ and $\q_t$ according to (\ref{eqn:upate:wt}) and (\ref{eqn:upate:qt:1}), respectively
\ENDFOR
\RETURN $\wb=\frac{1}{T} \sum_{t=1}^T \w_t$  and $\qb=\frac{1}{T} \sum_{t=1}^T \q_t$
\end{algorithmic}\label{alg:1}
\end{algorithm}
Based on the theoretical guarantee of SMD for stochastic convex-concave optimization \citep[\S~3.1]{nemirovski-2008-robust}, we have the following theorem for Algorithm~\ref{alg:1}.
\begin{theorem} \label{thm:1} Under Assumptions~\ref{ass:1}, \ref{ass:2a}, \ref{ass:2} and \ref{ass:3}, and setting $\eta_w =  \Dw^2\sqrt{\frac{8  }{5T (\Dw^2 G^2+\ln m)}}$ and $\eta_q = (\ln m)\sqrt{\frac{8}{5T (\Dw^2 G^2+\ln m)}}$ in Algorithm~\ref{alg:1}, we have
 \[
 \E \big[ \epsilon_{\phi}(\wb, \qb) \big] \leq 2 \sqrt{\frac{10 (\Dw^2 G^2 + \ln m )}{T} }
 \]
 and with probability at least $1-\delta$,
\[
\epsilon_{\phi}(\wb, \qb) \leq  \left(8+2\ln \frac{2}{\delta}\right) \sqrt{\frac{10(\Dw^2 G^2 + \ln m )}{T} }.
\]
\end{theorem}
\begin{myremark}
\textnormal{
Theorem~\ref{thm:1} shows that Algorithm~\ref{alg:1} achieves an $O(\sqrt{(\log m)/T})$ convergence rate. Since it consumes $m$ samples per iteration, the sample complexity is $O(m (\log m)/\epsilon^2)$, which nearly matches the $\Omega(m/\epsilon^2)$ lower bound \citep[Theorem 5]{DRO:Online:Game}.}
\end{myremark}

\paragraph{Comparisons with \citet{Gouop_DRO}} 
Given the fact that the number of samples used in each round of Algorithm~\ref{alg:1} is $m$, it is natural to ask whether it can be reduced to a small constant. Indeed, the stochastic algorithm of \citet{Gouop_DRO} only requires $1$ sample per iteration, but suffers a large  sample complexity. In each round $t$, they first generate a random index $i_t \in [m]$ uniformly, and  draw $1$ sample $\z_t^{(i_t)}$ from $\P_{i_t}$. The stochastic gradients are constructed as follows:
\begin{equation}\label{eqn:stoch:grad:3}
\gh_{w}(\w_t,\q_t) = q_{t,i_t}m \nabla \ell(\w_{t};\z_t^{(i_t)}), \textrm{ and } \gh_{q}(\w_t,\q_t)=  [0, \ldots,m \ell(\w_{t};\z_t^{(i_t)}), \ldots,0]^\top
\end{equation}
where $\gh_{q}(\w_t,\q_t)$ is a vector with $m \ell(\w_{t};\z_t^{(i_t)})$ in position $i_t$ and $0$ elsewhere. Then, the two stochastic gradients are used to update $\w_t$ and $\q_t$, in the same way as  (\ref{eqn:upate:wt})  and (\ref{eqn:upate:qt:1}). However, it only attains a slow convergence rate: $O(m \sqrt{(\log m)/T})$, leading to an  $O(m^2 (\log m)/\epsilon^2)$  sample complexity, which is higher than that of Algorithm~\ref{alg:1} by a factor of $m$. The slow convergence is due to the fact that the optimization error depends on the dual norm of the stochastic gradients in (\ref{eqn:stoch:grad:3}), which blows up by a factor of $m$, compared with the gradients in (\ref{eqn:stoch:grad:1}).

\paragraph{Comparisons with \citet{Online:Multiple:Distribution}} 
To reduce the number of samples required in each round, \citet{Online:Multiple:Distribution} propose to reuse samples for multiple iterations. To approximate $\nabla_\w \phi(\w_t,\q_t)$, they construct the stochastic gradient $\gt_{w}(\w_t,\q_t)$ in the same way as (\ref{eqn:stoch:grad:2}), which needs $1$ sample. To approximate $\nabla_\q \phi(\w_t,\q_t)$, they draw $m$ samples $\z_{\tau}^{(1)},\ldots,  \z_{\tau}^{(m)}$, one from each distribution,  at round $\tau=mk+1$, $k \in\mathbb{Z}$, and reuse them for $m$ iterations to construct the following gradient:
\begin{equation} \label{eqn:gradient:wrong}
\g'_{q}(\w_t,\q_t)= [\ell(\w_{t};\z_\tau^{(1)}), \ldots, \ell(\w_{t};\z_\tau^{(m)})]^\top, \  t=\tau,\ldots, \tau+m-1.
\end{equation}
Then, they treat $\gt_{w}(\w_t,\q_t)$ and $\g'_{q}(\w_t,\q_t)$ as stochastic gradients, and update $\w_t$ and $\q_t$ by SMD. In this way, their algorithm uses $2$ samples on average in each iteration. However, the gradient in (\ref{eqn:gradient:wrong}) is no longer an unbiased estimator of the true gradient $\nabla_\q \phi(\w_t,\q_t)$ at rounds $t=\tau+2,\ldots,\tau+m-1$, making their analysis ungrounded. To see this, from the updating rule of SMD, we know that $\w_{\tau+2}$ depends on $\q_{\tau+1}$, which in turn depends on the $m$ samples drawn at round $\tau$, and thus
\[
\E \left[\ell(\w_{\tau+2}; \z_\tau^{(i)} )\right] \neq R_i(\w_{\tau+2}), \ i =1,\ldots, m.
\]
%$\g_{q}(\w_t,\q_t)$

\subsubsection{Anytime Extensions} \label{sec:anytime:smd}
The step sizes $\eta_w$ and $\eta_q$ in Theorem~\ref{thm:1} depend on the total number of iterations $T$, which complicates practical implementation as it requires setting $T$ beforehand. Additionally, the theorem only offers theoretical guarantees for the final solution. To avoid these limitations,  we propose an anytime extension of Algorithm~\ref{thm:1} by employing time-varying step sizes. We note that there is a long-standing history of designing anytime algorithms in optimization and related areas \citep{anytime_IS,10.5555/2074094.2074123,pmlr-v97-cutkosky19a}.

Specifically, we replace the fixed step sizes $\eta_w$ and $\eta_q$ in (\ref{eqn:upate:wt}) and (\ref{eqn:upate:qt:1}) with time-varying step sizes \citep{nemirovski-2008-robust}
\begin{equation} \label{eqn:alg1_anytime_stepsize}
	\eta^w_t =  \Dw^2\sqrt{\frac{2}{t(\Dw^2 G^2+\ln m)}}, \textrm{ and }  \eta^q_t = (\ln m)\sqrt{\frac{2}{t(\Dw^2 G^2+\ln m)}},
\end{equation}
respectively. To enable anytime capability, we maintain the weighted averages of the iterates: 
\begin{equation} \label{eqn:anytime_output}
\begin{split}
\wb_t=&\sum_{j=1}^t\frac{\eta^w_j\w_j}{\sum_{k=1}^t\eta^w_k}= \frac{(\sum_{j=1}^{t-1} \eta^w_j)\wb_{t-1} + \eta^w_t \w_t}{\sum_{k=1}^t \eta^w_k}, \\
\qb_t=&\sum_{j=1}^t\frac{\eta^q_j\q_j}{\sum_{k=1}^t\eta^q_k}= \frac{(\sum_{j=1}^{t-1} \eta^q_j)\qb_{t-1} + \eta^q_t \q_t}{\sum_{k=1}^t \eta^q_k}
\end{split}
\end{equation} 
which can be returned as solutions whenever required, and provide the following theoretical guarantee  for the solution $(\wb_t, \qb_t)$ at each round $t$.
\begin{theorem} \label{thm:alg1_anytime} Under Assumptions~\ref{ass:1}, \ref{ass:2a}, \ref{ass:2} and \ref{ass:3}, and setting step sizes as \eqref{eqn:alg1_anytime_stepsize} in Algorithm~\ref{alg:1},  we have
\begin{equation} \label{thm:alg1_anytime_exp}
\E \big[ \epsilon_{\phi}(\wb_t, \qb_t) \big] \leq \frac{\sqrt{\Dw^2 G^2+\ln m}}{\sqrt{2}\left(\sqrt{t+1}-1\right)} \left(5+3\ln t\right)=O\left(\frac{\sqrt{\log m} \log t}{\sqrt{t}}\right), \ \forall  t \in \zn.
\end{equation}
Furthermore, with probability at least $1-\delta$,  we have
	\[
	\epsilon_{\phi}(\wb_t, \qb_t)\leq\frac{\sqrt{\Dw^2 G^2+ \ln m}}{\sqrt{2}\left(\sqrt{t+1}-1\right)}
	\left(9+11\ln\frac{2}{\delta}+7\ln t+3\ln\frac{2}{\delta}\ln t\right)=O\left(\frac{\sqrt{\log m} \log t}{\sqrt{t}}\right)
	\]
for each $t \in \zn$.
\end{theorem}

\begin{myremark}
\textnormal{The convergence rate of the anytime extension  is slower by a factor of  $O(\log t)$ compared to Algorithm~\ref{alg:1} with fixed step sizes. However, the modified algorithm possesses the anytime characteristic, indicating it is capable of returning a solution at any round.}
\end{myremark}

\subsection{Non-oblivious Online Learning for GDRO} \label{sec:smd:2}
In this section, we explore methods to reduce the number of samples used in each iteration from $m$ to $1$. As shown in (\ref{eqn:stoch:grad:2}), we can use $1$ sample to construct a stochastic gradient for $\w_t$ with small norm, since $\|\gt_{w}(\w_t,\q_t)\|_{w,*}\leq G$ under Assumption~\ref{ass:3}. Thus, it is relatively easy to control the error related to $\w_t$.  However, we do not have such guarantees for the stochastic gradient of $\q_t$. Recall that the infinity norm of $\gh_{q}(\w_t,\q_t)$ in (\ref{eqn:stoch:grad:3}) is upper bounded by $m$. The reason is that we insist on the unbiasedness of the stochastic gradient, which leads to a large variance. To control the variance, \citet{NEURIPS2022_e90b00ad} have applied gradient clipping to $\gh_{q}(\w_t,\q_t)$, and established an $O(m (\log m)/\epsilon^2)$ sample complexity that holds in expectation. Different from their approach, we borrow techniques from online learning to balance the bias and the variance.

In the studies of convex-concave saddle-point problems, it is now well-known that they can be solved by playing two online learning algorithms against each other \citep{FREUND199979,Predictable:NIPS:2013,Game:NIPS:15,Online:Bandit:Minimax}. This transformation allows us to exploit no-regret algorithms developed in online learning to bound the optimization error. To solve  problem  (\ref{eqn:convex:concave}), we ask the 1st player to minimize a sequence of convex functions
\begin{equation}\label{eqn:first:player}
\phi(\w,\q_1)=\sum_{i=1}^m q_{1,i} R_i(\w), \ \ \phi(\w,\q_2)=\sum_{i=1}^m q_{2,i} R_i(\w), \ \cdots, \ \phi(\w,\q_T)=\sum_{i=1}^m q_{T,i} R_i(\w)
\end{equation}
under the constraint $\w \in \W$, and the 2nd player to maximize a sequence of linear functions
\begin{equation}\label{eqn:second:player}
\phi(\w_1,\q)=\sum_{i=1}^m q_{i} R_i(\w_1), \ \  \phi(\w_2,\q)=\sum_{i=1}^m q_{i} R_i(\w_2), \ \cdots, \ \phi(\w_T,\q)=\sum_{i=1}^m q_{i} R_i(\w_T)
\end{equation}
subject to the constraint $\q \in \Delta_m$. We highlight that there exists an important difference between our stochastic convex-concave problem and its deterministic counterpart. Here, the two players cannot directly observe the loss function, and can only approximate $R_i(\w)=\E_{\z \sim \P_i}\big[\ell(\w;\z)\big]$ by drawing samples from $\P_i$. The stochastic setting makes the problem more challenging, and in particular, we need to take care of the \emph{non-oblivious} nature of the learning process. Here, ``non-oblivious'' refers to the fact that the online functions depend on the past decisions of the players.

Next, we discuss the online algorithms that will be used by the two players. As shown in Section~\ref{sec:smd:1}, the 1st player can easily obtain a stochastic gradient with small norm by using $1$ sample. So, we model the problem faced by the 1st player as ``non-oblivious online convex optimization (OCO) with stochastic gradients'', and still use SMD to update its solution. In each round $t$, with $1$ sample drawn from $\P_{i}$, the 2nd player can estimate the value of $R_{i}(\w_t)$ which is the coefficient of $q_{i}$. Since the 2nd player is maximizing a linear function over the simplex, the problem can be modeled as ``non-oblivious multi-armed bandits (MAB) with stochastic rewards''. And fortunately, we have powerful online algorithms for non-oblivious MAB \citep{Auer:2003:NMB,2020:bandit-book}, whose regret has a sublinear dependence on $m$. In this paper, we choose the Exp3-IX algorithm \citep{NIPS2015_e5a4d6bf}, and generalize its theoretical guarantee to stochastic rewards. In contrast, if we apply SMD with $\gh_{q}(\w_t,\q_t)$ in (\ref{eqn:stoch:grad:3}), the regret scales at least linearly with $m$.

\begin{algorithm}[t]
\caption{Non-oblivious Online Learning for GDRO}
{\bf Input}: step size $\eta_w$ and $\eta_q$, and IX coefficient $\gamma$
\begin{algorithmic}[1]
\STATE Initialize $\w_1=\argmin_{\w \in \W}\nu_w(\w)$, and $\q_1=[1/m, \ldots, 1/m]^\top \in \R^m$
\FOR{$t=1$ to $T$}
\item Generate $i_t \in [m]$ according to $\q_t$, and draw a sample $\z_t^{(i_t)}$ from distribution $\P_{i_t}$
\STATE Construct the stochastic gradient in (\ref{eqn:stoch:grad:2}) and the IX loss estimator in (\ref{eq:IX-loss-estimator})
\STATE Update $\w_t$ and $\q_t$ according to (\ref{eqn:upate:wt:2}) and (\ref{eq:algo-q}), respectively
\ENDFOR
\RETURN $\wb=\frac{1}{T} \sum_{t=1}^T \w_t$  and $\qb=\frac{1}{T} \sum_{t=1}^T \q_t$
\end{algorithmic}\label{alg:2}
\end{algorithm}
The complete procedure is presented in Algorithm~\ref{alg:2}, and we explain key steps below. In each round $t$, we generate an index $i_t \in [m]$ from the probability distribution $\q_t$, and then draw a sample $\z_t^{(i_t)}$ from the distribution $\P_{i_t}$. With the stochastic gradient in (\ref{eqn:stoch:grad:2}), we use SMD to update $\w_t$:
\begin{equation} \label{eqn:upate:wt:2}
\w_{t+1}= \argmin_{\w \in \W} \big\{ \eta_w \langle \gt_{w}(\w_t,\q_t) , \w -\w_t\rangle + B_w(\w,\w_t) \big\}.
\end{equation}
Then, we reuse the sample $\z_t^{(i_t)}$  to update $\q_t$ according to Exp3-IX, which first constructs the Implicit-eXploration (IX) loss estimator \citep{NIPS'14:IX-estimator}:
\begin{equation} \label{eq:IX-loss-estimator}
    \tildes_{t,i} = \frac{1 - \ell(\w_t,\z_t^{(i_t)})}{q_{t,i} + \gamma} \cdot \ind[i_t = i], \ \forall i \in [m]
\end{equation}
where $\gamma >0$ is the IX coefficient and $\ind[A]$ equals to $1$ when the event $A$ is true and $0$ otherwise,  and then performs a mirror descent update:
\begin{equation}    \label{eq:algo-q}
    \q_{t+1} = \argmin_{\q \in \Delta_m} \big \{\eta_q \langle\tildesb_t, \q - \q_t\rangle + B_{q}(\q,\q_t)\big\}.
\end{equation}
Compared with (\ref{eqn:upate:qt:1}), the only difference is that the stochastic gradient $-\g_{q}(\w_t,\q_t)$ is now replaced with the IX loss estimator $\tildesb_t$. However, it is not an instance of SMD, because $\tildesb_t$ is no longer an unbiased stochastic gradient. The main advantage of $\tildesb_t$ is that it reduces the variance of the gradient estimator by sacrificing a little bit of unbiasedness, which turns out to be crucial for a high probability guarantee, and thus can deal with non-oblivious adversaries. Since we still use the entropy regularizer in (\ref{eq:algo-q}), it also enjoys an explicit form that is similar to (\ref{eqn:upate:qt:2}).

We present the theoretical guarantee of Algorithm~\ref{alg:2}. To this end, we first bound the regret of the 1st player. In the analysis, we address the non-obliviousness by the ``ghost iterate'' technique of \citet{nemirovski-2008-robust}.
\begin{theorem} \label{thm:2} Under Assumptions~\ref{ass:1}, \ref{ass:2a} and \ref{ass:3}, and setting $\eta_w=\frac{2\Dw}{G\sqrt{5T}}$, we have
\[
\E \left[ \sum_{t=1}^T \phi(\w_t,\q_t) - \min_{\w \in \W} \sum_{t=1}^T \phi(\w,\q_t) \right] \leq 2 \Dw G \sqrt{5 T}
\]
and with probability at least $1-\delta$,
\[
\sum_{t=1}^T \phi(\w_t,\q_t) - \min_{\w \in \W} \sum_{t=1}^T \phi(\w,\q_t) \leq   \Dw G \sqrt{T} \left( 2 \sqrt{5} +8 \sqrt{\ln \frac{1}{\delta}} \right).
\]
\end{theorem}

By extending Exp3-IX to stochastic rewards, we have the following bound for the 2nd player.

\begin{theorem} \label{thm:3} Under Assumption \ref{ass:2}, and setting $\eta_q = \sqrt{\frac{\ln m}{mT}}$ and the IX coefficient $\gamma = \frac{\eta_q}{2}$, we have
\[
    \E\left[\max_{\q \in \Delta_m} \sum_{t=1}^T \phi(\w_t,\q) - \sum_{t=1}^T \phi(\w_t,\q_t)\right] \leq 3\sqrt{m T \ln m} + \sqrt{\frac{T}{2}} + 3\left(\sqrt{\frac{mT}{\ln m}} + \sqrt{\frac{T}{2}} + 1 \right)
\]
and with probability at least $1-\delta$,
\[
    \max_{\q \in \Delta_m} \sum_{t=1}^T \phi(\w_t,\q) - \sum_{t=1}^T \phi(\w_t,\q_t) \leq 3\sqrt{m T \ln m} + \sqrt{\frac{T}{2}} + \left(\sqrt{\frac{mT}{\ln m}} + \sqrt{\frac{T}{2}} + 1 \right)\ln \frac{3}{\delta}.
\]
\end{theorem}

Combining the above two theorems directly leads to the following optimization error bound.

\begin{theorem} \label{thm:4} Under Assumptions~\ref{ass:1}, \ref{ass:2a}, \ref{ass:2} and \ref{ass:3}, and setting $\eta_w=\frac{2\Dw}{G\sqrt{5T}}$, $\eta_q = \sqrt{\frac{\ln m}{m T}}$ and $\gamma = \frac{\eta_q}{2}$ in Algorithm~\ref{alg:2}, we have
\begin{equation} \label{eqn:alg2:exp}
\E \big[ \epsilon_{\phi}(\wb, \qb) \big] \leq 2DG\sqrt{\frac{5}{T}} + 3\sqrt{\frac{m  \ln m}{T}} + \sqrt{\frac{1}{2T}} + 3\left(\sqrt{\frac{m}{T\ln m}} + \sqrt{\frac{1}{2T}} + \frac{1}{T} \right)
\end{equation}
and with probability at least $1-\delta$,
\begin{equation} \label{eqn:alg2:high}
\begin{split}
    &\epsilon_{\phi}(\wb, \qb) \\
\leq &\Dw G \sqrt{\frac{1}{T}} \left( 2 \sqrt{5} +8 \sqrt{ \ln \frac{2}{\delta}} \right) + 3\sqrt{\frac{m  \ln m}{T}} + \sqrt{\frac{1}{2T}} + \left(\sqrt{\frac{m}{T\ln m}} + \sqrt{\frac{1}{2T}} + \frac{1}{T} \right)\ln \frac{6}{\delta}.
\end{split}
\end{equation}
\end{theorem}

\begin{myremark}
\textnormal{The above theorem shows that with $1$ sample per iteration,  Algorithm~\ref{alg:2} is able to achieve an $O(\sqrt{ m (\log m)/T})$ convergence rate, thus maintaining the $O(m (\log m)/\epsilon^2)$ sample complexity. It is worth  mentioning that one may attempt to reduce the $\log m$ factor by employing mirror descent with the Tsallis entropy ($\nu_q(\q) = 1 - \sum_{i=1}^m \sqrt{q_i}$) for the 2nd player \citep[Theorem~13]{JMLR'10:AudibertB10}. However, even in the standard MAB problem, such an improvement only happens in the oblivious setting, and is conjectured to be impossible in the non-oblivious case \citep[Remark~14]{JMLR'10:AudibertB10}.}
\end{myremark}

\paragraph{Comparisons with \citet{DRO:Online:Game}} 
In a recent work, \citet{DRO:Online:Game} have deployed online algorithms to optimize $\w$ and $\q$, but did not consider the non-oblivious property. As a result, their theoretical guarantees, which build upon the analysis  for oblivious online learning \citep{Modern:Online:Learning}, cannot justify the optimality of their algorithm for (\ref{eqn:convex:concave}). Specifically, their results imply that for any \emph{fixed} $\w$ and $\q$ that are independent from $\wb$ and $\qb$ \citep[Theorem 3]{DRO:Online:Game},
\begin{equation} \label{eqn:exp:regret:weak}
\E \left[\phi(\wb,\q)- \phi(\w,\qb) \right] = O \left( \sqrt{\frac{m}{T}} \right).
\end{equation}
However, (\ref{eqn:exp:regret:weak}) cannot be used to bound $\epsilon_{\phi}(\wb, \qb)$  in (\ref{eqn:per:measure}), because of the dependency issue. To be more clear, we have
\[
\epsilon_{\phi}(\wb, \qb) = \max_{\q\in \Delta_m}  \phi(\wb,\q)- \min_{\w\in \W}  \phi(\w,\qb)=\phi(\wb,\qh)- \phi(\wh,\qb),
 \]
where $\wh=\argmin_{\w \in \W}  \phi(\w,\qb)$ and $\qh= \argmax_{\q \in \Delta_m} \phi(\wb,\q)$ \emph{depend} on $\qb$ and $\wb$, respectively.

\begin{myremark}
\textnormal{After we pointed out the dependence issue of reusing samples, \citet{Haghtalab:GDRO:2023} modified their method by incorporating bandit algorithms to optimize $\q$. From our understanding, the idea of applying bandits to GDRO was \emph{firstly} proposed by \citet{DRO:Online:Game}, and subsequently refined by us.}
\end{myremark}

\subsubsection{Anytime Extensions} \label{sec:online:anytime}
Similar to Algorithm~\ref{alg:1}, Algorithm~\ref{alg:2} also requires the prior specification of the total number of iterations $T$, as the values of $\eta_w$ in SMD, as well as $\eta_q$ and $\gamma$ in Exp3-IX, are dependent on $T$. Following the extension in Section~\ref{sec:anytime:smd}, we can also adapt Algorithm~\ref{alg:2} to be anytime by employing time-varying parameters in SMD and Exp3-IX. Specifically, in the $t$-th round, we replace $\eta_w$ in \eqref{eqn:upate:wt:2}, $\eta_q$ in \eqref{eq:algo-q}, and $\gamma$ in \eqref{eq:IX-loss-estimator} with
\begin{equation} \label{eqn:alg2_anytime_stepsize}
\eta^w_t = \frac{D}{G\sqrt{t}}, \ \eta^q_t=\sqrt{\frac{\ln m}{mt}}, \textrm{ and }  \gamma_{t}=\frac{\eta^q_t}{2}
\end{equation}
respectively, and output $\wb_t$ and $\qb_t$ in \eqref{eqn:anytime_output} as the current solution. 

Compared to the original Algorithm~\ref{alg:2}, our modifications are relatively minor. However, the theoretical analysis differs significantly. The reason is because the optimization error of $(\wb_t, \qb_t)$ is governed by the \emph{weighted average regret} of the two players, rather than the standard regret. That is,
\begin{equation} \label{eqn:alg2_anytime_O1O2}
\begin{split}
 &\epsilon_{\phi}(\wb_t, \qb_t) = \max_{\q\in \Delta_m}  \phi(\wb_t,\q)- \min_{\w\in \W}  \phi(\w,\qb_t)\\
\leq & \underbrace{\left(\sum_{j=1}^t\eta^w_j\right)^{-1} \left(\max_{\w\in \W} \sum_{j=1}^t \eta^w_j\left[\phi(\w_j,\q_j)- \phi(\w,\q_j)\right]\right) }_{:=O_1}\\
		&+\underbrace{\left(\sum_{j=1}^t\eta^q_j\right)^{-1} \left(\max_{\q\in \Delta_m}\sum_{j=1}^t \eta^q_j \left[\phi(\w_j,\q)-\phi(\w_j,\q_j)\right]\right)}_{:=O_2}.
	\end{split}
\end{equation} 

For the 1st player, we extend the analysis of SMD in Theorem~\ref{thm:2}, and obtain the results below for bounding  $O_1$.
\begin{theorem} \label{thm:alg2_anytime_w} 
Under Assumptions~\ref{ass:1}, \ref{ass:2a} and \ref{ass:3}, and using $\eta^w_t$ in \eqref{eqn:alg2_anytime_stepsize} for the 1st player, we have
	\[
\E\big[O_1\big] \leq \frac{DG}{\left(\sqrt{t+1}-1\right)}\left(\frac{9}{4} + \frac{5}{4} \ln t\right), \ \forall  t \in \zn.
	\]
Furthermore, with probability at least $1-\delta$,  we have
\begin{equation*}
O_1\leq \frac{DG}{\sqrt{t+1}-1}\left(\frac{17}{4}+\frac{13}{4}\ln t+2\ln\frac{1}{\delta}\right)
\end{equation*}
for each $t \in \zn$.
\end{theorem}

While \citet{NIPS2015_e5a4d6bf} have analyzed the regret of Exp3-IX with time-varying step sizes, our focus is on the weighted average regret $O_2$. To achieve this, we conduct a different analysis to bound $O_2$, and establish the following theoretical guarantee.
\begin{theorem} \label{thm:alg2_anytime_q} 
Under Assumption \ref{ass:2}, and using $\eta^q_t$ and  $\gamma_t$ in \eqref{eqn:alg2_anytime_stepsize} for the 2nd player, we have
\[
\E\big[O_2\big]\leq \frac{1}{2\left(\sqrt{t+1}-1\right)}\left(\left(3+\ln t\right)\sqrt{m\ln m}+6\sqrt{\frac{m}{\ln m}}+4\sqrt{\frac{1+\ln t}{2}}\right), \ \forall  t \in \zn.
\]
Furthermore, with probability at least $1-\delta$,  we have
\begin{equation*}
O_2
\leq
\frac{1}{2\left(\sqrt{t+1}-1\right)} \left(\left(3+\ln t\right)\sqrt{m\ln m}+\left(2\sqrt{\frac{m}{\ln m}}+\sqrt{\frac{1+\ln t}{2}}\right)\ln\frac{3}{\delta}+\sqrt{\frac{1+\ln t}{2}}\right)
\end{equation*}
for each $t \in \zn$.
\end{theorem}

By directly integrating the above two theorems, we derive the following theorem for the optimization error at each round.
\begin{theorem} \label{thm:alg2_anytime} 
Under Assumptions~\ref{ass:1}, \ref{ass:2a}, \ref{ass:2} and \ref{ass:3}, and setting parameters as \eqref{eqn:alg2_anytime_stepsize} in Algorithm~\ref{alg:2},  we have
\begin{equation} \label{eqn:alg2_anytime_thm_exp} 
	\begin{split}
	\E \big[ \epsilon_{\phi}(\wb_t, \qb_t) \big] 
	\leq &\frac{\left(3+\ln t\right)\sqrt{m\ln m}+6\sqrt{m/\ln m}+4\sqrt{(1+\ln t)/2}+DG\left(5+3\ln t\right)}{2\left(\sqrt{t+1}-1\right)}\\
	= & O\left(\frac{\sqrt{m \log m}\log t}{\sqrt{t}}\right),  \quad\quad\quad\quad  \forall  t \in \zn.
\end{split}
\end{equation}
Furthermore, with probability at least $1-\delta$,  we have
\begin{equation} \label{eqn:alg2_anytime_thm_highpro} 
	\begin{split}
	&\epsilon_{\phi}(\wb_t, \qb_t) \\
	\leq & \frac{\left(3+\ln t\right)\sqrt{m\ln m}+\left(2\sqrt{\frac{m}{\ln m}}+\sqrt{\frac{1+\ln t}{2}}\right)\ln\frac{6}{\delta}+\sqrt{\frac{1+\ln t}{2}}+DG\left(9+7\ln t+4\ln\frac{2}{\delta}\right)}{2\left(\sqrt{t+1}-1\right)} \\
	= & O\left(\frac{\sqrt{m \log m}\log t}{\sqrt{t}}\right)
	\end{split}
	\end{equation}
for each $t \in \zn$.
\end{theorem}
\begin{myremark}
\textnormal{Similar to the conclusion in Section~\ref{sec:anytime:smd}, the convergence rate in the above theorem is $O(\log t)$ times slower than that in Theorem~\ref{thm:4}.}
\end{myremark}

\section{Weighted GDRO for Imbalanced Data} \label{sec:sa:weight:grdo}
When designing SA approaches for GDRO, it is common to assume that the algorithms are free to draw samples from every distribution \citep{Gouop_DRO}, as we do in Section~\ref{sec:sa:grdo}. However, this assumption may not hold in practice. For example, data collection costs can vary widely among distributions~\citep{RADIVOJAC2004224}, and data collected from various channels can have different throughputs~\citep{zhou2023stream}. In this section, we investigate the scenario where the number of samples can be drawn from each distribution could be different. Denote by $n_i$ the number of samples that can be drawn from $\P_i$. Without loss of generality, we assume that $n_1 \geq n_2 \geq \cdots \geq n_m$. Note that we have a straightforward \textbf{Baseline} which just runs Algorithm~\ref{alg:1} for $n_m$ iterations, and the optimization error $\epsilon_{\phi}(\wb, \qb)=O(\sqrt{(\log m) /n_m})$.

\subsection{Stochastic Mirror Descent with Non-uniform Sampling}
To meet the budget, we propose to incorporate non-uniform sampling into SMD. Before getting into technical details, we first explain the main idea of using non-uniform sampling. One way is to draw $1$ sample from every distribution $\P_i$ with probability $p_i=n_i/n_1$ in each iteration. Then, after $n_1$ iterations, the \emph{expected} number of samples drawn from $\P_i$ will be $n_1 p_i=n_i$, and thus the budget is satisfied in expectation.% Of course, other approach is also possible, e.g., running the algorithm for $n$ iterations, and in each iteration selecting $1$ distribution with probability $[n_1/n, \ldots, n_m/n]$ and drawing $1$ sample from the chosen distribution.

Specifically, in each round $t$, we first generate a set of Bernoulli random variables  $\{b_t^{(1)},\ldots,b_t^{(m)}\}$ with $\Pr[b_t^{(i)}=1]=p_i$ to determine whether to sample from each distribution. If $b_t^{(i)}=1$, we draw a sample $\z_t^{(i)}$ from $\P_i$. The question then becomes how to construct stochastic gradients from these samples. Let $\C_t=\{i|b_t^{(i)}=1\}$ be the indices of selected distributions. If we stick to the original problem in (\ref{eqn:convex:concave}), then the stochastic gradients should be constructed in the following way
\begin{equation}\label{eqn:stoch:grad:4}
\g_{w}(\w_t,\q_t) = \sum_{i\in C_t}  \frac{q_{t,i}}{p_i}\nabla \ell(\w_{t};\z_t^{(i)}), \textrm{ and } [\g_{q}(\w_t,\q_t)]_i= \left\{
\begin{array}{ll}                           \ell(\w_{t};\z_t^{(i)})/p_i, & i \in \C_t \\
0, & \textrm{otherwise}
\end{array}
\right.
\end{equation}
to ensure unbiasedness. Then, they can be used by SMD to update $\w_t$ and $\q_t$. To analyze the optimization error, we need to bound the norm of stochastic gradients in (\ref{eqn:stoch:grad:4}). To this end, we have $\|\g_{w}(\w_t,\q_t)\|_{w,*} \leq G n_1/n_m$ and $\|\g_{q}(\w_t,\q_t)_i\|_\infty \leq n_1/n_m$. Following the arguments of Theorem~\ref{thm:1}, we can prove that the error $\epsilon_{\phi}(\wb, \qb)=O( \sqrt{(\log m) /n_1} \cdot n_1/n_m)= O(\sqrt{n_1\log m}/n_m)$, which is even larger than the $O(\sqrt{(\log m) /n_m})$ error of the Baseline.

In the following, we demonstrate that a simple twist of the above procedure can still yield meaningful results that are complementary to the Baseline. We observe that the large norm of the stochastic gradients in (\ref{eqn:stoch:grad:4}) is caused by the inverse probability $1/p_i$. A natural idea is to ignore $1/p_i$, and define the following stochastic gradients:
\begin{equation}\label{eqn:stoch:grad:5}
\g_{w}(\w_t,\q_t) = \sum_{i\in C_t}  q_{t,i}\nabla \ell(\w_{t};\z_t^{(i)}), \textrm{ and } [\g_{q}(\w_t,\q_t)]_i= \left\{
\begin{array}{ll}
\ell(\w_{t};\z_t^{(i)}), & i \in \C_t \\
 0, & \textrm{otherwise.}
\end{array}
\right.
\end{equation}
In this way, they are no longer stochastic gradients of (\ref{eqn:convex:concave}), but can be treated as stochastic gradients of a weighted GDRO problem:
\begin{equation} \label{eqn:convex:concave:weight}
\min_{\w \in \W} \max_{\q \in \Delta_m}  \   \left\{\varphi(\w,\q)= \sum_{i=1}^m q_i p_i R_i(\w) \right\}
\end{equation}
where each risk $R_i(\cdot)$ is scaled by a factor $p_i$. Based on the gradients in (\ref{eqn:stoch:grad:5}),  we still use (\ref{eqn:upate:wt}) and (\ref{eqn:upate:qt:1}) to update $\w_t$ and $\q_t$. We summarize the complete procedure in Algorithm~\ref{alg:3}.

\begin{algorithm}[t]
\caption{Stochastic Mirror Descent for Weighted GDRO}
{\bf Input}: step size $\eta_w$ and $\eta_q$
\begin{algorithmic}[1]
\STATE Initialize $\w_1=\argmin_{\w \in \W}\nu_w(\w)$, and $\q_1=[1/m, \ldots, 1/m]^\top \in \R^m$
\FOR{$t=1$ to $n_1$}
\STATE For each $i\in[m]$, generate a Bernoulli random variable $b_t^{(i)}$ with $\Pr[b_t^{(i)}=1]=p_i$, and if $b_t^{(i)}=1$, draw a sample $\z_{t}^{(i)}$ from distribution $\P_i$
\STATE Construct the stochastic gradients defined in (\ref{eqn:stoch:grad:5})
\STATE Update $\w_t$ and $\q_t$ according to (\ref{eqn:upate:wt}) and (\ref{eqn:upate:qt:1}), respectively
\ENDFOR
\RETURN $\wb=\frac{1}{n_1} \sum_{t=1}^{n_1} \w_t$  and $\qb=\frac{1}{n_1} \sum_{t=1}^{n_1} \q_t$
\end{algorithmic}\label{alg:3}
\end{algorithm}

We omit the optimization error of Algorithm~\ref{alg:3} for (\ref{eqn:convex:concave:weight}), since it has exactly the same form as Theorem~\ref{thm:1}. What we are really interested in is the theoretical guarantee of its solution on multiple distributions. To this end, we have the following theorem.
\begin{theorem} \label{thm:5}   Under Assumptions~\ref{ass:1}, \ref{ass:2a}, \ref{ass:2} and \ref{ass:3}, and setting $\eta_w =  \Dw^2\sqrt{\frac{8  }{5n_1(\Dw^2 G^2+\ln m)}}$ and $\eta_q = (\ln m)\sqrt{\frac{8}{5n_1 (\Dw^2 G^2+\ln m)}}$ in Algorithm~\ref{alg:3},  with probability at least $1-\delta$, we have
 \[
 R_i(\wb) - \frac{n_1}{n_i} p_{\varphi}^* \leq \frac{1}{p_i} \mu(\delta) \sqrt{\frac{10 (\Dw^2 G^2 + \ln m )}{n_1} }= \mu(\delta) \frac{\sqrt{10 (\Dw^2 G^2 + \ln m ) n_1}}{n_i} , \ \forall i \in [m]
 \]
 where $p_{\varphi}^*$ is the optimal value of (\ref{eqn:convex:concave:weight}) and $\mu(\delta)= 8+2\ln \frac{2}{\delta}$.
\end{theorem}
\begin{myremark}
\textnormal{While the value of $p_{\varphi}^*$ is generally unknown, it can be regard as a small constant when there exists one model that attains small risks on all distributions. We see that Algorithm~\ref{alg:3} exhibits a \emph{distribution-dependent} convergence behavior: the larger the number of samples $n_i$, the smaller the target risk $n_1 p_{\varphi}^*/n_i$, and the faster the convergence rate $O(\sqrt{n_1 \log m}/n_i)$. Note that its rate is always better than the $O(\sqrt{n_1\log m}/n_m)$ rate of SMD with (\ref{eqn:stoch:grad:4}) as gradients. Furthermore, it converges faster than the Baseline when $n_i \geq \sqrt{n_1 n_m}$. In particular, for distribution $\P_1$, Algorithm~\ref{alg:3} attains an $O(\sqrt{(\log m)/n_1})$ rate, which almost matches the optimal $O(\sqrt{1/n_1})$ rate of learning from a single distribution. Finally, we would like to emphasize that a similar idea of introducing ``scale factors'' has been used by \citet[\S~4.3.1]{Nemirovski:SMP} for stochastic semidefinite feasibility problems and \citet{Regret:RML:DS} for empirical MRO.}
\end{myremark}

\subsection{Stochastic Mirror-Prox Algorithm with Mini-batches}
In Algorithm~\ref{alg:3}, distributions with more samples take their advantage by appearing more frequently in the stochastic gradients. In this section, we propose a different way, which lets them reduce the variance in the elements of stochastic gradients by mini-batches \citep{NIPS2007_Topmoumoute}.

The basic idea is as follows. We run our algorithm for a small number of iterations $\nb$ that is no larger than $n_m$. Then, in each iteration, we draw a mini-batch of $n_i/\nb$ samples from every distribution $\P_i$. For $\P_i$ with more samples, we can estimate the associated risk $R_i(\cdot)$ and its gradient more accurately, i.e., with a smaller variance. However, to make this idea work, we need to tackle two obstacles: (i) the performance of the SA algorithm should depend on the variance of gradients instead of the norm, and for this reason SMD is unsuitable; (ii) even some elements of the stochastic gradient have small variances, the entire gradient may still have a large variance. To address the first challenge, we resort to a more advanced SA approach---stochastic mirror-prox algorithm (SMPA), whose convergence rate depends on the variance \citep{Nemirovski:SMP}. To overcome the second challenge, we again introduce scale factors into the optimization problem and the stochastic gradients. And in this way, we can ensure faster convergence rates for distributions with more samples.

In SMPA, we need to maintain two sets of solutions: $(\w_t, \q_t)$ and $(\w_t', \q_t')$. In each round $t$, we first draw $n_i/n_m$ samples from every distribution $\P_i$, denoted by $\z_t^{(i,j)}$, $j=1,\ldots,n_i/n_m$. Then, we use them to construct stochastic gradients at $(\w_t', \q_t')$ of a  weighted GDRO problem (\ref{eqn:convex:concave:weight}), where the value of $p_i$ will be determined later. Specifically, we define
\begin{equation}\label{eqn:stoch:grad:6}
\begin{split}
\g_{w}(\w_t',\q_t') =& \sum_{i=1}^m q_{t,i}' p_i \left(\frac{n_m}{n_i} \sum_{j=1}^{n_i/n_m} \nabla \ell(\w_{t}';\z_t^{(i,j)}) \right), \\
\g_{q}(\w_t',\q_t')= &\left[p_1 \frac{n_m}{n_1}  \sum_{j=1}^{n_1/n_m} \ell(\w_{t}';\z_t^{(1,j)}), p_2 \frac{n_m}{n_2} \sum_{j=1}^{n_2/n_m} \ell(\w_{t}';\z_t^{(2,j)}), \ldots,  p_m \ell(\w_{t}';\z_t^{(m)}) \right]^\top . \!
\end{split}
\end{equation}
Let's take the stochastic gradient $\g_{q}(\w_t',\q_t')$, whose variance will be measured in terms of $\|\cdot\|_\infty$, as an example to explain the intuition of inserting $p_i$. Define $u_i=\frac{n_m}{n_i} \sum_{j=1}^{n_i/n_m}  \ell(\w_{t}';\z_t^{(i,j)})$.  With a larger mini-batch size $n_i/n_m$, $u_i$ will approximate $R_i(\w_{t}')$ more accurately, and thus have a smaller variance. Then, it allows us to insert a larger value of $p_i$, without affecting the variance of $\|\g_{q}(\w_t',\q_t')\|_\infty$, since $\|\cdot\|_\infty$ is insensitive to perturbations of small elements. Similar to the case in Theorem~\ref{thm:5}, the convergence rate of $R_i(\cdot)$ depends on $1/p_i$, and becomes faster if $p_i$ is larger.

Based on (\ref{eqn:stoch:grad:6}), we use SMD to update $(\w_t', \q_t')$, and denote the solution by $(\w_{t+1}, \q_{t+1})$:
\begin{align}
\w_{t+1}= &\argmin_{\w \in \W} \big\{ \eta_w \langle \g_{w}(\w_t',\q_t') , \w -\w_t'\rangle + B_w(\w,\w_t') \big\}, \label{eqn:upate:wt:mirror}  \\
\q_{t+1}=& \argmin_{\q \in \Delta_m} \big\{ \eta_q \langle -\g_{q}(\w_t',\q_t') , \q -\q_t'\rangle + B_q(\q,\q_t') \big\}. \label{eqn:upate:qt:mirror}
\end{align}
Next, we draw another $n_i/n_m$ samples from each distribution $\P_i$, denoted by $\zh_t^{(i,j)}$, $j=1,\ldots,n_i/n_m$,  to construct stochastic gradients at $(\w_{t+1}, \q_{t+1})$:
\begin{equation}\label{eqn:stoch:grad:7}
\begin{split}
\g_{w}(\w_{t+1},\q_{t+1}) =& \sum_{i=1}^m q_{t+1,i} p_i \left(\frac{n_m}{n_i} \sum_{j=1}^{n_i/n_m} \nabla \ell(\w_{t+1};\zh_t^{(i,j)}) \right), \\
\g_{q}(\w_{t+1},\q_{t+1})= &\left[p_1 \frac{n_m}{n_1}  \sum_{j=1}^{n_1/n_m} \ell(\w_{t+1};\zh_t^{(1,j)}),  \ldots,  p_m \ell(\w_{t+1};\zh_t^{(m)}) \right]^\top .
\end{split}
\end{equation}
Then, we use them  to update $(\w_t', \q_t')$ again, and denote the result by $(\w_{t+1}', \q_{t+1}')$:
\begin{align}
\w_{t+1}'= &\argmin_{\w \in \W} \big\{ \eta_w \langle \g_{w}(\w_{t+1},\q_{t+1}) , \w -\w_t'\rangle + B_w(\w,\w_t') \big\} ,\label{eqn:upate:wt:mirror:2}  \\
\q_{t+1}'=& \argmin_{\q \in \Delta_m} \big\{ \eta_q \langle -\g_{q}(\w_{t+1},\q_{t+1}) , \q -\q_t'\rangle + B_q(\q,\q_t') \big\}. \label{eqn:upate:qt:mirror:2}
\end{align}
To meet the budget constraints, we repeat the above process for $n_m/2$ iterations. Finally, we return $\wb=\frac{2}{n_m} \sum_{t=2}^{1+n_m/2} \w_t$  and $\qb=\frac{2}{n_m} \sum_{t=2}^{1+n_m/2} \q_t$ as solutions. The complete procedure is summarized in Algorithm~\ref{alg:4}.

\begin{algorithm}[t]
\caption{Stochastic Mirror-Prox Algorithm for Weighted GDRO}
{\bf Input}: step size $\eta_w$ and $\eta_q$
\begin{algorithmic}[1]
\STATE Initialize $\w_1'=\argmin_{\w \in \W}\nu_w(\w)$, and $\q_1'=[1/m, \ldots, 1/m]^\top \in \R^m$
\FOR{$t=1$ to $n_m/2$}
\STATE For each $i\in[m]$, draw $n_i/n_m$ samples $\{\z_t^{(i,j)}| j=1,\ldots,n_i/n_m\}$ from distribution $\P_i$

\STATE Construct the stochastic gradients defined in (\ref{eqn:stoch:grad:6})
\STATE Calculate $\w_{t+1}$ and $\q_{t+1}$ according to (\ref{eqn:upate:wt:mirror}) and (\ref{eqn:upate:qt:mirror}), respectively
\STATE For each $i\in[m]$, draw $n_i/n_m$ samples $\{\zh_t^{(i,j)}| j=1,\ldots,n_i/n_m\}$ from distribution $\P_i$

\STATE Construct the stochastic gradients defined in (\ref{eqn:stoch:grad:7})
\STATE Calculate $\w_{t+1}'$ and $\q_{t+1}'$ according to (\ref{eqn:upate:wt:mirror:2}) and (\ref{eqn:upate:qt:mirror:2}), respectively
\ENDFOR
\RETURN $\wb=\frac{2}{n_m} \sum_{t=2}^{1+n_m/2} \w_t$  and $\qb=\frac{2}{n_m} \sum_{t=2}^{1+n_m/2} \q_t$
\end{algorithmic}\label{alg:4}
\end{algorithm}

To analysis the performance of Algorithm~\ref{alg:4}, we further assume the risk function $R_i(\cdot)$ is smooth, and the dual norm $\|\cdot\|_{w,*}$ satisfies a regularity condition.
\begin{ass}\label{ass:4} All the risk functions are $L$-smooth, i.e.,
\begin{equation}\label{eqn:smooth:R}
\|\nabla R_i(\w) - \nabla R_i(\w')\|_{w,*} \leq  L \|\w-\w'\|_w, \ \forall  \w,  \w' \in \W, i \in [m].
\end{equation}
\end{ass}
Note that even in the studies of stochastic convex optimization (SCO), smoothness is necessary to obtain a variance-based convergence rate \citep{Lan:SCO}.
\begin{ass} \label{ass:5} The dual norm $\|\cdot\|_{w,*}$ is $\kappa$-regular for some small constant $\kappa \geq 1$.
\end{ass}
The regularity condition is used when analyzing the effect of mini-batches on stochastic gradients. For a formal definition, please refer to \citet{Vector:Martingales}. Assumption~\ref{ass:5} is satisfied by most of papular norms considered in the literature, such as the vector $\ell_p$-norm and the infinity norm.

Then, we have the following theorem for Algorithm~\ref{alg:4}.
\begin{theorem} \label{thm:7}  Define
\begin{equation} \label{eqn:mirror:parameters}
\begin{split}
&p_{\max}=\max_{i\in[m]} p_i, \quad \omega_{\max}=  \max_{i \in [m]}  \frac{p_i^2 n_m}{n_i} ,\\
&\Lt= 2\sqrt{2} p_{\max} (\Dw^2 L +  \Dw^2  G \sqrt{\ln m}), \textrm{ and } \sigma^2=2 c \omega_{\max}(\kappa \Dw^2  G^2 +   \ln^2 m)
\end{split}
\end{equation}
where $c>0$ is an absolute constant. Under Assumptions~\ref{ass:1}, \ref{ass:2a}, \ref{ass:2}, \ref{ass:3}, \ref{ass:4} and \ref{ass:5}, and setting
\[
\eta_w =2  \Dw^2 \min \left(  \frac{1}{\sqrt{3} \Lt}, \frac{2}{\sqrt{7 \sigma^2 n_m}}  \right), \textrm{ and } \eta_q = 2   \min \left( \frac{1}{\sqrt{3} \Lt}, \frac{2}{\sqrt{7 \sigma^2 n_m}}  \right) \ln m
\]
in Algorithm~\ref{alg:4}, with probability at least  $1-\delta$, we have
\[
 R_i(\wb)  - \frac{1}{p_i} p_{\varphi}^*  =  \frac{1}{p_i} \left( \frac{7 \Lt}{n_m} + \sqrt{\frac{\sigma^2}{n_m}} \left( 14\sqrt{\frac{2}{ 3}} + 7\sqrt{3\log \frac{2}{\delta}} + \frac{14}{  n_m} \log \frac{2}{\delta} \right) \right)
\]
where $p_{\varphi}^*$ is the optimal value of (\ref{eqn:convex:concave:weight}).\\
Furthermore, by setting $p_i$ as
\begin{equation} \label{eqn:pi:mirror}
p_i = \frac{1/\sqrt{n_m} + 1}{1/\sqrt{n_m} + \sqrt{n_m/n_i}},
\end{equation}
with high probability, we have
\[
 R_i(\wb) - \frac{1}{p_i} p_{\varphi}^*  =  O\left(\left( \frac{1}{n_m} + \frac{1}{\sqrt{n_i}} \right) \sqrt{\kappa+\ln^2 m}\right).
\]
\end{theorem}
\begin{myremark}
\textnormal{
Compared with Algorithm~\ref{alg:3}, Algorithm~\ref{alg:4} has two advantages: (i) the budget constraint is satisfied exactly;  (ii) we obtain a faster $O((\log m)/\sqrt{n_i})$ rate for all distributions $\P_i$ such that $n_i \leq n_m^2$, which is much better than the $O(\sqrt{n_1 \ln m}/n_i)$ rate of Algorithm~\ref{alg:3}, and the $O(\sqrt{(\log m) /n_m})$ rate of the Baseline. For distributions with a larger budget, i.e., $n_i > n_m^2$, it maintains a fast $O((\log m) /n_m)$ rate. Since it only updates $n_m$ times, and the best  we can expect is the $O(1 /n_m)$ rate of deterministic settings \citep{nemirovski-2005-prox}. So, there is a performance limit for mini-batch based methods, and after that increasing the batch-size cannot reduce the rate, which consists with the usage of mini-batches in SCO \citep{NIPS2011_0942,ICML:13:Zhang:logT}.
}
\end{myremark}

\begin{myremark}
\label{remark:best-of-both-world}
\textnormal{
To further improve the convergence rate, we can design a hybrid algorithm that combines  non-uniform sampling and mini-batches. Specifically, we run our algorithm for $\nb \in [n_m, n_1]$ rounds, and for distributions with $n_i \geq \nb$, we use mini-batches to reduce the variance, and for distributions with $n_i < \nb$, we use random sampling to satisfy the budget constraint. }
\end{myremark}
% Last, we note that there is an additional $\sqrt{\ln m}$ term in Theorem~\ref{thm:7}, compared with other theorems. That is because we use tools from sub-Gaussian distributions (instead of a bounded condition) to analyze the variance of the stochastic gradients, and the presence of the infinity norm contributes a $\sqrt{\ln m}$ term, which seems unavoidable \citep[Lemma A.12]{bianchi-2006-prediction}.

\section{AT$_k$RO for Heterogeneous Distributions}
GDRO is effective in dealing with homogeneous distributions, where the risks of all distributions are roughly of the same order. However, its effectiveness diminishes when confronted with heterogeneous distributions. This stems from the sensitivity of the \emph{max} operator to outlier distributions with significantly high risks, causing it to focus solely on outliers and overlook others \citep{pmlr-v48-shalev-shwartzb16}. To address this issue, research in robust supervised learning has introduced the approach of minimizing  the average of the $k$ largest individual losses \citep{ATk_SVM_loss,ADA-CVAR}. Inspired by these studies, we propose to optimize the average top-$k$ risk $\L_{k}(\w)$ in \eqref{eqn:ATkR}, which can mitigate the influence of outliers.

\subsection{Preliminaries}
By replacing $\L_{\text{max}}(\w)$ in \eqref{eqn:Max_risk} with $\L_{k}(\w)$, we obtain the average top-$k$ risk optimization (AT$_k$RO) problem:
\begin{equation} \label{eqn:ATk-objective}
\min_{\w\in\mathcal{W}}\max_{\I\in \B_{m,k}}\left\{\frac{1}{k}\sum_{i\in \I} R_i(\w)\right\}
\end{equation} 
which reduces to GDRO when $k=1$. Before introducing specific optimization algorithms, we present an example to illustrate the difference between GDRO and AT$_k$RO.

\begin{example}  \label{example:GDRO&ATk}
We define the hypothesis space as $\W = [0,1]$ and the Bernoulli distribution as $\operatorname{Ber}(\mu,1)$, which outputs 1 with probability $\mu$ and 0 with probability $1-\mu$. Then, we consider 16 distributions: $\operatorname{Ber}(\mu_i,1)$ where $\mu_i$ is sequentially set to $0.5, 0.86, 0.87, \ldots, 0.99, 1$. The loss function is defined as $\ell(\w; \z) = (\w - \z)^2$ for a random sample $\z \in \{0, 1\}$ drawn from these distributions.  We denote the solutions of GDRO and AT$_5$RO by $\w^*_G$ and $\w^*_A$, respectively. It is easy to show that $\w^*_G = 0.5$ and $\w^*_A = 0.8$, as detailed in Appendix~\ref{sec:prof:example}.
\end{example}
\begin{figure}
	\begin{center}
		\subfigure[The individual risk $R_i(\w)$ for 9 out of 16 distributions]{
			\label{fig:Example_single} %% label for second subfigure
			\includegraphics[width=0.51\textwidth]{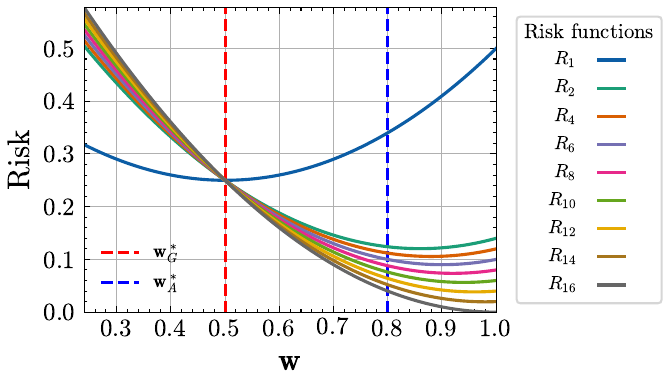}}%
		\
		\subfigure[$\L_{\text{max}}(\w)$ and $\L_{5}(\w)$ ]{
			\label{fig:Example_objective} %% label for second subfigure
			\includegraphics[width=0.395\textwidth]{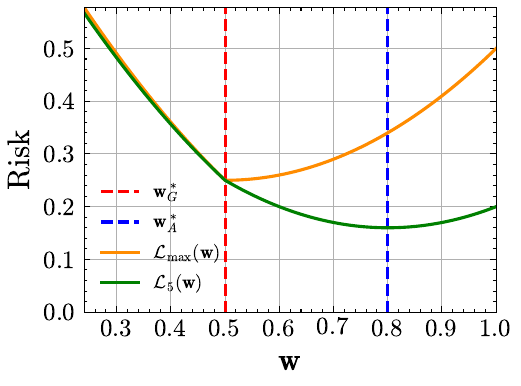}}%
		\caption{Graphical illustrations of Example \ref{example:GDRO&ATk}.}
		\label{fig:Example}
	\end{center}
\end{figure}

To visualize the results, in Fig.~\ref{fig:Example} we plot a portion of the risk functions, the objectives of GDRO and AT$_5$RO, as well as their respective solutions. From Fig.~\ref{fig:Example_single}, it is evident that distribution $\P_1$ is significantly different from the other 15 distributions, indicating it could be an outlier. Fig.~\ref{fig:Example_objective} demonstrates that GDRO primarily focuses on $\P_1$, yielding the solution $\w_G^*=\argmin_{\w \in \W}R_1(\w)=0.5$. Although its solution performs well on $\P_1$, it underperforms on the other 15 distributions. Note that a slight increase in $\w_G^*$ leads to a  noticeable  reduction in  $R_2,\ldots, R_{16}$, with the cost of a minor increase in $R_1$. AT$_5$RO offers a relatively balanced solution $\w_A^*=0.8$  by considering the top-$5$ high-risk distributions. Specifically, the average risk of $\w_A^*$ on distributions $\P_2,\ldots,\P_{16}$ is 0.168 lower than that of $\w_G^*$, with a 0.09 increase in the risk on $\P_1$. Therefore, AT$_5$RO effectively mitigates the influence of the outlier distribution $\P_1$, showing superior robustness compared to GDRO.

Similar to the case of GDRO, \eqref{eqn:ATk-objective} can  be cast as a stochastic convex-concave saddle-point problem:
\begin{equation}	\label{eqn:ATk-convex-concave}
\min_{\w\in\mathcal{W}}\max_{\q\in \SS_{m,k}}\left\{\phi(\w,\q)=\sum_{i=1}^{m} q_i R_i(\w)\right\}
\end{equation}
where 
\[
\SS_{m,k}=\left\{\q\in\mathbb{R}^m \mid 0\leq q_i\leq\frac1k,\sum_{i=1}^m q_i=1\right\}
\]
 is the capped simplex which can be viewed as the slice of the hyper-cube $[0,1/k]^m$ cut by a hyper-plane $\q^{\top}\mathbf{1}=1$. The difference between \eqref{eqn:convex:concave} and \eqref{eqn:ATk-convex-concave} lies in the domain of $\q$, which is $\Delta_m$ and $\SS_{m,k}$ respectively.

Note that a similar convex-concave optimization problem has been studied by \citet{ADA-CVAR} and \citet{Online:Bandit:Minimax}. However, their works investigate the deterministic setting, whereas our paper considers a stochastic problem. Consequently, their algorithms are not applicable here, necessitating the design of efficient stochastic approaches for \eqref{eqn:ATk-convex-concave}. By replacing $\Delta_m$ in \eqref{eqn:per:measure} with $\SS_{m,k}$, we obtain the performance measure of an approximate solution $(\wb, \qb)$ to \eqref{eqn:ATk-convex-concave}, i.e.,
\begin{equation} \label{eqn:AT_k:measure}
\epsilon_\phi'(\bar{\w},\bar{\q})=\max_{\q\in\SS_{m,k}}\phi(\bar{\w},\q)-\min_{\w\in\mathcal{W}}\phi(\w,\bar{\q})
\end{equation}
which also controls the optimality of $\bar{\w}$ to \eqref{eqn:ATk-objective} by replacing $\Delta_m$ with $\SS_{m,k}$ in \eqref{eqn:relation:error}.

\subsection{Stochastic Mirror Descent for AT$_k$RO}
Following the procedure in Section~\ref{sec:smd:1},  we also use SMD to optimize \eqref{eqn:ATk-convex-concave}, with the only difference being the update rule for $\q$. 

Since the objectives of \eqref{eqn:ATk-convex-concave} and \eqref{eqn:convex:concave} are identical,  the stochastic gradients $\g_{w}(\w_t,\q_t)$ and $\g_{q}(\w_t,\q_t)$ in \eqref{eqn:stoch:grad:1} also serve as unbiased estimators of true gradients $\nabla_\w \phi(\w_t,\q_t)$ and $\nabla_\q \phi(\w_t,\q_t)$, respectively. In the $t$-th round, we reuse \eqref{eqn:upate:wt} to update $\w_t$, and modify the update of $\q_t$ as
\begin{equation} \label{eqn:upate:qt:ATk:1}
	\q_{t+1}= \argmin_{\q \in \SS_{m,k}} \big\{ \eta_q \langle -\g_{q}(\w_t,\q_t) , \q -\q_t\rangle + B_q(\q,\q_t) \big\}.
\end{equation}
Because the domain is no longer the simplex $\Delta_m$, the explicit form in \eqref{eqn:upate:qt:2} does not apply to \eqref{eqn:upate:qt:ATk:1}.  In the following lemma, we demonstrate that \eqref{eqn:upate:qt:ATk:1} can be reduced to a neg-entropy Bregman projection problem onto the capped simplex \citep{Coaching}.
\begin{lemma}\label{lemma:capped_simplex_opt}
Consider a mirror descent defined as 
\begin{equation} \label{q_OMD_0}
\q= \argmin_{\q \in \SS_{m,k}} \big\{\eta \langle \g, \q -\q_0\rangle + B_q(\q,\q_0) \big\}
\end{equation}
where $\g,\q_0\in\R^m$ and $B_q(\cdot,\cdot)$ is the Bregman distance defined in terms of the neg-entropy.  Then, (\ref{q_OMD_0}) is equivalent to $\q=\argmin_{\q \in \SS_{m,k}} B_q(\q,\hat{\q})$ where $\hat{q}_i=q_{0,i} e^{-\eta g_i}$.
\end{lemma}
By Lemma~\ref{lemma:capped_simplex_opt}, we can leverage existing algorithms for neg-entropy Bregman projections onto the capped simplex to compute 
\begin{equation} \label{eqn:upate:qt:ATk:2}
\q_{t+1}=\argmin_{\q \in \SS_{m,k}} B_q(\q,\hat{\q}_t), \textrm{ where } \hat{q}_{t,i}=q_{t,i} e^{\eta_q \ell(\w_{t};\z_t^{(i)})}.
\end{equation}
In particular, we choose Algorithm 2 of \citet{Coaching}, summarized in Appendix~\ref{appendix:section:Projection}, whose time complexity is $O(m+k\ln k)$. 

\begin{algorithm}[t]
	\caption{Stochastic Mirror Descent for AT$_k$RO}
	{\bf Input}: step size $\eta_w$ and $\eta_q$
	\begin{algorithmic}[1]
		\STATE Initialize $\w_1=\argmin_{\w \in \W}\nu_w(\w)$, and $\q_1=[1/m, \ldots, 1/m]^\top \in \R^m$
		\FOR{$t=1$ to $T$}
		\STATE For each $i\in[m]$, draw a sample $\z_{t}^{(i)}$ from distribution $\P_i$
		\STATE Construct the stochastic gradients defined in \eqref{eqn:stoch:grad:1}
		\STATE Update $\w_t$ and $\q_t$ according to \eqref{eqn:upate:wt} and \eqref{eqn:upate:qt:ATk:2}, respectively
		\ENDFOR
		\RETURN $\wb=\frac{1}{T} \sum_{t=1}^T \w_t$  and $\qb=\frac{1}{T} \sum_{t=1}^T \q_t$
	\end{algorithmic}\label{alg:5}
\end{algorithm}

We present the entire procedure in Algorithm~\ref{alg:5}, and have the following theorem.
\begin{theorem} \label{thm:8} 
Under Assumptions~\ref{ass:1}, \ref{ass:2a}, \ref{ass:2} and \ref{ass:3}, and setting $\eta_w = \Dw^2\sqrt{\frac{8}{5T (\Dw^2 G^2+\ln\frac{m}{k})}}$ and $\eta_q = (\ln \frac{m}{k})\sqrt{\frac{8}{5T (\Dw^2 G^2+\ln\frac{m}{k})}}$ in Algorithm~\ref{alg:5},  we have
	\[
	\E \big[ \epsilon_{\phi}'(\wb, \qb) \big] \leq 2 \sqrt{\frac{10 (\Dw^2 G^2 + \ln\frac{m}{k} )}{T} }
	\]
	and with probability at least $1-\delta$,
	\[
	\epsilon_{\phi}'(\wb, \qb) \leq  \left(8+2\ln \frac{2}{\delta}\right) \sqrt{\frac{10(\Dw^2 G^2 + \ln\frac{m}{k} )}{T} }.
	\]
\end{theorem}
\begin{myremark}
\textnormal{The above theorem indicates that Algorithm~\ref{alg:5} attains an $O(\sqrt{(\log (m/k))/T})$ convergence rate. Since it requires $m$ samples in each iteration, the sample complexity is $O((m \log (m/k))/\epsilon^2)$.}
\end{myremark}

\subsubsection{Anytime Extensions}
As discussed in  Section~\ref{sec:anytime:smd}, we can adapt Algorithm~\ref{alg:5} for anytime use by employing time-varying step sizes.  In the $t$-th round, we use step sizes
\begin{equation} \label{eqn:alg5_anytime_stepsize}
\eta^w_t =  \Dw^2\sqrt{\frac{2}{t(\Dw^2 G^2+\ln\frac{m}{k})}}, \textrm{ and }\eta^q_t = (\ln\frac{m}{k})\sqrt{\frac{2}{t(\Dw^2 G^2+\ln\frac{m}{k})}}
\end{equation}
in  (\ref{eqn:upate:wt}) and  (\ref{eqn:upate:qt:ATk:1})/(\ref{eqn:upate:qt:ATk:2}) to update $\w_t$ and $\q_t$, respectively. When required, we return $\wb_t$ and $\qb_t$ in \eqref{eqn:anytime_output} as outputs. 

Similar to Theorem~\ref{thm:alg1_anytime}, we have the following theoretical guarantee for $(\wb_t, \qb_t)$. 
\begin{theorem} \label{thm:alg5_anytime} 
Under Assumptions~\ref{ass:1}, \ref{ass:2a}, \ref{ass:2} and \ref{ass:3}, and setting step sizes as \eqref{eqn:alg5_anytime_stepsize} in Algorithm~\ref{alg:5}, we have
	\[
	\E \big[ \epsilon_{\phi}'(\wb_t, \qb_t) \big] \leq \frac{\sqrt{\Dw^2 G^2+\ln \frac{m}{k}}}{\sqrt{2}\left(\sqrt{t+1}-1\right)} \left(5+3\ln t\right)=O\left(\frac{\sqrt{\log \frac{m}{k}} \log t}{\sqrt{t}}\right) , \ \forall  t \in \zn.
	\]
Furthermore, with probability at least $1-\delta$,  we have
	\[
	\epsilon_{\phi}'(\wb_t, \qb_t)\leq\frac{\sqrt{\Dw^2 G^2+ \ln \frac{m}{k}}}{\sqrt{2}\left(\sqrt{t+1}-1\right)}
	\left(9+11\ln\frac{2}{\delta}+7\ln t+3\ln\frac{2}{\delta}\ln t\right)
	=O\left(\frac{\sqrt{\log \frac{m}{k}} \log t}{\sqrt{t}}\right)
	\]
for each $t \in \zn$.
\end{theorem}
\begin{myremark}
\textnormal{Similar to previous cases, the convergence rate of the anytime extension is slower by a factor of $O(\log t)$.}
\end{myremark}
\subsection{Non-oblivious Online Learning for AT$_k$RO} \label{sec:online:ATK}
Building on the two-player game in Section~\ref{sec:smd:2}, we can leverage online learning techniques to reduce the number of samples used in each round from $m$ to $k$.

The 1st player  faces the same problem, specifically minimizing the sequence of convex functions in (\ref{eqn:first:player}) under the constraint $\w \in \W$. Therefore, it can still be framed as ``non-oblivious OCO with stochastic gradients'' and solved using SMD. In contrast, the 2nd player tackles a different challenge: maximizing the sequence of linear functions in (\ref{eqn:second:player}), constrained by $\q \in \SS_{m,k}$ rather than $\q \in \Delta_m$.  Because the domain is the capped simplex, it is natural to ask the 2nd player  to select the $k$ highest-risk options from $m$ distributions, reflecting the combinatorial nature of the problem. After drawing one sample from each selected distribution, the 2nd player  observes $k$ stochastic rewards,  which fits into a semi-bandit structure. This leads to modeling the 2nd player's problem as ``non-oblivious combinatorial semi-bandits with stochastic rewards''.  For the 2nd player, we  can certainly apply existing algorithms designed for non-oblivious combinatorial semi-bandits \citep{doi:10.1287/moor.2013.0598,JMLR:v17:15-091,Exp4.MP}. Here, to maintain consistency with Algorithm~\ref{alg:2}, we will extend the Exp3-IX algorithm to address this scenario.

In the following, we elaborate on the details and modifications compared to Algorithm~\ref{alg:2}. To select $k$ distributions from $m$ in each round, we require a sampling algorithm that, given the value of $k$ and a probability vector $\p\in\SS_{m,k}$, can generate a set $\I$ such that
\begin{equation}\label{DepRound:prop}
	|\I|=k, \textrm{ and } \Pr[i \in \I] = k p_i,  \ \forall i \in [m].
\end{equation}
For this purpose, we can use the DepRound algorithm \citep{DepRound}, which satisfies the above requirement and has $O(m)$ time and space complexities. A detailed description of its procedure is provided  in  Appendix~\ref{appendix:section:DepRound}.  We note that DepRound has been used by many combinatorial semi-bandit algorithms \citep{DepRound-app1,Exp4.MP,Online:Bandit:Minimax}.  In each round $t$,  we first invoke the DepRound algorithm with $(k,\q_t)$ as inputs to generate a set $\I_t$ containing the indices of $k$ selected distributions. For each $i \in \I_t$, we then draw a sample $\z_t^{(i)}$ from the corresponding distribution $\P_i$.

Next, the  1st player constructs the stochastic gradient as shown below:
\begin{equation} \label{eqn:ATk:grad:w}
	\tilde{\g}_w(\w_t,\q_t)=\frac1k\sum_{i\in \I_t}\nabla\ell(\w_t;\z_t^{(i)})
\end{equation}
which can be easily verified, based on \eqref{DepRound:prop}, as an unbiased estimator of  $\nabla_\w\phi(\w_t,\q_t)$. Then, we update $\w_{t}$ by applying the mirror descent \eqref{eqn:upate:wt:2} with  $\tilde{\g}_w(\w_t,\q_t)$ in (\ref{eqn:ATk:grad:w}). For the 2nd player, we modify the IX loss estimator for the combinatorial semi-bandit setting: 
\begin{equation}\label{eq:IX-loss-estimator-2}
\tilde{s}_{t,i}=\frac{1-\ell(\w_t,\z_t^{(i)})}{kq_{t,i}+\gamma}\cdot\ind[i\in \I_t], \ \forall i\in[m]
\end{equation}
and then update $\q_t$ by mirror descent
\begin{equation} \label{eqn:upate:qt:ATk:3}
	\q_{t+1}=\argmin_{\q \in \SS_{m,k}}\left\{\eta_q\langle\tilde{\s}_{t},\q-\q_t\rangle+B_q(\q,\q_t)\right\}.
\end{equation} 
Compared with \eqref{eq:IX-loss-estimator}, \eqref{eq:IX-loss-estimator-2}  incorporates two key changes. First, we replace $\ind[i_t=i]$ with $\ind[i\in \I_t]$ to utilize all the $k$ observed losses $\{\ell(\w_t,\z_t^{(i)})| i \in \I_t\}$. Second, since $\Pr[i\in \I_t]=kq_{t,i}$, the denominator of $\tilde{s}_{t,i}$ is adjusted accordingly. By Lemma~\ref{lemma:capped_simplex_opt}, we can similarly transform \eqref{eqn:upate:qt:ATk:3} into a neg-entropy Bregman projection problem:
\begin{equation} \label{eqn:upate:qt:ATk:4}
\q_{t+1}=\argmin_{\q \in \SS_{m,k}} B_q(\q,\hat{\q}_t), \textrm{ where } \hat{q}_{t,i}=q_{t,i} e^{-\eta_q \tilde{s}_{t,i}}
\end{equation} 
which can be solved by Algorithm 2 of \citet{Coaching}. The complete procedure is presented in Algorithm~\ref{alg:6}.

\begin{algorithm}[t]
	\caption{Non-oblivious Online Learning for AT$_k$RO}
	{\bf Input}: step sizes $\eta_w$ and $\eta_{q}$, and IX coefficient $\gamma$ 
	\label{alg:6}
	\begin{algorithmic}[1]
		\STATE  Initialize $\w_1=\argmin_{\w \in \W}\nu_w(\w)$, and $\q_1=[1/m, \ldots, 1/m]^\top \in \R^m$
		\FOR{$t=1$ to $T$}
		\STATE Generate $\I_t=\text{\rm DepRound}(k,\q_t)$
		\STATE For each $i\in \I_t$, draw a sample $\z_t^{(i)}$ from distribution $\mathcal{P}_i$
		\STATE Construct the stochastic gradient in \eqref{eqn:ATk:grad:w} and the modified IX loss estimator in \eqref{eq:IX-loss-estimator-2}
		\STATE Update $\w_t$ and $\q_t$ according to \eqref{eqn:upate:wt:2} and \eqref{eqn:upate:qt:ATk:4}, respectively
		\ENDFOR
		\RETURN $\wb=\frac1T\sum_{t=1}^T\w_t$ and $\qb=\frac1T\sum_{t=1}^T\q_t$
	\end{algorithmic}
\end{algorithm}

For the 1st player, Theorem~\ref{thm:2} remains applicable because the only change in the proof of Theorem~\ref{thm:2} is that $\|\tilde{\g}_w(\w_t,\q_t)\|_{w,*}^2=\|\frac1k\sum_{i\in \I_t}\nabla\ell(\w_t;\z_t^{(i)})\|_{w,*}^2\leq G^2$, which does not alter the conclusion. For the 2nd player,  we prove the following theorem.

\begin{theorem} \label{thm:9}
	Under Assumption~\ref{ass:2}, and setting $\eta_{q}=\sqrt{\frac{k \ln m}{m T}}$ and $\gamma=\frac{\eta_{q}}{2}$,  we have
\[
\E \left[\max_{\q\in\SS_{m,k}}\sum_{t=1}^T\phi(\w_t,\q)-\sum_{t=1}^T\phi(\w_t,\q_t)\right]
		\leq3\sqrt{\frac{T}{2}}+\frac{2m}{k}+2\sqrt{\frac{mT}{k\ln m}}+3\sqrt{\frac{m T \ln m}{k}}+\frac{m \ln m}{k}
\]
	and with probability at least $1-\delta$,
\[
\begin{split}
& \max_{\q\in\SS_{m,k}}\sum_{t=1}^T\phi(\w_t,\q)-\sum_{t=1}^T\phi(\w_t,\q_t)	\\
\leq & \sqrt{\frac{T}{2}}+\left(\sqrt{\frac{T}{2}}+\frac{m}{k}+\sqrt{\frac{mT}{k\ln m}}\right)\ln\frac{2}{\delta}+3\sqrt{\frac{mT \ln m}{k}}+\frac{m \ln m}{k}.
\end{split}
\]
\end{theorem}
Based on Theorems~\ref{thm:2} and~\ref{thm:9}, we directly obtain the optimization error of Algorithm~\ref{alg:6} as follows.
\begin{theorem} \label{thm:10}
Under Assumptions~\ref{ass:1}, \ref{ass:2a}, \ref{ass:2} and \ref{ass:3}, and setting $\eta_w=\frac{2D}{G\sqrt{5T}}$, $\eta_{q}=\sqrt{\frac{k\ln m}{m T}}$ and $\gamma=\frac{\eta_{q}}{2}$ in Algorithm~\ref{alg:6}, we have
\[
\mathrm{E}\left[\epsilon_\phi'(\bar{\w}, \bar{\q})\right] \leq 2 D G \sqrt{\frac{5}{T}}+3\sqrt{\frac{1}{2T}}+2\sqrt{\frac{m}{kT\ln m}}+3\sqrt{\frac{m\ln m}{kT}}+\frac{m(2+\ln m)}{kT}
\]
	and with probability at least $1-\delta$,
\[
		\begin{split}
		\epsilon_\phi'(\bar{\w}, \bar{\q})
		\leq & DG\sqrt{\frac{1}{T}}\left(2\sqrt{5}+8\sqrt{\ln\frac2\delta}\right)+\sqrt{\frac{1}{2T}}+\left(\sqrt{\frac{1}{2T}}+\frac{m}{kT}+\sqrt{\frac{m}{kT\ln m}}\right)\ln\frac{4}{\delta}\\
		&+3\sqrt{\frac{m\ln m}{kT}}+\frac{m \ln m}{kT}.
		\end{split}
\]
\end{theorem}

\begin{myremark}
\textnormal{Theorem~\ref{thm:10} demonstrates that Algorithm~\ref{alg:6} obtains an $O(\sqrt{m(\log m)/(kT)})$ convergence rate. Since it consumes $k$ samples per iteration, the sample complexity is $O(m (\log m)/\epsilon^2)$, slightly higher than that of Algorithm~\ref{alg:5}.}
\end{myremark}

\subsubsection{Anytime Extensions} 
Based on the discussion in Section~\ref{sec:online:anytime}, it is natural to adopt time-varying parameters to make Algorithm~\ref{alg:6} anytime. However, during the theoretical analysis, we encountered a technical obstacle. In the original paper of Exp3-IX, there are two concentration results concerning the IX loss estimator (\ref{eq:IX-loss-estimator}): one for fixed parameters and the other for time-varying parameters, i.e., Corollary 1 and Lemma 1 of \citet{NIPS2015_e5a4d6bf} respectively. In Section~\ref{sec:online:ATK}, we successfully extended their Corollary 1 to combinatorial semi-bandits, resulting in Theorem~\ref{thm:9}. However, we are unable to extend their Lemma 1 to combinatorial semi-bandits,\footnote{In combinatorial semi-bandits,  there are $k$ non-zero $\{\tilde{s}_{t,i} | i \in \I_t\}$ in each round $t$. Consequently, we need to handle $k$ non-zero $\{\bar{\xi}_{t,i}| i \in \I_t\}$  in \eqref{eq:concentration-2}, which renders the original analysis invalid, and it remains unclear how to resolve this issue.} and therefore cannot provide theoretical guarantees for Algorithm~\ref{alg:6} when using time-varying parameters. Additionally, we have not found any algorithms in the literature that utilize time-varying parameters in non-oblivious combinatorial semi-bandits.

To circumvent the aforementioned challenge, we present an anytime algorithm for AT$_k$RO from a different perspective. The key observation is that 
 \emph{we are not dealing with a true bandit problem but are instead exploiting bandit techniques to solve} \eqref{eqn:ATk-convex-concave}. 
During the execution of our algorithm, the 2nd player is not necessarily required to select $k$ distinct arms. It is perfectly fine to select just $1$ arm, as long as we can bound the regret in terms of the linear functions in (\ref{eqn:second:player}), subject to the constraint $\q \in \SS_{m,k}$. To this end, we propose to modify the anytime extension of Algorithm~\ref{alg:2} described in Section~\ref{sec:online:anytime}.  

In the following, we describe the key steps. Recall the three time-varying parameters $\eta^w_t$, $\eta^q_t$ and $\gamma_{t}$ in (\ref{eqn:alg2_anytime_stepsize}).
In each round, we use SMD in (\ref{eqn:upate:wt:2}) with a time-varying step size to update $\w_t$:
\begin{equation} \label{eqn:upate:wt:3}
\w_{t+1}= \argmin_{\w \in \W} \big\{\eta^w_t \langle \gt_{w}(\w_t,\q_t) , \w -\w_t\rangle + B_w(\w,\w_t) \big\}
\end{equation}
where $\gt_{w}(\w_t,\q_t)$ is defined in (\ref{eqn:stoch:grad:2}). Similarly, we use a  time-varying parameter to define the IX loss estimator 
\begin{equation} \label{eq:IX-loss-estimator:ATK}
    \tildes_{t,i} = \frac{1 - \ell(\w_t,\z_t^{(i_t)})}{q_{t,i} + \gamma_t} \cdot \ind[i_t = i], \ \forall i \in [m].
\end{equation}
 The only change required is to adjust the domain in the mirror descent (\ref{eq:algo-q}) to $\SS_{m,k}$:
\begin{equation}    \label{eq:algo-q:ATK1}
    \q_{t+1} = \argmin_{\q \in \SS_{m,k}} \big \{\eta^q_t \langle\tildesb_t, \q - \q_t\rangle + B_{q}(\q,\q_t)\big\}
\end{equation}
which can also be reduced to a neg-entropy Bregman projection problem. If demanded, we will return $(\wb_t, \qb_t)$ in (\ref{eqn:anytime_output}) as the current solution. We summarize the complete procedure in Algorithm~\ref{alg:7}.

\begin{algorithm}[t]
\caption{Non-oblivious Online Learning for AT$_k$RO with Anytime Capability}
\begin{algorithmic}[1]
\STATE Initialize $\w_1=\argmin_{\w \in \W}\nu_w(\w)$, and $\q_1=[1/m, \ldots, 1/m]^\top \in \R^m$
\FOR{$t=1$ to $T$}
\item Generate $i_t \in [m]$ according to $\q_t$, and draw a sample $\z_t^{(i_t)}$ from distribution $\P_{i_t}$
\STATE Construct the stochastic gradient in (\ref{eqn:stoch:grad:2}) and the IX loss estimator in (\ref{eq:IX-loss-estimator:ATK})
\STATE Update $\w_t$ and $\q_t$ according to (\ref{eqn:upate:wt:3}) and (\ref{eq:algo-q:ATK1}), respectively
\ENDFOR
\end{algorithmic}\label{alg:7}
\end{algorithm}

Following the proof of Theorem~\ref{thm:alg2_anytime}, we establish the following theoretical guarantee regarding the optimization error.
\begin{theorem} \label{thm:alg7} 
Under Assumptions~\ref{ass:1}, \ref{ass:2a}, \ref{ass:2} and \ref{ass:3}, for Algorithm~\ref{alg:7} we have
\[ 
	\begin{split}
	\E \big[ \epsilon_{\phi}'(\wb_t, \qb_t) \big] 
	\leq &\frac{\left(3+\ln t\right)\sqrt{m\ln m}+6\sqrt{m/\ln m}+4\sqrt{(1+\ln t)/2}+DG\left(5+3\ln t\right)}{2\left(\sqrt{t+1}-1\right)}\\
	= & O\left(\frac{\sqrt{m \log m}\log t}{\sqrt{t}}\right),  \quad\quad\quad\quad  \forall  t \in \zn.
\end{split}
\]
Furthermore, with probability at least $1-\delta$,  we have
\[
	\begin{split}
	&\epsilon_{\phi}'(\wb_t, \qb_t) \\
	\leq & \frac{\left(3+\ln t\right)\sqrt{m\ln m}+\left(2\sqrt{\frac{m}{\ln m}}+\sqrt{\frac{1+\ln t}{2}}\right)\ln\frac{6}{\delta}+\sqrt{\frac{1+\ln t}{2}}+DG\left(9+7\ln t+4\ln\frac{2}{\delta}\right)}{2\left(\sqrt{t+1}-1\right)} \\
	= & O\left(\frac{\sqrt{m \log m}\log t}{\sqrt{t}}\right)
	\end{split}
\]
for each $t \in \zn$.
\end{theorem}

\begin{myremark}
\textnormal{Note that the upper bounds in this theorem are exactly the same as  in Theorem~\ref{thm:alg2_anytime}. Since Algorithm~\ref{alg:7} uses only $1$ sample per iteration,  it is not surprising that its convergence rate is slower than Algorithm~\ref{alg:6} by a factor of $\O(\sqrt{k})$.}
\end{myremark}

\section{Analysis}
In this section, we present proofs of main theorems, and defer the analysis of supporting lemmas to Appendix~\ref{appendix:Lemmas}.

\subsection{Proof of Theorem~\ref{thm:1}} \label{sec:prof:1}
The proof is based on Lemma 3.1 and Proposition 3.2 of \citet{nemirovski-2008-robust}. To apply them, we show that their preconditions are satisfied  under our assumptions.

Although two instances of SMD are invoked to update $\w$ and $\q$ separately, they can be merged as $1$ instance by concatenating $\w$ and $\q$ as a single variable $[\w;\q] \in \W \times \Delta_m$, and redefine the norm and the distance-generating function \citep[\S~3.1]{nemirovski-2008-robust}. Let $\mathcal{E}$ be the space that $\W$ lies in. We equip the Cartesian product $\mathcal{E} \times \R^m$ with the following norm and dual norm:
\begin{equation} \label{eqn:norm:new}
\big\|[\w;\q] \big\|= \sqrt{ \frac{1}{2 \Dw^2} \|\w\|_{w}^2 + \frac{1}{2 \ln m} \|\q\|_1^2 } , \textrm{ and } \big\|[\u;\v] \big\|_*= \sqrt{2 \Dw^2 \|\u\|_{w,*}^2 +  2 \|\v\|_\infty^2   \ln m}.
\end{equation}
We use the notation $\x=[\w;\q]$, and equip the set $\W \times \Delta_m$  with  the distance-generating function
\begin{equation} \label{eqn:dis:fun}
\nu(\x) =\nu([\w;\q])  = \frac{1}{2 \Dw^2} \nu_w(\w) +\frac{1}{2 \ln m} \nu_q(\q) .
\end{equation}
It is easy to verify that $\nu(\x)$ is $1$-strongly convex w.r.t.~the norm $\|\cdot\|$. Let $B(\cdot,\cdot)$ be the Bregman distance associated with $\nu(\cdot)$:
\begin{equation} \label{eqn:Bregman:merge}
\begin{split}
B(\x,\x')= &\nu(\x) - \big[\nu(\x') + \langle \nabla \nu(\x') , \x-\x'\rangle\big] \\
=& \frac{1}{2 \Dw^2} \left(  \nu_w(\w) - \big[\nu_w(\w') + \langle \nabla \nu_w(\w') , \w-\w'\rangle\big]
 \right)\\
&+\frac{1}{2 \ln m} \left( \nu_q(\q) - \big[\nu_q(\q') + \langle \nabla \nu_q(\q') , \q-\q'\rangle\big]
\right)\\
=& \frac{1}{2 \Dw^2} B_w(\w,\w') + \frac{1}{2 \ln m} B_q(\q,\q')
\end{split}
\end{equation}
where $\x'=[\w';\q']$.

Then, we consider the following version of SMD for updating $\x_t$:
\begin{equation} \label{eqn:update:xt}
\x_{t+1}= \argmin_{\x \in \W \times \Delta_m} \Big\{ \eta \big \langle [ \g_{w}(\w_t,\q_t); -\g_{q}(\w_t,\q_t)] , \x -\x_t \big\rangle + B(\x,\x_t) \Big\}
\end{equation}
where $\eta>0$ is the step size. In the beginning, we set $\x_1=\argmin_{\x \in \W \times \Delta_m}\nu(\x) = [\w_1;\q_1]$. From the decomposition of the Bregman distance  in (\ref{eqn:Bregman:merge}), we observe that (\ref{eqn:update:xt}) is equivalent to (\ref{eqn:upate:wt}) and (\ref{eqn:upate:qt:1}) by setting
\[
\eta_w =2 \eta \Dw^2, \textrm{ and } \eta_q = 2 \eta \ln m .
\]

Next, we show that the stochastic gradients are well-bounded.  Under our assumptions, we have
\[
\begin{split}
& \|\g_{w}(\w_t,\q_t)\|_{w,*}=\left\|\sum_{i=1}^m q_{t,i} \nabla \ell(\w_{t};\z_t^{(i)})\right\|_{w,*} \leq  \sum_{i=1}^m q_{t,i} \left\| \nabla \ell(\w_{t};\z_t^{(i)})\right\|_{w,*} \overset{\text{(\ref{eqn:gradient})}}{\leq} \sum_{i=1}^m q_{t,i}  G = G,\\
&\|\g_{q}(\w_t,\q_t)\|_\infty = \big \| [\ell(\w_{t};\z_t^{(1)}), \ldots, \ell(\w_{t};\z_t^{(m)})]^\top \big \|_\infty \overset{\text{(\ref{eqn:value})}}{\leq} 1.
\end{split}
\]
As a result, the concatenated gradients used in (\ref{eqn:update:xt}) is also bounded in term of the dual norm $\|\cdot \|_*$:
\begin{equation} \label{eqn:alg1_M}
\begin{split}
\big\| [ \g_{w}(\w_t,\q_t); -\g_{q}(\w_t,\q_t)] \big\|_*= &\sqrt{ 2 \Dw^2 \|\g_{w}(\w_t,\q_t)\|_{w,*}^2 +  2 \|\g_{q}(\w_t,\q_t)\|_\infty^2   \ln m} \\
\leq &\underbrace{\sqrt{2 \Dw^2 G^2  +  2 \ln m}}_{:=M}.
\end{split}
\end{equation}

Now, we are ready to state our theoretical guarantees. By setting
\[
\eta=\frac{2}{M \sqrt{5T}}  =  \sqrt{\frac{2}{5T (\Dw^2 G^2+\ln m)}},
\]
 (3.13) of \citet{nemirovski-2008-robust} implies that
\[
\E \big[ \epsilon_{\phi}(\wb, \qb) \big] \leq 2 M \sqrt{\frac{5}{T}}= 2 \sqrt{\frac{10 (\Dw^2 G^2 + \ln m )}{T} }.
\]
Furthermore, from Proposition 3.2 of \citet{nemirovski-2008-robust}, we have, for any $\Omega>1$
\[
\Pr\left[ \epsilon_{\phi}(\wb, \qb) \geq   (8+2\Omega) M \sqrt{\frac{5}{T}}  = (8+2\Omega) \sqrt{\frac{10( \Dw^2 G^2 + \ln m )}{T} }\right] \leq 2 \exp(-\Omega).
\]
We complete the proof by setting $\delta=2 \exp(-\Omega)$.
\subsection{Proof of Theorem~\ref{thm:alg1_anytime}}\label{sec:prof:1_anytime}
In a manner similar to the proof of Theorem~\ref{thm:1} in Section~\ref{sec:prof:1}, we combine the updates for $\w_t$ and $\q_t$ into a unified expression:
\[
\x_{t+1}= \argmin_{\x \in \W \times \Delta_m} \Big\{ \eta_t \big \langle [ \g_{w}(\w_t,\q_t); -\g_{q}(\w_t,\q_t)] , \x -\x_t \big\rangle + B(\x,\x_t) \Big\}
\]
where the step size $\eta_t$ satisfying
\[
\eta^w_t =2 \eta_t \Dw^2, \textrm{ and }  \eta^q_t= 2 \eta_t \ln m.
\]
Then, from (3.11) of \citet{nemirovski-2008-robust}, we have
\begin{equation} \label{eqn:alg1_anytime_1}
\begin{split}
\E \big[ \epsilon_{\phi}(\wb_t, \qb_t) \big] \leq & \left(\sum_{j=1}^t \eta_t \right)^{-1} \left( 2 + \frac{5}{2} M^2 \sum_{j=1}^t \eta_t^2 \right)\\
= & \left(\sum_{j=1}^t\frac{1}{\sqrt{j}}\right)^{-1} \left(2M+\frac{5M}{2}\sum_{j=1}^t\frac{1}{j}\right)
\end{split}
\end{equation}
where we set $\eta_t=\frac{1}{M \sqrt{t}}$, and $M$ is defined in \eqref{eqn:alg1_M}. Combining \eqref{eqn:alg1_anytime_1} with the following inequalities
\begin{equation} \label{eqn:anytime_fact}
	\begin{split}
		& \sum_{j=1}^t \frac{1}{j} \leq 1+\int_1^t \frac{1}{x} d x=1+\left.\ln x\right|_1 ^t=1+\ln t \\
		& \sum_{j=1}^t \frac{1}{\sqrt{j}} \geq \int_1^{t+1} \frac{1}{\sqrt{x}} d x=\left.2 \sqrt{x}\right|_1 ^{t+1}=2(\sqrt{t+1}-1)
	\end{split}
\end{equation}
we obtain
\[
	\E \big[ \epsilon_{\phi}(\wb_t, \qb_t) \big] \leq \frac{M}{2\left(\sqrt{t+1}-1\right)} \left(5+3\ln t\right).
\]

Next, we focus on the high-probability bound. Although Proposition 3.2 of \citet{nemirovski-2008-robust}   provides a high-probability bound only for a fixed step size, its proof actually supports time-varying step sizes. By setting $\Theta=2\sqrt{\Omega}$ in their analysis, we have
\begin{equation}\label{eqn:anytime_high_1}
	\Pr\left[\sum_{j=1}^t \eta_{j} \epsilon_{\phi}(\wb_t, \qb_t)>2+\frac{5}{2}(1+\Omega) M^2 \sum_{j=1}^t \eta_j^2+8 \sqrt{2\Omega} M \sqrt{\sum_{j=1}^t \eta_j^2}\right] \leq 2\exp (-\Omega)
\end{equation}
for any $\Omega>0$.  Substituting $\eta_t=\frac{1}{M \sqrt{t}}$ into \eqref{eqn:anytime_high_1}, we have
\[
\begin{split}
&2\exp (-\Omega)\\
  \geq & \Pr\left[\epsilon_{\phi}(\wb_t, \qb_t)> M \left(\sum_{j=1}^t\frac{1}{\sqrt{j}}\right)^{-1}\left(2+\frac{5}{2}(1+\Omega)\left(\sum_{j=1}^t \frac{1}{\sqrt{j}}\right) +  8\sqrt{2\Omega\left(\sum_{j=1}^t \frac{1}{\sqrt{j}}\right)}\right)\right] \\
   \overset{\text{(\ref{eqn:anytime_fact})}}{\geq} &  \Pr\left[\epsilon_{\phi}(\wb_t, \qb_t)>\frac{M}{2\left(\sqrt{t+1}-1\right)}\left(2+\frac{5}{2}(1+\Omega)(1+\ln t) +  8\sqrt{2\Omega(1+\ln t)}\right)\right] \\
 \geq &\Pr\left[\epsilon_{\phi}(\wb_t, \qb_t)>\frac{M}{2\left(\sqrt{t+1}-1\right)}\left(2+\frac{5}{2}(1+\Omega)(1+\ln t) +  4\left(2\Omega+1+\ln t\right)\right)\right] \\
\geq&\Pr\left[\epsilon_{\phi}(\wb_t, \qb_t)>\frac{M}{2\left(\sqrt{t+1}-1\right)}\left(9+11\Omega+7\ln t+3\Omega\ln t\right)\right] 
.
\end{split}
\]
We complete the proof by setting $\delta=2 \exp(-\Omega)$.

\subsection{Proof of Theorem~\ref{thm:2}} \label{sec:prof:2}
Our goal is to analyze SMD for non-oblivious OCO with stochastic gradients. In the literature, we did not find a convenient reference for it. A very close one is the Lemma 3.2 of \citet{Flaxman:2005:OCO}, which bounds the expected regret of SGD for non-oblivious OCO. But it is insufficient for our purpose, so we provide our proof by following the analysis of SMD for stochastic convex-concave optimization \citep[\S~3]{nemirovski-2008-robust}. Notice that we cannot use the theoretical guarantee of SMD for SCO \citep[\S~2.3]{nemirovski-2008-robust}, because the objective function is fixed in SCO.

From the standard analysis of mirror descent, e.g., Lemma 2.1 of \citet{nemirovski-2008-robust}, we have
\begin{equation} \label{eqn:smd:0}
\begin{split}
\langle  \gt_{w}(\w_t,\q_t), \w_t - \w \rangle  \leq \frac{B_w(\w, \w_t) - B_w(\w,\w_{t+1}) }{\eta_w} + \frac{\eta_w}{2} \| \gt_{w}(\w_t,\q_t)\|_{w,*}^2.
\end{split}
\end{equation}
Summing the above inequality over $t=1,\ldots,T$, we have
\begin{equation} \label{eqn:smd:1}
\begin{split}
\sum_{t=1}^T \langle  \gt_{w}(\w_t,\q_t), \w_t - \w &\rangle  \leq   \frac{B_w(\w, \w_1)}{\eta_w} + \frac{\eta_w}{2} \sum_{t=1}^T \| \gt_{w}(\w_t,\q_t)\|_{w,*}^2\\
&\overset{\text{(\ref{eqn:gradient}),(\ref{eqn:stoch:grad:2})}}{\leq}  \frac{B_w(\w, \w_1)}{\eta_w} + \frac{\eta_w T G^2}{2} \leq \frac{\Dw^2}{\eta_w} + \frac{\eta_w T G^2}{2}\\
\end{split}
\end{equation}
where the last step is due to \citep[(2.42)]{nemirovski-2008-robust}
\begin{equation} \label{eqn:smd:2}
\max_{\w \in \W} B_w(\w, \w_1) \leq \max_{\w \in \W} \nu_w(\w) -\min_{\w \in \W} \nu_w(\w)  \overset{\text{(\ref{eqn:domain:W})}}{\leq} \Dw^2.
\end{equation}

By Jensen's inequality, we have
\[
\begin{split}
&\sum_{t=1}^T \left[ \phi(\w_t,\q_t) -  \phi(\w,\q_t) \right] \leq  \sum_{t=1}^T \langle \nabla_\w \phi(\w_t,\q_t), \w_t-\w\rangle\\
=& \sum_{t=1}^T\langle \gt_{w}(\w_t,\q_t), \w_t-\w\rangle + \sum_{t=1}^T \langle \nabla_\w \phi(\w_t,\q_t) - \gt_{w}(\w_t,\q_t), \w_t-\w\rangle \\
\overset{\text{(\ref{eqn:smd:1})}}{\leq} & \frac{\Dw^2}{\eta_w} + \frac{\eta_w TG^2}{2}  + \sum_{t=1}^T \langle \nabla_\w \phi(\w_t,\q_t) - \gt_{w}(\w_t,\q_t), \w_t-\w\rangle.
\end{split}
\]
Maximizing each side over $\w \in \W$, we arrive at
\begin{equation} \label{eqn:smd:3}
\begin{split}
&\max_{\w\in\W}  \sum_{t=1}^T \left[\phi(\w_t,\q_t) -  \phi(\w,\q_t) \right]= \sum_{t=1}^T \phi(\w_t,\q_t) - \min_{\w\in\W} \sum_{t=1}^T \phi(\w,\q_t)\\
\leq &  \frac{\Dw^2}{\eta_w} + \frac{\eta_w TG^2}{2}  +\max_{\w \in \W} \left\{\underbrace{\sum_{t=1}^T \langle \nabla_\w \phi(\w_t,\q_t) - \gt_{w}(\w_t,\q_t), \w_t-\w\rangle}_{:=F(\w)} \right\}.
\end{split}
\end{equation}

Next, we bound the last term in (\ref{eqn:smd:3}), i.e., $\max_{\w \in \W} F(\w)$. Because $\E_{t-1}[\gt_{w}(\w_t,\q_t)]=\nabla_\w \phi(\w_t,\q_t)$, $F(\w)$ is the sum of a martingale difference sequence for any \emph{fixed} $\w$. However, it is not true for $\wt=\argmax_{\w \in \W} F(\w)$, because $\wt$ depends on the randomness of the algorithm. Thus, we cannot directly apply techniques for martingales to bounding  $\max_{\w \in \W} F(\w)$. This is the place where the analysis differs from that of SCO.

To handle the above challenge, we introduce a virtual sequence of variables to decouple the dependency \citep[proof of Lemma 3.1]{nemirovski-2008-robust}. Imagine there is an online algorithm which performs SMD by using $\nabla_\w \phi(\w_t,\q_t) - \gt_{w}(\w_t,\q_t)$ as the gradient:
\begin{equation} \label{eqn:virtual:smd}
\v_{t+1}= \argmin_{\w \in \W} \big\{ \eta_w \langle  \nabla_\w \phi(\w_t,\q_t) - \gt_{w}(\w_t,\q_t) , \w -\v_t\rangle + B_w(\w,\v_t) \big\}
\end{equation}
where $\v_1=\w_1$. By repeating the derivation of (\ref{eqn:smd:1}), we can show that
\begin{equation} \label{eqn:smd:4}
\begin{split}
&\sum_{t=1}^T \langle  \nabla_\w \phi(\w_t,\q_t) - \gt_{w}(\w_t,\q_t), \v_t - \w \rangle  \\
\leq  & \frac{B_w(\w, \w_1)}{\eta_w} + \frac{\eta_w}{2} \sum_{t=1}^T \| \nabla_\w \phi(\w_t,\q_t) - \gt_{w}(\w_t,\q_t)\|_{w,*}^2 \leq \frac{\Dw^2}{\eta_w} + 2 \eta_w T G^2
\end{split}
\end{equation}
where in the last inequality, we make use of (\ref{eqn:smd:2}) and
\begin{equation} \label{eqn:smd:5}
\begin{split}
 &\| \nabla_\w \phi(\w_t,\q_t) - \gt_{w}(\w_t,\q_t)\|_{w,*} \leq \|\phi(\w_t,\q_t)\|_{w,*} + \|\gt_{w}(\w_t,\q_t)\|_{w,*}  \\
\leq &\E_{t-1}[\| \gt_{w}(\w_t,\q_t)\|_{w,*}] + \|\gt_{w}(\w_t,\q_t)\|_{w,*}  \overset{\text{(\ref{eqn:gradient}),(\ref{eqn:stoch:grad:2})}}{\leq} 2G.
\end{split}
\end{equation}

Then, we have
\begin{equation} \label{eqn:smd:6}
\begin{split}
& \max_{\w \in \W} \left\{\sum_{t=1}^T \langle \nabla_\w \phi(\w_t,\q_t) - \gt_{w}(\w_t,\q_t), \w_t-\w\rangle \right\} \\
= & \max_{\w \in \W} \left\{\sum_{t=1}^T \langle \nabla_\w \phi(\w_t,\q_t) - \gt_{w}(\w_t,\q_t), \v_t-\w\rangle\right\}  \\
&+ \sum_{t=1}^T \langle \nabla_\w \phi(\w_t,\q_t) - \gt_{w}(\w_t,\q_t), \w_t-\v_t\rangle   \\
\overset{\text{(\ref{eqn:smd:4})}}{\leq} & \frac{\Dw^2}{\eta_w} + 2 \eta_w T G^2 + \sum_{t=1}^T \underbrace{\langle \nabla_\w \phi(\w_t,\q_t) - \gt_{w}(\w_t,\q_t), \w_t-\v_t\rangle}_{:=V_t}.
\end{split}
\end{equation}
From the updating rule of $\v_t$ in (\ref{eqn:virtual:smd}), we know that $\v_t$ is independent from $\nabla_\w \phi(\w_t,\q_t) - \gt_{w}(\w_t,\q_t)$, and thus $V_1,\ldots,V_T$ is a martingale difference sequence.

Substituting (\ref{eqn:smd:6}) into (\ref{eqn:smd:3}), we have
\begin{equation} \label{eqn:smd:7}
\sum_{t=1}^T \phi(\w_t,\q_t) - \min_{\w\in\W} \sum_{t=1}^T \phi(\w,\q_t) \leq  \frac{2\Dw^2}{\eta_w} + \frac{5\eta_w TG^2}{2} + \sum_{t=1}^T V_t.
\end{equation}
Taking expectation over both sides, we have
\[
\E \left[ \sum_{t=1}^T \phi(\w_t,\q_t) - \min_{\w\in\W} \sum_{t=1}^T \phi(\w,\q_t) \right] \leq  \frac{2\Dw^2}{\eta_w} + \frac{5\eta_w TG^2}{2} = 2 \Dw G \sqrt{5 T}
\]
where we set $\eta_w= \frac{2 \Dw}{G\sqrt{5T}}$.

To establish high probability bounds, we make use of the Hoeffding-Azuma inequality for martingales stated below \citep{bianchi-2006-prediction}.
\begin{lemma} \label{eqn:azuma}
Let $V_1, V_2,  \ldots$  be a martingale difference sequence with respect to some sequence $X_1, X_2, \ldots$ such that $V_i \in [A_i , A_i + c_i ]$ for some random variable $A_i$, measurable with respect to $X_1, \ldots , X_{i-1}$ and a positive constant $c_i$. If $S_n = \sum_{i=1}^n V_i$, then for any
$t > 0$,
\[
\Pr[ S_n > t] \leq \exp \left( -\frac{2t^2}{\sum_{i=1}^n c_i^2} \right).
\]
\end{lemma}
To apply the above lemma, we need to show that $V_t$ is bounded. Indeed, we have
\begin{equation} \label{eqn:smd:upper_bound_Vt}
\begin{split}
&\left|\langle \nabla_\w \phi(\w_t,\q_t) - \gt_{w}(\w_t,\q_t), \w_t-\v_t\rangle \right|\\
\leq &\|\nabla_\w \phi(\w_t,\q_t) - \gt_{w}(\w_t,\q_t)\|_{w,*} \|\w_t-\v_t\|_w \\
\overset{\text{(\ref{eqn:smd:5})}}{\leq} & 2G \|\w_t-\v_t\|_w \leq 2G \left( \|\w_t-\w_1\|_w + \|\v_t-\w_1\|_w \right)\\
\leq &2G \left( \sqrt{2 B_w(\w_t, \w_1)} + \sqrt{2 B_w(\v_t, \w_1)} \right) \overset{\text{(\ref{eqn:smd:2})}}{\leq} 4 \sqrt{2} \Dw G.
\end{split}
\end{equation}
From Lemma~\ref{eqn:azuma}, with probability at least $1-\delta$, we have
\begin{equation} \label{eqn:smd:8}
\sum_{t=1}^T V_t \leq 8\Dw G \sqrt{T \ln \frac{1}{\delta}}.
\end{equation}
We complete the proof by substituting (\ref{eqn:smd:8}) into (\ref{eqn:smd:7}).

\subsection{Proof of Theorem~\ref{thm:3}} \label{sec:prof:3}
Since we can only observe $\ell(\w_t,\z_t^{(i_t)})$ instead of $R_{i_t}(\w_t)$, the  theoretical guarantee of Exp3-IX \citep{NIPS2015_e5a4d6bf} cannot be directly applied to  Algorithm~\ref{alg:2}. To address this challenge, we generalize the regret analysis of Exp3-IX to stochastic rewards.

By the definition of $\phi(\w,\q)$ in~\eqref{eqn:convex:concave} and the property of linear optimization over the simplex, we have
\begin{equation}\label{eq:transform2}
\begin{split}
   & \max_{\q \in \Delta_m} \sum_{t=1}^T \phi(\w_t,\q) - \sum_{t=1}^T \phi(\w_t,\q_t) =\max_{\q \in \Delta_m}  \sum_{i=1}^m q_i \left(\sum_{t=1}^T R_i(\w_t) \right) - \sum_{t=1}^T \sum_{i=1}^m q_{t,i}R_{i}(\w_t)\\
    = & \sum_{t=1}^T R_{j^*}(\w_t) - \sum_{t=1}^T \sum_{i=1}^m q_{t,i}R_{i}(\w_t)     =  \sum_{t=1}^T \E_{\z \sim \P_{j^*}}[\ell(\w_t;\z)] - \sum_{t=1}^T \sum_{i=1}^m q_{t,i}  \E_{\z \sim \P_{i}}[\ell(\w_t;\z)]  \\
    =& \sum_{t=1}^T \sum_{i=1}^m q_{t,i} s_{t,i} - \sum_{t=1}^T s_{t,j^*}= \sum_{t=1}^T \inner{\q_t}{\s_t} - \sum_{t=1}^T s_{t,j^*}
\end{split}
\end{equation}
where $j^* \in \argmax_{j \in [m]} \sum_{t=1}^T R_j(\w_t)$ and the vector $\s_t \in \R^m$ is defined as
\begin{equation} \label{eq:definition:sti}
s_{t,i} = 1 - \E_{\z \sim \P_i}[\ell(\w_t;\z)] \overset{\text{(\ref{eqn:value})}}{\in}  [0,1], \ \forall i \in [m].
\end{equation}
To facilitate the analysis, we introduce a vector $\hatsb_t \in \R^m$ with
\begin{equation} \label{eq:shat:t}
\hats_{t,i} = 1 -\ell(\w_t; \z_t^{(i)}) \overset{\text{(\ref{eqn:value})}}{\in}  [0,1], \ \forall i \in [m]
\end{equation}
where $\z_t^{(i)}$ denotes a random sample drawn from the $i$-th distribution. Note that $\hatsb_t$ is only used for \emph{analysis}, with the purpose of handling the stochastic rewards. In the algorithm, only $\hats_{t,i_t} = 1 -\ell(\w_t; \z_t^{(i_t)})$ is observed in the $t$-th iteration.

Following the proof of Theorem 1 of \citet{NIPS2015_e5a4d6bf}, we have
\begin{equation}   \label{eq:MAB-Hedge}
\sum_{t=1}^T \inner{\q_t}{\tildesb_t} - \sum_{t=1}^T \tildes_{t,j^*}   \leq \frac{\ln m}{\eta_q} + \frac{\eta_q}{2} \sum_{t=1}^T  \sum_{i=1}^m \tildes_{t,i}
\end{equation}
which makes use of the property of online mirror descent with local norms \citep{Bandit:suvery}.  From (5) of \citet{NIPS2015_e5a4d6bf}, we have
\begin{equation}    \label{eq:highprob-i_t}
 \inner{\q_t}{\tildesb_t}= \sum_{i=1}^m q_{t,i} \tildes_{t,i} = \hats_{t,i_t} - \gamma \sum_{i=1}^m \tildes_{t,i}.
\end{equation}
Combining (\ref{eq:MAB-Hedge}) and (\ref{eq:highprob-i_t}), we have
\begin{equation}    \label{eq:random:bound:1}
\sum_{t=1}^T \hats_{t,i_t}  \leq  \sum_{t=1}^T \tildes_{t,j^*} + \left( \frac{\eta_q}{2} + \gamma\right)\sum_{t=1}^T  \sum_{i=1}^m \tildes_{t,i}  +\frac{\ln m}{\eta_q}.
\end{equation}
From (\ref{eq:transform2}), we have
\begin{equation}    \label{eq:random:bound:2}
\begin{split}
   & \max_{\q \in \Delta_m} \sum_{t=1}^T \phi(\w_t,\q) - \sum_{t=1}^T \phi(\w_t,\q_t) \\
=  &  \sum_{t=1}^T \hats_{t,i_t} - \sum_{t=1}^T s_{t,j^*}  + \sum_{t=1}^T \inner{\q_t}{\s_t} -\sum_{t=1}^T \hats_{t,i_t}\\
 \overset{\eqref{eq:random:bound:1}}{\leq} & \underbrace{\sum_{t=1}^T \big(\tildes_{t,j^*}- s_{t,j^*} \big)}_{:=A}  + \underbrace{\left( \frac{\eta_q}{2} + \gamma\right)\sum_{t=1}^T  \sum_{i=1}^m \tildes_{t,i}}_{:=B}+ \underbrace{\sum_{t=1}^T \big( \inner{\q_t}{\s_t} - \hats_{t,i_t} \big)}_{:=C} +\frac{\ln m}{\eta_q}.
\end{split}
\end{equation}
We proceed to bound the above three terms $A$, $B$ and $C$ respectively.

To bound term $A$, we need the following concentration result concerning the IX loss estimates~\citep[Lemma 1]{NIPS2015_e5a4d6bf}, which we further generalize to the setting with stochastic rewards.
\begin{lemma}
\label{lemma:high-prob-martingale}
Let $\xi_{t,i} \in [0,1]$ for all $t \in [T]$ and $i \in [m]$, and $\tilde{\xi}_{t,i}$ be its IX-estimator defined as $\tilde{\xi}_{t,i} = \frac{\hat{\xi}_{t,i}}{p_{t,i} + \gamma_t} \indicator{i_t = i}$, where  $\hat{\xi}_{t,i} \in [0,1]$, $\E[\hat{\xi}_{t,i}] = \xi_{t,i}$, and the index $i_t$ is sampled from $[m]$ according to the distribution $\p_t \in \Delta_m$. Let $\{\gamma_t\}_{t=1}^T$ be a fixed non-increasing sequence with $\gamma_t \geq 0$ and let $\alpha_{t,i}$ be non-negative $\mathcal{F}_{t-1}$-measurable random variables satisfying $\alpha_{t,i} \leq 2\gamma_t$ for all $t \in [T]$ and $i \in [m]$. Then, with
probability at least $1 -\delta$,
\begin{equation} \label{eq:high-prob-margingale}
    \sum_{t=1}^T \sum_{i=1}^m \alpha_{t,i} (\tilde{\xi}_{t,i} - \xi_{t,i}) \leq \ln \frac{1}{\delta}.
\end{equation}
Furthermore, when $\gamma_t = \gamma \geq 0$ for all $t \in [T]$, the following holds with probability at least $1 -\delta$,
\begin{equation} \label{eq:high-prob-margingale2}
    \sum_{t=1}^T (\tilde{\xi}_{t,i} - \xi_{t,i}) \leq \frac{1}{2\gamma}\ln \frac{m}{\delta}
\end{equation}
simultaneously for all $i \in [m]$.
\end{lemma}

Notice that our construction of $\tildesb_t$ in~\eqref{eq:IX-loss-estimator} satisfies that $\tildes_{t,i} = \frac{\hats_{t,i}}{q_{t,i} + \gamma} \indicator{i_t = i}$ and $i_t$ is drawn from $[m]$ according to $\q_t \in \Delta_m$ as well as $\E[\hats_{t,i}] = s_{t,i}$, which meets the conditions required by Lemma~\ref{lemma:high-prob-martingale}. As a result, according to~\eqref{eq:high-prob-margingale2}, we have
\[
\sum_{t=1}^T (\tildes_{t,j} - s_{t,j}) \leq \frac{1}{2\gamma} \ln \frac{m}{\delta}
\]
for all $j\in [m]$ (including $j^*$) with probability at least $1 -\delta$.

To bound term $B$, we can directly use Lemma 1 of \citet{NIPS2015_e5a4d6bf}, because our setting $\frac{\eta_q}{2}= \gamma$  satisfies its requirement. Thus,  with probability at least $1 -\delta$, we have
\[
\left( \frac{\eta_q}{2} + \gamma\right)\sum_{t=1}^T  \sum_{i=1}^m \tildes_{t,j} \leq \left( \frac{\eta_q}{2} + \gamma\right)\sum_{t=1}^T  \sum_{i=1}^m \hats_{t,j} + \ln \frac{1}{\delta}  \overset{\text{(\ref{eq:shat:t})}}{\leq}  \left( \frac{\eta_q}{2} + \gamma\right) mT + \ln \frac{1}{\delta}.
\]

We now consider term $C$ in~\eqref{eq:random:bound:2}. Let $V_t= \inner{\q_t}{\s_t} - \hats_{t,i_t}$. Then, it is easy to verify that $\E_{t-1}[V_t] = 0$. So, the process $\{V_t\}_{t=1}^T$ forms a martingale difference sequence and it also satisfies $|V_t| \leq 1$ for all $t$. Hence, we can apply Lemma~\ref{eqn:azuma} and have
\[
    \sum_{t=1}^T \big(\inner{\q_t}{\s_t} - \hats_{t,i_t}\big) \leq  \sqrt{2 T \ln \frac{1}{\delta}} \leq \sqrt{\frac{T}{2}} \left(  1+\ln \frac{1}{\delta}\right),
\]
with probability at least $1 -\delta$.

Combining the three upper bounds for the terms $A$, $B$ and $C$, and further taking the union bound, we have, with probability at least $1-\delta$
\[
\begin{split}
    & \max_{q \in \Delta_m} \sum_{t=1}^T \phi(\w_t,\q) - \sum_{t=1}^T \phi(\w_t,\q_t) \\
    \leq & \frac{1}{2\gamma}\ln \frac{3m}{\delta} + \left( \frac{\eta_q}{2} + \gamma\right) mT + \ln \frac{3}{\delta} + \sqrt{\frac{T}{2}} \left(  1+\ln \frac{3}{\delta}\right) + \frac{\ln m}{\eta_q}\\
    = & 2\sqrt{m T \ln m} + \sqrt{\frac{mT}{\ln m}} \cdot \ln \frac{3m}{\delta} + \sqrt{\frac{T}{2}} + \left(\sqrt{\frac{T}{2}}+1 \right) \ln \frac{3}{\delta}\\
    = & 3\sqrt{m T \ln m} + \sqrt{\frac{T}{2}} + \left(\sqrt{\frac{mT}{\ln m}} + \sqrt{\frac{T}{2}} + 1 \right)\ln \frac{3}{\delta},
\end{split}
\]
where the third line holds because of our parameter settings $\gamma = \frac{\eta_q}{2}$ and $\eta_q = \sqrt{\frac{\ln m}{m T}}$.

To obtain the expected regret upper bound based on high probability guarantee, we use the formula as follows \citep[\S~3.2]{Bandit:suvery}.
\begin{lemma}  \label{lemma:high_pro_bound_to_Exp}
	For any real-valued random variable $X$, 
	\[
	\E[X] \leq \int_{0}^1 \frac{1}{\delta} \Pr \left[X > \ln \frac{1}{\delta}\right]~\mathrm{d} \delta.
	\]
\end{lemma}
By setting
\[
X =  \left(\sqrt{\frac{mT}{\ln m}} + \sqrt{\frac{T}{2}} + 1 \right)^{-1} \cdot \left(\max_{q \in \Delta_m} \sum_{t=1}^T \phi(\w_t,\q) - \sum_{t=1}^T \phi(\w_t,\q_t) - 3\sqrt{m T \ln m} - \sqrt{\frac{T}{2}}\right),
\]
we derive $\E[X] \leq 3$ by Lemma~\ref{lemma:high_pro_bound_to_Exp}, which implies
\[
\E\left[\max_{q \in \Delta_m} \sum_{t=1}^T \phi(\w_t,\q) - \sum_{t=1}^T \phi(\w_t,\q_t)\right] \leq 3\sqrt{m T \ln m} + \sqrt{\frac{T}{2}} + 3\left(\sqrt{\frac{mT}{\ln m}} + \sqrt{\frac{T}{2}} + 1 \right). 
\]

\subsection{Proof of Theorem~\ref{thm:4}} \label{sec:prof:4}
By Jensen's inequality and the outputs $\wb = \frac{1}{T} \sum_{t=1}^T \w_t$ and $\qb = \frac{1}{T} \sum_{t=1}^T \q_t$, we have
\begin{equation} \label{eqn:decom:error}
\begin{split}
& \epsilon_{\phi}(\wb, \qb) = \max_{\q\in \Delta_m}  \phi(\wb,\q)- \min_{\w\in \W}  \phi(\w,\qb)\\
\leq & \frac{1}{T} \left(\max_{\q\in \Delta_m}  \sum_{t=1}^T \phi(\w_t,\q)- \min_{\w\in \W}  \sum_{t=1}^T \phi(\w,\q_t) \right) \\
=& \frac{1}{T} \left(\max_{\q\in \Delta_m}  \sum_{t=1}^T \phi(\w_t,\q) - \sum_{t=1}^T \phi(\w_t,\q_t) \right) + \frac{1}{T} \left(\sum_{t=1}^T \phi(\w_t,\q_t) - \min_{\w\in \W}  \sum_{t=1}^T \phi(\w,\q_t) \right)
\end{split}
\end{equation}
and thus
\begin{equation} \label{eqn:decom:exp:error}
\begin{split}
\E \big[\epsilon_{\phi}(\wb, \qb)\big] \leq & \frac{1}{T} \E \left[\left(\max_{\q\in \Delta_m}  \sum_{t=1}^T \phi(\w_t,\q) - \sum_{t=1}^T \phi(\w_t,\q_t) \right)\right] \\
&+ \frac{1}{T}  \E \left[\left(\sum_{t=1}^T \phi(\w_t,\q_t) - \min_{\w\in \W}  \sum_{t=1}^T \phi(\w,\q_t) \right)\right].
\end{split}
\end{equation}

We obtain (\ref{eqn:alg2:high}) by substituting the high probability bounds in Theorems~\ref{thm:2} and \ref{thm:3} into (\ref{eqn:decom:error}), and taking the union bound. Similarly, we obtain (\ref{eqn:alg2:exp})  by substituting the expectation bounds in Theorems~\ref{thm:2} and \ref{thm:3} into (\ref{eqn:decom:exp:error}).

\subsection{Proof of Theorem~\ref{thm:alg2_anytime_w}}
The proof of Theorem~\ref{thm:alg2_anytime_w} closely follows that of Theorem~\ref{thm:2},  with the difference being the use of a time-varying step size $\eta^w_t$.

Similar to \eqref{eqn:smd:0}, by Lemma 2.1 of \citet{nemirovski-2008-robust}, we have 
\begin{equation} \label{eqn:alg2_anytime:1}
 \eta^w_j \langle \gt_{w}(\w_j,\q_j), \w_j - \w \rangle  \leq B_w(\w, \w_j) - B_w(\w,\w_{j+1}) + \frac{(\eta^w_j)^2}{2} \| \gt_{w}(\w_j,\q_j)\|_{w,*}^2.
\end{equation}
Summing \eqref{eqn:alg2_anytime:1} over $j=1,\cdots,t$, we have 
\begin{equation} \label{eqn:alg2_anytime:2}
	\begin{split}
\sum_{j=1}^t \eta^w_j \langle \gt_{w}(\w_j,\q_j), \w_j - \w \rangle
\leq   B_w&(\w, \w_1) + \sum_{j=1}^t \frac{(\eta^w_j)^2}{2} \| \gt_{w}(\w_j,\q_j)\|_{w,*}^2\\
\overset{\text{(\ref{eqn:gradient}),(\ref{eqn:stoch:grad:2}),(\ref{eqn:smd:2})}}{\leq} & \Dw^2 + \frac{G^2}{2} \sum_{j=1}^t (\eta^w_j)^2.
	\end{split}
\end{equation}

By Jensen's inequality, we get
\[
\begin{split}
	&\sum_{j=1}^t \eta^w_j\left[ \phi(\w_j,\q_j) -  \phi(\w,\q_j) \right] 
	\leq  \sum_{j=1}^t \eta^w_j\langle \nabla_\w \phi(\w_j,\q_j), \w_j-\w\rangle\\
	=& \sum_{j=1}^t \eta^w_j\langle \gt_{w}(\w_j,\q_j), \w_j-\w\rangle + \sum_{j=1}^t \eta^w_j\langle \nabla_\w \phi(\w_j,\q_j) - \gt_{w}(\w_j,\q_j), \w_j-\w\rangle \\
	\overset{\text{(\ref{eqn:alg2_anytime:2})}}{\leq} & \Dw^2 + \frac{G^2}{2} \sum_{j=1}^t (\eta^w_j)^2 + \sum_{j=1}^t \eta^w_j\langle \nabla_\w \phi(\w_j,\q_j) - \gt_{w}(\w_j,\q_j), \w_j-\w\rangle.
\end{split}
\]
Maximizing both sides over $\w \in \W$, we obtain 
\begin{equation} \label{eqn:alg2_anytime_w_1}
\begin{split}
		&\max_{\w\in \W} \sum_{j=1}^t \eta^w_j\left[\phi\left(\w_j,\q_j\right)- \phi\left(\w,\q_j\right)\right]\\
		\leq &  \Dw^2 + \frac{G^2}{2}\sum_{j=1}^t (\eta^w_j)^2 +\max_{\w \in \W} \left\{\underbrace{\sum_{j=1}^t \eta^w_j\langle \nabla_\w \phi(\w_j,\q_j) - \gt_{w}(\w_j,\q_j), \w_j-\w\rangle}_{:=F_t(\w)} \right\}.
	\end{split}
\end{equation}

To handle the last term in (\ref{eqn:alg2_anytime_w_1}), we also construct a virtual sequence of variable:
\begin{equation} \label{eqn:virtual:smd:timevarying}
\v_{j+1}= \argmin_{\w \in \W} \big\{ \eta^w_j \langle  \nabla_\w \phi(\w_j,\q_j) - \gt_{w}(\w_j,\q_j) , \w -\v_j\rangle + B_w(\w,\v_j) \big\},
\end{equation}
where $\v_1=\w_1$. The difference between \eqref{eqn:virtual:smd:timevarying} and \eqref{eqn:virtual:smd} lies in the use of the time-varying step size $\eta^w_j$ in \eqref{eqn:virtual:smd:timevarying}. By repeating the derivation of (\ref{eqn:alg2_anytime:2}), we have
\begin{equation} \label{eqn:alg2_anytime:3}
	\begin{split}
&\sum_{j=1}^t \eta^w_j \langle \nabla_\w \phi(\w_j,\q_j) - \gt_{w}(\w_j,\q_j), \v_j - \w \rangle  \\
\leq  & B_w(\w, \w_1) + \sum_{j=1}^t \frac{(\eta^w_j)^2}{2} \| \nabla_\w \phi(\w_j,\q_j) - \gt_{w}(\w_j,\q_j)\|_{w,*}^2 
\overset{\eqref{eqn:smd:2},\eqref{eqn:smd:5}}{\leq} \Dw^2 + 2G^2\sum_{j=1}^t (\eta^w_j)^2
	\end{split}
\end{equation}

Then, we have
\begin{equation} \label{eqn:alg2_anytime_w_2}
	\begin{split}
\max_{\w \in \W} F_t(\w)
=& \max_{\w \in \W}\left\{\sum_{j=1}^t\eta^w_j \langle \nabla_\w \phi(\w_j,\q_j) - \gt_{w}(\w_j,\q_j), \v_j-\w\rangle\right\}\\
&+ \sum_{j=1}^t\eta^w_j \langle \nabla_\w \phi(\w_j,\q_j) - \gt_{w}(\w_j,\q_j), \w_j-\v_j\rangle \\
\overset{\eqref{eqn:alg2_anytime:3}}{\leq} & \Dw^2 + 2G^2\sum_{j=1}^t (\eta^w_j)^2 + \sum_{j=1}^t \underbrace{\eta^w_j \langle \nabla_\w \phi(\w_j,\q_j) - \gt_{w}(\w_j,\q_j), \w_j-\v_j\rangle}_{:=W_j}.
	\end{split}
\end{equation}
Combining \eqref{eqn:alg2_anytime_w_1} and \eqref{eqn:alg2_anytime_w_2}, we have 
\begin{equation} \label{eqn:alg2_anytime_w_3}
\max_{\w\in \W} \sum_{j=1}^t \eta^w_j\left[\phi\left(\w_j,\q_j\right)- \phi\left(\w,\q_j\right)\right] 	\leq   2\Dw^2 + \frac{5G^2}{2}\sum_{j=1}^t (\eta^w_j)^2 +\sum_{j=1}^t W_j.
\end{equation}

Following the same arguments in the proof of Theorem~\ref{thm:2}, we know that $\{W_j\}_{j=1}^t$ is a martingale difference sequence. Taking expectation over both sides of (\ref{eqn:alg2_anytime_w_3}), we have
\[
\E \left[\max_{\w\in \W} \sum_{j=1}^t \eta^w_j\left[\phi\left(\w_j,\q_j\right)- \phi\left(\w,\q_j\right)\right]\right] \leq 2\Dw^2 + \frac{5G^2}{2}\sum_{j=1}^t (\eta^w_j)^2
\]
which implies
\[
\begin{split}
 \E\big[O_1\big]  \leq &\left(\sum_{j=1}^t\eta^w_j\right)^{-1} \left[ 2\Dw^2 + \frac{5G^2}{2}\sum_{j=1}^t (\eta^w_j)^2 \right]\\
\overset{\eqref{eqn:alg2_anytime_stepsize}}{=} &\left(\sum_{j=1}^t \frac{1}{\sqrt{j}}\right)^{-1} \left[ 2\Dw G + \frac{5DG}{2}\sum_{j=1}^t \frac{1}{j} \right] \overset{\eqref{eqn:anytime_fact}}{\leq} \frac{DG}{\left(\sqrt{t+1}-1\right)}\left(\frac{9}{4} + \frac{5}{4} \ln t\right).
\end{split}
\]

Then, we proceed to establish the high probability bound. From \eqref{eqn:smd:upper_bound_Vt}, we have $|W_j|\leq4\sqrt{2}\eta^w_jDG$ for all $j\in \zn$. By Lemma~\ref{eqn:azuma}, with probability at least $1-\delta$, we have
\begin{equation} \label{eqn:alg2_anytime_w_4}
	\sum_{j=1}^t W_t \leq 8\Dw G \sqrt{\sum_{j=1}^t(\eta^w_j)^2 \ln \frac{1}{\delta}}.
\end{equation}
Substituting \eqref{eqn:alg2_anytime_w_4} into \eqref{eqn:alg2_anytime_w_3}, with probability at least $1-\delta$, we have
\begin{equation} \label{eqn:alg2_anytime_w_5}
\max_{\w\in \W} \sum_{j=1}^t \eta^w_j\left[\phi\left(\w_j,\q_j\right)- \phi\left(\w,\q_j\right)\right] \leq 2\Dw^2 + \frac{5G^2}{2} \sum_{j=1}^t (\eta^w_j)^2  + 8\Dw G \sqrt{\sum_{j=1}^t(\eta^w_j)^2 \ln \frac{1}{\delta}}.
\end{equation}
Thus,
\[
\begin{split}
O_1 \overset{\eqref{eqn:alg2_anytime_w_5},\eqref{eqn:alg2_anytime_stepsize}}{\leq} &  \left(\sum_{j=1}^t \frac{1}{\sqrt{j}}\right)^{-1} \left[ 2\Dw G + \frac{5DG}{2}\sum_{j=1}^t \frac{1}{j} + 8DG \sqrt{\left(\sum_{j=1}^t \frac{1}{j} \right)\ln \frac{1}{\delta}}\right]  \\
\overset{\eqref{eqn:anytime_fact}}{\leq} & \frac{DG}{2\left(\sqrt{t+1}-1\right)}\left(2+ \frac{5}{2} (1+\ln t)+8\sqrt{(1+\ln t )\ln \frac{1}{\delta}}\right]\\
\leq &\frac{DG}{\sqrt{t+1}-1}\left(\frac{17}{4}+\frac{13}{4}\ln t+2\ln\frac{1}{\delta}\right)
\end{split}
\]
where in the last step we use the fact that $2 \sqrt{(1+\ln t )\ln (1/\delta)} \leq 1+\ln t + \ln (1/\delta) $.

\subsection{Proof of Theorem~\ref{thm:alg2_anytime_q}} \label{sec:online:weighted}
We will modify the proof of Theorem~\ref{thm:3}  to bound the weighted average regret $O_2$. 

Similar to \eqref{eq:transform2}, we have
\begin{equation}\label{eqn:alg2_anytime_q_1}
	\begin{split}
	& \max_{\q\in \Delta_m}\sum_{j=1}^t \eta^q_j \phi\left(\w_j,\q\right)-\sum_{j=1}^t\eta^q_j\phi\left(\w_j,\q_j\right) \\
	= & \max_{\q \in \Delta_m}  \sum_{i=1}^m q_i \left(\sum_{j=1}^t\eta^q_j R_i(\w_j) \right) - \sum_{j=1}^t \eta^q_j \sum_{i=1}^m q_{j,i}R_{i}(\w_j)\\
	= & \sum_{j=1}^t \eta^q_jR_{k_t^*}(\w_j) - \sum_{j=1}^t \eta^q_j\sum_{i=1}^m q_{j,i}R_{i}(\w_j) \\
	= & \sum_{j=1}^t \eta^q_j\sum_{i=1}^m q_{j,i} s_{j,i} - \sum_{j=1}^t \eta^q_js_{j,k_t^*}
	= \sum_{j=1}^t \eta^q_j\inner{\q_j}{\s_j} - \sum_{j=1}^t \eta^q_js_{j,k_t^*}
	\end{split}
\end{equation}
where $k_t^* \in \argmax_{i \in [m]} \sum_{j=1}^t \eta^q_jR_i(\w_j)$ and $\s_t \in \R^m$ is defined in \eqref{eq:definition:sti}. 

By using the property of online mirror descent with local norms (\citeauthor{Bandit:suvery}, \citeyear{Bandit:suvery}, Theorem 5.5; \citeauthor{Modern:Online:Learning}, \citeyear{Modern:Online:Learning}, \S~6.5 and \S~6.6), we have 
\begin{equation}   \label{eqn:alg2_anytime_q_2}
\begin{split}
	\sum_{j=1}^t \eta^q_j\inner{\q_j}{\tildesb_j} - \sum_{j=1}^t \eta^q_j \tildes_{j,k_t^*}   \leq &  \ln m + \frac12 \sum_{j=1}^t (\eta^q_j)^2 \sum_{i=1}^m q_{j,i} \tildes_{j,i}^2  \\
\leq & \ln m + \frac12 \sum_{j=1}^t (\eta^q_j)^2 \sum_{i=1}^m \tildes_{j,i}
\end{split}
\end{equation}
where the last step follows from the fact that $ q_{j,i} \tildes_{j,i}  \leq 1$. We rewrite \eqref{eq:highprob-i_t} as
\begin{equation}    \label{eqn:alg2_anytime_q_3}
	\inner{\q_j}{\tildesb_j}= \sum_{i=1}^m q_{j,i} \tildes_{j,i} = \hats_{j,i_t} - \gamma_j \sum_{i=1}^m \tildes_{j,i}
\end{equation}
where $\hatsb_j  \in \R^m$ is defined in \eqref{eq:shat:t}. Then, we have
\begin{equation}    \label{eqn:alg2_anytime_q_4}
\begin{split}
\sum_{j=1}^t  \eta^q_j \hats_{j,i_t} \overset{\eqref{eqn:alg2_anytime_q_3}}{=} & \sum_{j=1}^t \eta^q_j \inner{\q_j}{\tildesb_j} + \sum_{j=1}^t \gamma_j \eta^q_j\sum_{i=1}^m \tildes_{j,i} \\
\overset{\eqref{eqn:alg2_anytime_q_2}}{\leq} & \sum_{j=1}^t \eta^q_j \tildes_{j,k_t^*} +  \sum_{j=1}^t \left( \frac{(\eta^q_j)^2}{2} +\gamma_j \eta^q_j \right) \sum_{i=1}^m \tildes_{j,i} +  \ln m .
\end{split}
\end{equation}

Based on \eqref{eqn:alg2_anytime_q_1}, we have
\[
	\begin{split}
		& \max_{\q\in \Delta_m}\sum_{j=1}^t \eta^q_j \phi\left(\w_j,\q\right)-\sum_{j=1}^t\eta^q_j\phi\left(\w_j,\q_j\right) \\
		= & \sum_{j=1}^t \eta^q_j\inner{\q_j}{\s_j} - \sum_{j=1}^t \eta^q_j\hats_{j,i_j} + \sum_{j=1}^t \eta^q_j\hats_{j,i_j} - \sum_{j=1}^t \eta^q_js_{j,k_t^*}\\
	\end{split}
\]
\begin{equation}\label{eqn:alg2_anytime_q_decom_ABC}
	\begin{split}
		\overset{\eqref{eqn:alg2_anytime_q_4}}{\leq} & \underbrace{\sum_{j=1}^t \eta^q_j\big(\tildes_{j,k_t^*}- s_{j,k_t^*} \big)}_{:=A_t}  + \underbrace{\sum_{j=1}^t \left( \frac{(\eta^q_j)^2}{2} + \gamma_j\eta^q_j\right)  \sum_{i=1}^m \tildes_{j,i}}_{:=B_t}+ \underbrace{\sum_{j=1}^t \eta^q_j\big( \inner{\q_j}{\s_j} - \hats_{j,i_j} \big)}_{:=C_t} + \ln m.
	\end{split}
\end{equation}
Next, we bound three terms $A_t$, $B_t$ and $C_t$, respectively.

For term $A_t$, recall that we set $\eta^q_t=2\gamma_t$ in (\ref{eqn:alg2_anytime_stepsize}). In Section~\ref{sec:prof:3}, we have verified that our constructions of $\tildesb_t$ and $\hatsb_t$ satisfy the requirement of Lemma~\ref{lemma:high-prob-martingale}. Then, by setting $\alpha_{t,i}=\eta^q_t\ind[i=k]\leq 2\gamma_t$ in \eqref{eq:high-prob-margingale},  with probability at least $1-\delta$,  we have
\[
	\sum_{j=1}^t \eta^q_j\big(\tildes_{j,k}- s_{j,k} \big)\leq\ln \frac{1}{\delta}
\]
 for each $k\in[m]$. Taking the union bound, we conclude that with probability at least $1-\delta$
\begin{equation}\label{eqn:alg2_anytime_q_res_A}
	\sum_{j=1}^t \eta^q_j\big(\tildes_{j,k_t^*}- s_{j,k_t^*} \big)\leq\ln \frac{m}{\delta}.
\end{equation}

For term $B_t$, we apply Lemma 1 of \citet{NIPS2015_e5a4d6bf} with $\alpha_{t,i} =\frac{(\eta^q_t)^2}{2} + \gamma_t\eta^q_t = (\eta^q_t)^2$. It is easy to very that $\eta^q_t \in [0,1]$, and thus $\alpha_{t,i} \leq \eta^q_t = 2 \gamma_t$. Then,  with probability at least $1-\delta$, we have
\begin{equation}\label{eqn:alg2_anytime_q_res_B}
	\begin{split}
		\sum_{j=1}^t \left( \frac{(\eta^q_j)^2}{2} + \gamma_j\eta^q_j\right)  \sum_{i=1}^m \tildes_{j,i}
		&\leq \sum_{j=1}^t \left( \frac{(\eta^q_j)^2}{2} + \gamma_j\eta^q_j\right)  \sum_{i=1}^m \hats_{j,i}+\ln \frac{1}{\delta}\\
		&\leq m\sum_{j=1}^t (\eta^q_j)^2+\ln \frac{1}{\delta}.
	\end{split}
\end{equation}

To bound term $C_t$, we define a martingale difference sequence $W_j=\eta^q_j(\inner{\q_j}{\s_j}-\hats_{j,i_j} )$, $j \in  \zn$. Then, it can be shown that $|W_j|\leq\eta^q_j$ for all $j$. Applying Lemma~\ref{eqn:azuma}, with probability at least $1-\delta$,  we have
\begin{equation}\label{eqn:alg2_anytime_q_res_C}
	\sum_{j=1}^t \eta^q_j\big( \inner{\q_j}{\s_j} - \hats_{j,i_j} \big) 	\leq \sqrt{2\sum_{j=1}^t(\eta^q_j)^2\ln\frac{1}{\delta}} 	\leq\sqrt{\frac{1}{2}\sum_{j=1}^t(\eta^q_j)^2}\left(1+\ln\frac{1}{\delta}\right).
\end{equation}

Substituting (\ref{eqn:alg2_anytime_q_res_A}), (\ref{eqn:alg2_anytime_q_res_B}) and (\ref{eqn:alg2_anytime_q_res_C}) into (\ref{eqn:alg2_anytime_q_decom_ABC}), and taking the union bound,  with probability at least $1-\delta$, we have 
\begin{equation}\label{eqn:alg2_anytime_q_decom_ABC:sum}
	\begin{split}
		& \max_{\q\in \Delta_m}\sum_{j=1}^t \eta^q_j \phi\left(\w_j,\q\right)-\sum_{j=1}^t\eta^q_j\phi\left(\w_j,\q_j\right) \\
		\leq & 2\ln m+  m\sum_{j=1}^t (\eta^q_j)^2+2 \ln \frac{3}{\delta} + \sqrt{\frac{1}{2}\sum_{j=1}^t(\eta^q_j)^2}\left(1+\ln\frac{3}{\delta}\right) .
	\end{split}
\end{equation}
Thus, 
\[
\begin{split}
O_2 \\
 \overset{\eqref{eqn:alg2_anytime_q_decom_ABC:sum},\eqref{eqn:alg2_anytime_stepsize}}{\leq} &\sqrt{\frac{m}{\ln m}} \left( \sum_{j=1}^t \frac{1}{\sqrt{t}} \right)^{-1} \left( 2\ln m +  (\ln m ) \sum_{j=1}^t \frac{1}{j} +2 \ln \frac{3}{\delta} +\sqrt{  \frac{\ln m}{2 m} \sum_{j=1}^t \frac{1}{t} }\left(1+\ln\frac{3}{\delta}\right) \right)\\
\overset{\eqref{eqn:anytime_fact}}{\leq}&
\frac{1}{2\left(\sqrt{t+1}-1\right)} \left(\left(3+\ln t\right)\sqrt{m\ln m}+\left(2\sqrt{\frac{m}{\ln m}}+\sqrt{\frac{1+\ln t}{2}}\right)\ln\frac{3}{\delta}+\sqrt{\frac{1+\ln t}{2}}\right).
\end{split}
\]

To obtain the expected upper bound of $O_2$, we define 
\[
	X=\left(\left(2\sqrt{\frac{m}{\ln m}}+\sqrt{\frac{1+\ln t}{2}}\right)\right)^{-1}\cdot\left(2\left(\sqrt{t+1}-1\right)O_2-\left(3+\ln t\right)\sqrt{m\ln m}-\sqrt{\frac{1+\ln t}{2}}\right).
\]
By Lemma~\ref{lemma:high_pro_bound_to_Exp}, we have $\E[X]\leq 3$, which implies
\[
	\E\left[O_2\right]\leq \frac{1}{2\left(\sqrt{t+1}-1\right)}\left(\left(3+\ln t\right)\sqrt{m\ln m}+6\sqrt{\frac{m}{\ln m}}+4\sqrt{\frac{1+\ln t}{2}}\right).
\]

\subsection{Proof of Theorem~\ref{thm:alg2_anytime}} \label{sec:thm:anytime:online}
According to \eqref{eqn:alg2_anytime_stepsize}, we can rewrite $\eta^w_t=c_w\eta_t$ and $\eta^q_t=c_q\eta_t$ with $c_w=D/G, c_q=\sqrt{(\ln m)/m}$ and $\eta_t=\sqrt{1/t}$ . Then, we decompose the optimization error in the $t$-th round using the convexity-concavity of $\phi\left(\cdot,\cdot\right)$: 
%\cite[(3.9)]{nemirovski-2008-robust}
\begin{equation} \label{eqn:decom:error_anytime}
	\begin{split}
		& \epsilon_{\phi}(\wb_t, \qb_t) = \max_{\q\in \Delta_m}  \phi(\wb_t,\q)- \min_{\w\in \W}  \phi(\w,\qb_t)\\
		\overset{\eqref{eqn:anytime_output}}{=} & \max_{\q\in \Delta_m}  \phi\left(\sum_{j=1}^t\frac{\eta^w_j\w_j}{\sum_{k=1}^t\eta^w_k},\q\right)- \min_{\w\in \W}  \phi\left(\w,\sum_{j=1}^t\frac{\eta^q_j\q_j}{\sum_{k=1}^t\eta^q_k}\right)\\
		= & \max_{\q\in \Delta_m}  \phi\left(\sum_{j=1}^t\frac{\eta_{j}\w_j}{\sum_{k=1}^t\eta_{k}},\q\right)- \min_{\w\in \W}  \phi\left(\w,\sum_{j=1}^t\frac{\eta_{j}\q_j}{\sum_{k=1}^t\eta_{k}}\right)\\
		\leq & \max_{\q\in \Delta_m} \left(\sum_{j=1}^t\eta_{j}\right)^{-1} \sum_{j=1}^t \eta_{j} \phi\left(\w_j,\q\right) -\min_{\w\in \W} \left(\sum_{j=1}^t\eta_{j}\right)^{-1} \sum_{j=1}^t \eta_{j} \phi\left(\w,\q_j\right)\\
		= & \left(\sum_{j=1}^t\eta_{j}\right)^{-1} \left(\max_{\q\in \Delta_m}\sum_{j=1}^t \eta_{j} \left[\phi(\w_j,\q)-\phi(\w_j,\q_j)\right]\right) \\
		&+\left(\sum_{j=1}^t\eta_{j}\right)^{-1} \left(\max_{\w\in \W} \sum_{j=1}^t \eta_{j}\left[\phi(\w_j,\q_j)- \phi(\w,\q_j)\right]\right)\\
		= & O_1+O_2,
	\end{split}
\end{equation}
where $O_1$ and $O_2$ are defined in \eqref{eqn:alg2_anytime_O1O2}. And thus
\begin{equation} \label{eqn:decom:error_anytime:exp}
\E \big[ \epsilon_{\phi}(\wb_t, \qb_t) \big] \leq \E\big[ O_1\big]+ \E\big[ O_2\big].
\end{equation}
We derive \eqref{eqn:alg2_anytime_thm_highpro} by substituting the high probability bounds in Theorems~\ref{thm:alg2_anytime_w} and \ref{thm:alg2_anytime_q} into \eqref{eqn:decom:error_anytime} and taking the union bound. Moreover, we obtain \eqref{eqn:alg2_anytime_thm_exp} by substituting the expectation bounds in Theorems~\ref{thm:alg2_anytime_w} and \ref{thm:alg2_anytime_q} into \eqref{eqn:decom:error_anytime:exp}.

\subsection{Proof of Theorem~\ref{thm:5}}
For the stochastic gradients in (\ref{eqn:stoch:grad:5}), their norm can be upper bounded in the same way as (\ref{eqn:stoch:grad:1}). That is,
\[
\begin{split}
& \|\g_{w}(\w_t,\q_t)\|_{w,*}=\left\| \sum_{i\in C_t}  q_{t,i} \nabla \ell(\w_{t};\z_t^{(i)}) \right\|_{w,*} \leq  \sum_{i\in C_t}   q_{t,i} \left\| \nabla \ell(\w_{t};\z_t^{(i)})\right\|_{w,*} \overset{\text{(\ref{eqn:gradient})}}{\leq} \sum_{i\in C_t} q_{t,i}  G = G,\\
&\|\g_{q}(\w_t,\q_t)\|_\infty = \max_{i\in C_t} | \ell(\w_{t};\z_t^{(i)})| \overset{\text{(\ref{eqn:value})}}{\leq} 1.
\end{split}
\]
So, with exactly the same analysis as Theorem~\ref{thm:1}, we have
\[
\E \big[ \epsilon_{\varphi}(\wb, \qb) \big] \leq 2 \sqrt{\frac{10 (\Dw^2 G^2 + \ln m )}{n_1} }
\]
and with probability at least $1-\delta$,
\begin{equation} \label{eqn:high:bound:1}
\epsilon_{\varphi}(\wb, \qb) \leq  \left(8+2\ln \frac{2}{\delta}\right) \sqrt{\frac{10(\Dw^2 G^2 + \ln m )}{n_1} }.
\end{equation}

Next, we discuss how to bound the risk of $\wb$ on every distribution $\P_i$, i.e., $R_i(\wb)$.  Following the derivation in (\ref{eqn:relation:error}), we know
\[
\max_{i\in[m]} p_i R_i(\wb) - \min_{\w \in \W}  \max_{i\in[m]}  p_i  R_i(\w)  \leq \epsilon_{\varphi}(\wb, \qb).
\]
Thus, for every distribution $\P_i$, $R_i(\wb)$ can be bounded in the following way:
\[
R_i(\wb)  \leq  \frac{1}{p_i}\min_{\w \in \W}  \max_{i\in[m]} p_i  R_i(\w)  + \frac{1}{p_i}\epsilon_{\varphi}(\wb, \qb).
\]
Taking the high probability bound in (\ref{eqn:high:bound:1}) as an example, we have with probability at $1-\delta$
\begin{equation} \label{eqn:single:R:1}
\begin{split}
R_i(\wb)  \leq & \frac{1}{p_i} \min_{\w \in \W}  \max_{i\in[m]} p_i  R_i(\w)  + \frac{1}{p_i} \left(8+2\ln \frac{2}{\delta}\right) \sqrt{\frac{10 (\Dw^2 G^2 + \ln m )}{n_1} } \\
=& \frac{n_1}{n_i}\min_{\w \in \W}  \max_{i\in[m]} p_i  R_i(\w) + \left(8+2\ln \frac{2}{\delta}\right) \frac{\sqrt{10 (\Dw^2 G^2 + \ln m ) n_1}}{n_i}.
\end{split}
\end{equation}

\subsection{Proof of Theorem~\ref{thm:7}}
We first provide some simple facts that will be used later.  From Assumption~\ref{ass:2}, we immediately know that each risk function $R_i(\cdot)$ also belongs to $[0,1]$. As a result, the difference between each risk function and its estimator is well-bounded, i.e., for all $i\in[m]$,
\begin{equation} \label{eqn:error:R}
-1 \leq R_i(\w) - \ell(\w;\z)  \leq 1, \ \forall \w \in \W,  \ \z \sim \P_i.
\end{equation}
From Assumption~\ref{ass:3}, we can prove that each risk function $R_i(\cdot)$ is $G$-Lipschitz continuous. To see this, we have
\begin{equation}\label{eqn:R:Lipschitz:1}
\|\nabla R_i(\w)\|_{w,*} = \| \E_{\z \sim \P_i} \nabla \ell(\w;\z)\|_{w,*}  \leq \E_{\z \sim \P_i} \| \nabla \ell(\w;\z)\big\|_{w,*} \overset{\text{(\ref{eqn:gradient})}}{\leq} G, \ \forall \w \in \W, i \in[m].
\end{equation}
As a result, we have
\begin{equation}\label{eqn:R:Lipschitz:2}
|R_i(\w) - R_i(w')| \leq G \|\w-\w'\|_w, \ \forall \w, \w' \in \W,  i \in[m].
\end{equation}
Furthermore,  the difference between the gradient of $R_i(\cdot)$ and its estimator is also well-bounded, i.e., for all $i\in[m]$,
\begin{equation} \label{eqn:error:RGrad}
\|\nabla R_i(\w)-  \nabla \ell(\w;\z)\|_{w,*} \leq \|\nabla R_i(\w)\|_{w,*} + \| \nabla \ell(\w;\z)\|_{w,*} \overset{\text{(\ref{eqn:gradient}), (\ref{eqn:R:Lipschitz:1})}}{\leq} 2 G, \ \forall \w \in \W,   \z \sim \P_i.
\end{equation}

Recall the definition of the norm $\| \cdot \|$ and dual norm  $\|\cdot\|_*$  for the space $\mathcal{E} \times \R^m$ in (\ref{eqn:norm:new}), and the  distance-generating function $\nu(\cdot)$ in (\ref{eqn:dis:fun}). Following the arguments in Section~\ref{sec:prof:1}, the two updating rules in (\ref{eqn:upate:wt:mirror}) and (\ref{eqn:upate:qt:mirror}) can be merged as
\begin{equation*} \label{eqn:update:merge:1}
[\w_{t+1};\q_{t+1}]= \argmin_{\x \in \W \times \Delta_m} \Big\{ \eta \big \langle[  \g_{w}(\w_t',\q_t'); -\g_{q}(\w_t',\q_t')] , \x -[\w_t';\q_t'] \big\rangle + B(\x,[\w_t';\q_t']) \Big\}
\end{equation*}
where $\eta_w =2 \eta \Dw^2$ and $\eta_q = 2 \eta \ln m$. Similarly, (\ref{eqn:upate:wt:mirror:2}) and (\ref{eqn:upate:qt:mirror:2}) are equivalent to
\begin{equation*} \label{eqn:update:merge:2}
[\w_{t+1}';\q_{t+1}']= \argmin_{\x \in \W \times \Delta_m} \Big\{ \eta \big \langle [\g_{w}(\w_{t+1},\q_{t+1}); -\g_{q}(\w_{t+1},\q_{t+1})] , \x -[\w_t';\q_t'] \big\rangle + B(\x,[\w_t';\q_t']) \Big\}.
\end{equation*}

Let $F([\w;\q])$ be the monotone operator associated with the weighted GDRO problem in (\ref{eqn:convex:concave:weight}), i.e.,
\begin{equation*}\label{eqn:monotone}
%\begin{split}
F([\w;\q])=[\nabla_\w \varphi(\w,\q); -\nabla_\q \varphi(\w,\q) ]\\
=\left[ \sum_{i=1}^m q_i p_i \nabla R_i(\w)  ; -\big[p_1 R_1(\w), \ldots, p_m R_m(\w) \big]^\top \right] .
%\end{split}
\end{equation*}
From our constructions of stochastic gradients in (\ref{eqn:stoch:grad:6}) and (\ref{eqn:stoch:grad:7}), we clearly have
\[
\begin{split}
\E_{t-1} &\left\{[ \g_{w}(\w_t',\q_t'); -\g_{q}(\w_t',\q_t')] \right\}= F([\w_t';\q_t']),\\
 \E_{t-1} &\left\{[ \g_{w}(\w_{t+1},\q_{t+1}); -\g_{q}(\w_{t+1},\q_{t+1})]\right\}= F([\w_{t+1};\q_{t+1}]).
\end{split}
\]
Thus, Algorithm~\ref{alg:4} is indeed an instance of SMPA \citep[Algorithm 1]{Nemirovski:SMP}, and we can use their Theorem 1 and Corollary 1 to bound the optimization error.

Before applying their results, we show that all the preconditions are satisfied. The parameter $\Omega$ defined in (16) of \citet{Nemirovski:SMP} can be upper bounded by
\begin{equation} \label{eqn:omega:nemi}
\begin{split}
\Omega=& \sqrt{2 \max_{\x \in \W \times \Delta_m} B(\x,[\w_1';\q_1']) } \overset{\text{(\ref{eqn:Bregman:merge})}}{=} \sqrt{\frac{1}{ \Dw^2}\max_{\w \in \W} B_w(\w_1,\w_1') + \max_{\q \in  \Delta_m} \frac{1}{\ln m} B_q(\q,\q_1') }  \\
\overset{\text{(\ref{eqn:smd:2})}}{\leq} & \sqrt{\frac{1}{ \Dw^2} \left( \max_{\w \in \W} \nu_w(\w) -\min_{\w \in \W}  \nu_w(\w) \right)+  \frac{1}{\ln m} \left( \max_{\q \in  \Delta_m}  \nu_q(\q) -\min_{\q \in  \Delta_m} \nu_q(\q) \right) }  \overset{\text{(\ref{eqn:domain:W})}}{=}\sqrt{2}.
\end{split}
\end{equation}
Next, we need to demonstrate that  $F([\w;\q])$ is continuous.
\begin{lemma} \label{lem:1} For the monotone operator $F([\w;\q])$, we have
\[
\| F([\w;\q]) -F([\w';\q'])\|_* \leq \Lt \big\|[\w-\w';\q-\q'] \big\|
\]
where $\Lt$ is defined in (\ref{eqn:mirror:parameters}).
\end{lemma}
We proceed to show the variance of the stochastic gradients satisfies the light tail condition. To this end, we introduce the stochastic oracle used in  Algorithm~\ref{alg:4}:
\begin{equation*}\label{eqn:monotone:stochastic}
\g([\w;\q]) = \left[\g_w(\w,\q); -\g_q(\w,\q)\right]
\end{equation*}
where
\[
\begin{split}
\g_w(\w,\q) &=\sum_{i=1}^m q_{i} p_i \left(\frac{n_m}{n_i} \sum_{j=1}^{n_i/n_m} \nabla \ell(\w;\z^{(i,j)}) \right),\\
\g_q(\w,\q) & =\left[p_1 \frac{n_m}{n_1}  \sum_{j=1}^{n_1/n_m} \ell(\w;\z^{(1,j)}), \ldots, p_m \ell(\w;\z^{(m)}) \right]^\top
\end{split}
\]
and $\z^{(i,j)}$ is the $j$-th sample drawn from distribution $\P_i$. The following lemma shows that the variance is indeed sub-Gaussian.
\begin{lemma} \label{lem:2} For the stochastic oracle $\g([\w;\q])$, we have
\[
 \E \left[ \exp \left( \frac{\| F([\w;\q]) -\g([\w;\q])\|_*^2}{\sigma^2 } \right ) \right] \leq  2
\]
where $\sigma^2$ is defined in (\ref{eqn:mirror:parameters}).
\end{lemma}

Based on (\ref{eqn:omega:nemi}), Lemma~\ref{lem:1}, and Lemma~\ref{lem:2}, we can apply the theoretical guarantee of SMPA. Recall that the total number of iterations is $n_m/2$ in Algorithm~\ref{alg:4}. From Corollary 1 of \citet{Nemirovski:SMP}, by setting
\[
\eta=\min \left( \frac{1}{\sqrt{3} \Lt}, \frac{2}{\sqrt{7 \sigma^2 n_m}} \right)
\]
we have
\[
\Pr\left[\epsilon_{\varphi}(\wb, \qb) \geq \frac{7 \Lt}{n_m} + 14\sqrt{\frac{2\sigma^2}{ 3n_m}} + 7\Lambda \sqrt{\frac{\sigma^2}{n_m}} \right] \leq \exp\left(-\frac{\Lambda^2}{3} \right) + \exp\left(-\frac{\Lambda n_m}{2}\right)
\]
for all $\Lambda >0$. Choosing $\Lambda$ such that $\exp(-\Lambda^2/3) \leq \delta/2$ and $\exp(-\Lambda n_m/2) \leq \delta/2$, we have with probability at least $1-\delta$
\[
\epsilon_{\varphi}(\wb, \qb) \leq \frac{7 \Lt}{n_m} +  14\sqrt{\frac{2\sigma^2}{ 3n_m}} + 7\left(\sqrt{3 \log \frac{2}{\delta}} + \frac{2}{n_m} \log \frac{2}{\delta}\right) \sqrt{\frac{\sigma^2}{n_m}}.
\]
Following the derivation of (\ref{eqn:single:R:1}), we have
\begin{equation} \label{eqn:single:R:2}
\begin{split}
& R_i(\wb) - \frac{1}{p_i} \min_{\w \in \W}  \max_{i\in[m]} p_i  R_i(\w) \\
 \leq & \frac{1}{p_i} \left( \frac{7 \Lt}{n_m} + \sqrt{\frac{\sigma^2}{n_m}} \left( 14\sqrt{\frac{2}{ 3}} + 7\sqrt{3\log \frac{2}{\delta}} + \frac{14}{  n_m} \log \frac{2}{\delta} \right) \right).
\end{split}
\end{equation}

Inspired by \citet[\S~4.3.1]{Nemirovski:SMP}, we use the value of $p_i$ in (\ref{eqn:pi:mirror}) to simplify (\ref{eqn:single:R:2}). It is easy to verify that
\begin{align}
&\frac{p_{\max}}{p_i}=\frac{1/\sqrt{n_m} + \sqrt{n_m/n_i}}{1/\sqrt{n_m} + \sqrt{n_m/n_1}} \leq \left(1+ \frac{n_m}{\sqrt{n_i}}\right),
\nonumber \\
&\frac{1}{p_i} \frac{\Lt}{n_m} = O\left( \frac{p_{\max}}{p_i} \frac{\sqrt{\ln m}}{n_m} \right) = O\left( \left(\frac{1}{n_m} + \frac{1}{\sqrt{n_i}}\right)\sqrt{\ln m}\right),\label{eqn:single:R:3} \\
&  p_i  \leq \left(\frac{1}{\sqrt{n_m}} + 1\right)\sqrt{\frac{n_i}{n_m}}, \quad  \omega_{\max}= \max_{i \in [m]}  \frac{p_i^2 n_m}{n_i} \leq  \left( \frac{1}{\sqrt{n_m}} + 1\right)^2,  \nonumber \\
 &\frac{1}{p_i}\sqrt{\omega_{\max}} =  \frac{1/\sqrt{n_m} + \sqrt{n_m/n_i}}{1/\sqrt{n_m} + 1} \sqrt{\omega_{\max}} \leq \frac{1}{\sqrt{n_m}} + \sqrt{\frac{n_m}{n_i}} , \nonumber\\
&\frac{1}{p_i} \sqrt{\frac{\sigma^2}{n_m}} = O\left( \frac{1}{p_i}  \sqrt{\frac{\omega_{\max}(\kappa+\ln^2 m)}{n_m}} \right) = O\left(\left( \frac{1}{n_m} + \frac{1}{\sqrt{n_i}} \right) \sqrt{\kappa+\ln^2 m}\right). \label{eqn:single:R:4}
\end{align}
Substituting (\ref{eqn:single:R:3}) and (\ref{eqn:single:R:4}) into (\ref{eqn:single:R:2}), we have
\[
 R_i(\wb) - \frac{1}{p_i} \min_{\w \in \W}  \max_{i\in[m]} p_i  R_i(\w)  =  O\left(\left( \frac{1}{n_m} + \frac{1}{\sqrt{n_i}} \right) \sqrt{\kappa+\ln^2 m}\right). 
\]

\subsection{Proof of Theorem~\ref{thm:8}}\label{sec:prof:8}
The proof of Theorem~\ref{thm:8} is almost identical to that of Theorem~\ref{thm:1} in Section~\ref{sec:prof:1}, with the only difference being the replacement of the simplex $\Delta_m$ with the capped simplex $\SS_{m,k}$.

To obtain specific convergence rates, we need to analyze the diameter of $\SS_{m,k}$ measured by the neg-entropy function. First, it is easy to verify that $\frac{1}{m}\mathbf{1}=\argmin_{\q\in\SS_{m,k}}\nu_q(\q)$ and $\min_{\q\in\SS_{m,k}}\nu_q(\q)=-\ln m$.  Note that $\nu_q(\q)$ is convex in $\SS_{m,k}$, indicating that the maximum value is attained at the extreme points of $\SS_{m,k}$, i.e., the vectors in $\SS_{m,k}$ that cannot be expressed as a convex combination of other vectors in $\SS_{m,k}$ \cite[Section 4]{Online:Bandit:Minimax}. Specifically, such vectors comprise $k$ elements equal to $1$ and the remaining $m-k$ elements equal to $0$. Thus, $\max_{\q\in\SS_{m,k}}\nu_q(\q)=-\ln k$. In summary, we have
\[
\max_{\q\in\SS_{m,k}}\nu_q(\q)-\min_{\q\in\SS_{m,k}}\nu_q(\q)= -\ln k +\ln m=\ln \frac{m}{k}.
\]

Then, we replace the diameter of the domain of $\q$ from $\ln m$ to $\ln \frac{m}{k}$ in Section~\ref{sec:prof:1}, and obtain Theorem~\ref{thm:8}.

\subsection{Proof of Theorem~\ref{thm:alg5_anytime}}
In anytime extensions, the difference between Algorithm~\ref{alg:1} and Algorithm~\ref{alg:5} also lies in the domain of $\q$. Thus, we can follow the proof of Theorem~\ref{thm:alg1_anytime} in Section~\ref{sec:prof:1_anytime}, where we only need to replace the simplex $\Delta_m$ with the capped simplex $\SS_{m,k}$. From Section~\ref{sec:prof:8}, we know that the diameter of $\SS_{m,k}$ is upper bounded by $\ln \frac{m}{k}$. Therefore, we redefine $M=\sqrt{2 \Dw^2 G^2  +  2 \ln \frac{m}{k}}$, which leads to Theorem~\ref{thm:alg5_anytime}.

\subsection{Proof of Theorem~\ref{thm:9}} \label{sec:online:topk}
Recall the definition of $s_{t,i}$ and $\hat{s}_{t,i}$ in (\ref{eq:definition:sti}) and (\ref{eq:shat:t}) of Section~\ref{sec:prof:3}. Following the analysis of (\ref{eq:transform2}), we have
\begin{equation} \label{equation:p-player-decom-1}
\max_{\q\in\SS_{m,k}}\sum_{t=1}^T\phi(\w_t,\q)-\sum_{t=1}^T\phi(\w_t,\q_t)=\sum_{t=1}^T\langle\q_t,\s_t\rangle-\frac1k\sum_{i\in \I^*}\sum_{t=1}^Ts_{t,i}
\end{equation}
where $\I^*=\argmax_{\I\in \mathcal{B}_{m,k}} \sum_{i\in\I} \big[\sum_{t=1}^TR_i(\w_{t})\big]$. From the new construction of the IX loss estimator  (\ref{eq:IX-loss-estimator-2}), we have 
\begin{equation}\label{eq:IX-loss-estimator:inequality}
q_{t,i}\tilde{s}_{t,i}\leq \frac{q_{t,i}}{kq_{t,i}+\gamma} \leq \frac{1}{k}, \ \forall i\in[m].
\end{equation}
Similar to the derivation of (\ref{eq:MAB-Hedge}) and  (\ref{eqn:alg2_anytime_q_2}), we make use the property of online mirror descent with local norms and proceed with the following steps:
\begin{equation}\label{b-1}
\begin{split}
	\sum_{t=1}^T \inner{\q_t}{\tildesb_t}  -\sum_{t=1}^T\frac1k\sum_{i\in \I^*}\tilde{s}_{t,i} \leq &  \frac{\ln \frac{m}{k}}{\eta_{q}}+  \sum_{t=1}^T \frac{\eta_{q}}{2} \sum_{i=1}^m q_{t,i} \tildes_{t,i}^2  \\
\overset{\text{(\ref{eq:IX-loss-estimator:inequality})}}{\leq} & \frac{\ln \frac{m}{k}}{\eta_{q}}+\frac{\eta_{q}}{2k}\sum_{t=1}^T\sum_{i=1}^m\tilde{s}_{t,i}
\end{split}
\end{equation}
where in the first step we make use the fact that the diameter of $\SS_{m,k}$ is upper bounded by $\ln \frac{m}{k}$. Moreover, (\ref{eq:highprob-i_t}) becomes 
\begin{equation}\label{b-2}
\begin{split}
	\langle\mathbf{q}_t,\tilde{\mathbf{s}}_t\rangle
	\overset{\eqref{eq:IX-loss-estimator-2}, \eqref{eq:shat:t}}{=}\frac{1}{k}\sum_{i=1}^mkq_{t,i}\frac{\hat{s}_{t,i}}{kq_{t,i}+\gamma}\cdot\mathbb{I}[i\in \I_t]
	&=\frac{1}{k}\sum_{i=1}^m\left(1-\frac{\gamma}{kq_{t,i}+\gamma}\right)\hat{s}_{t,i}\cdot\mathbb{I}[i\in \I_t]\\
	&=\frac1k\sum_{i\in \I_t}\hat{s}_{t,i}-\frac{\gamma}{k}\sum_{i=1}^m\tilde{s}_{t,i}.
\end{split}
\end{equation}

Combining \eqref{b-1} and \eqref{b-2}, we have
\begin{equation}\label{b-3}
	\frac1k\sum_{t=1}^T\sum_{i\in \I_t}\hat{s}_{t,i}
	\leq\frac1k\sum_{t=1}^T\sum_{i\in \I^*}\tilde{s}_{t,i}+\frac{\ln \frac{m}{k}}{\eta_{q}}+\left(\frac{\eta_{q}}{2k}+\frac{\gamma}{k}\right)\sum_{t=1}^T\sum_{i=1}^m\tilde{s}_{t,i}.
\end{equation}
From \eqref{equation:p-player-decom-1}, we have
\begin{equation} \label{equation:p-player-decom-2}
	\begin{aligned}
		&\max_{\q\in\SS_{m,k}}\sum_{t=1}^T\phi(\w_t,\q)-\sum_{t=1}^T\phi(\w_t,\q_t)\\
		=&\sum_{t=1}^T\langle\q_t,\s_t\rangle-\frac1k\sum_{t=1}^T\sum_{i\in \I_t}\hat{s}_{t,i}+\frac1k\sum_{t=1}^T\sum_{i\in \I_t}\hat{s}_{t,i}-\frac1k\sum_{i\in \I^*}\sum_{t=1}^Ts_{t,i}\\
		\overset{\eqref{b-3}}{\leq}&\underbrace{\frac1k\sum_{i\in \I^*}\sum_{t=1}^T\left(\tilde{s}_{t,i}-s_{t,i}\right)}_{:=A}+\underbrace{\left(\frac{\eta_{q}}{2k}+\frac{\gamma}{k}\right)\sum_{t=1}^T\sum_{i=1}^m\tilde{s}_{t,i}}_{:=B}+\underbrace{\sum_{t=1}^T\left(\langle\q_t,\s_t\rangle-\frac1k\sum_{i\in \I_t}\hat{s}_{t,i}\right)}_{:=C}+\frac{\ln \frac{m}{k}}{\eta_{q}}. 
	\end{aligned}
\end{equation}
Next, we sequentially bound the above three items $A$, $B$, and $C$.

To bound $A$, we extend Corollary 1 of \citet{NIPS2015_e5a4d6bf} to the modified IX loss estimator \eqref{eq:IX-loss-estimator-2}. 
\begin{lemma} \label{lemma:high-prob-martingale-2}
Let $\xi_{t,i} \in [0,1]$ for all $t \in [T]$ and $i \in [m]$, and  $\tilde{\xi}_{t,i}$ be its IX-estimator defined as $\tilde{\xi}_{t,i} = \frac{\hat{\xi}_{t,i}}{kp_{t,i} + \gamma} \ind[i\in \I_t]$, where $\gamma\geq0$,  $\hat{\xi}_{t,i} \in [0,1]$, $\E[\hat{\xi}_{t,i}] = \xi_{t,i}$, $\p_t\in\SS_{m,k}$, and $\I_t$ is sampled by $\text{\rm DepRound}(k,\p_t)$. Then, with probability at least $1-\delta$,
	\begin{equation} \label{eq:high-prob-martingale-2}
		\sum_{t=1}^T\left(\tilde{\xi}_{t,i}-\xi_{t,i}\right)\leq \frac{1}{2\gamma}\ln\frac{m}{\delta}
	\end{equation}
	simultaneously hold for all $i \in [m]$.
\end{lemma}
Compared to Lemma~\ref{lemma:high-prob-martingale}, Lemma~\ref{lemma:high-prob-martingale-2}  only covers the case where a fixed $\gamma$ is used.

It is easy to verify that the construction of $\tilde{s}_{t,i}$ and $\I_t$ satisfy the conditions outlined in Lemma~\ref{lemma:high-prob-martingale-2}. Therefore, with probability at least $1-\delta$, we have
\begin{equation}	\label{equation:result-A}
	\frac1k \sum_{i\in \I^*} \sum_{t=1}^T\left(\tilde{s}_{t,i}-s_{t,i}\right) \leq \max_{i \in [m]} \sum_{t=1}^T \left(\tilde{s}_{t,i}-s_{t,i}\right) \overset{\eqref{eq:high-prob-martingale-2}}{\leq}  \frac{1}{2\gamma}\ln\frac{m}{\delta}.
\end{equation}
At the same time, we can also deliver an upper bound for $B$. From \eqref{eq:high-prob-martingale-2}, we have 
\begin{equation*}
	\sum_{t=1}^T\sum_{i=1}^m\tilde{s}_{t,i}
	\leq \sum_{t=1}^T\sum_{i=1}^ms_{t,i}+\frac{m}{2\gamma}\ln\frac{m}{\delta}
	\leq mT+\frac{m}{2\gamma}\ln\frac{m}{\delta}.
\end{equation*}
implying
\begin{equation}	\label{equation:result-B}
	\left(\frac{\eta_{q}}{2k}+\frac{\gamma}{k}\right)\sum_{t=1}^T\sum_{i=1}^m\tilde{s}_{t,i}
	\leq \frac{m}{k}\left(\frac{\eta_{q}}{2}+\gamma\right)\left(T+\frac{1}{2\gamma}\ln\frac{m}{\delta}\right).
\end{equation}

As for term $C$, we denote $V_t=\langle\q_t,\s_t\rangle-\frac1k\sum_{i\in \I_t}\hat{s}_{t,i}$. Since 
\[
\E_{t-1}\left[\frac{1}{k}\sum_{i\in \I_t}\hat{s}_{t,i}\right]=\frac{1}{k}\sum_{i=1}^m\Pr[i\in \I_t]s_{t,i}\overset{\eqref{DepRound:prop}}{=}  \sum_{i=1}^m q_{t,i}s_{t,i}
\]
we know that  $\{V_t\}_{t=1}^T$ is a martingale difference sequence. Furthermore, under Assumption~\ref{ass:2} and $\q\in\SS_{m,k}$, we have $|V_t|\leq 1$ for all $t$. By Lemma~\ref{eqn:azuma}, with probability at least $1-\delta$, we have 
\begin{equation}	\label{equation:result-C}
	\sum_{t=1}^T\left(\langle\q_t,\s_t\rangle-\frac1k\sum_{i\in \I_t}\hat{s}_{t,i}\right)\leq \sqrt{2T\ln \frac{1}{\delta}}\leq \sqrt{\frac{T}{2}}\left(1+\ln\frac{1}{\delta}\right).
\end{equation}

Substituting \eqref{equation:result-A}, \eqref{equation:result-B} and \eqref{equation:result-C} into \eqref{equation:p-player-decom-2}, and taking the union bound,\footnote{Because \eqref{equation:result-A} and \eqref{equation:result-B}  depend on the same random event, we can avoid one invocation of the union bound.} 
with probability at least $1-\delta$, we have
\[
\begin{split}
	&\max_{\q\in\SS_{m,k}}\sum_{t=1}^T\phi(\w_t,\q)-\sum_{t=1}^T\phi(\w_t,\q_t)\\
	\leq &\sqrt{\frac{T}{2}}\left(1+\ln\frac{2}{\delta}\right)+\frac{m}{k}\left(\frac{\eta_{q}}{2}+\gamma\right)\left(T+\frac{1}{2\gamma}\ln\frac{2m}{\delta}\right)+\frac{1}{2\gamma}\ln\frac{2m}{\delta}+\frac{\ln \frac{m}{k}}{\eta_{q}}\\
	= & \sqrt{\frac{T}{2}}\left(1+\ln\frac{2}{\delta}\right)+\frac{m}{k}\eta_{q}T+\frac{m}{k}\ln\frac{2m}{\delta}+\frac{1}{\eta_{q}}\ln m+\frac{1}{\eta_{q}}\left(\ln \frac{2}{\delta}+ \ln \frac{m}{k} \right)\\
	= & \sqrt{\frac{T}{2}}\left(1+\ln\frac{2}{\delta}\right)+2\sqrt{\frac{m}{k}T\ln m}+\frac{m}{k}\ln\frac{2m}{\delta}+\sqrt{\frac{mT}{k\ln m}}\ln\frac{2m}{\delta k}\\
	\leq {}& \sqrt{\frac{T}{2}}+\left(\sqrt{\frac{T}{2}}+\frac{m}{k}+\sqrt{\frac{mT}{k\ln m}}\right)\ln\frac{2}{\delta}+3\sqrt{\frac{m}{k}T\ln m}+\frac{m}{k}\ln m.
\end{split}
\]
where we set $\gamma=\frac{\eta_{q}}{2}$ in the 3rd line and  $\eta_{q}=\sqrt{\frac{k \ln m}{m T}}$ in the 4th line.

To get the expected regret bound, we define 
\begin{align*}
	X=&\left(\sqrt{\frac{T}{2}}+\frac{m}{k}+\sqrt{\frac{mT}{k\ln  m}}\right)^{-1}\cdot\\
	&\left(\max_{\q\in\SS_{m,k}}\sum_{t=1}^T\phi(\w_t,\q)-\sum_{t=1}^T\phi(\w_t,\q_t)-\sqrt{\frac{T}{2}}-3\sqrt{\frac{m}{k}T\ln m}-\frac{m}{k}\ln m\right),
\end{align*}
and Lemma~\ref{lemma:high_pro_bound_to_Exp} implies that  $\E[X]\leq2$. Then, we have
\begin{align*}
\E\left[\max_{\q\in\SS_{m,k}}\sum_{t=1}^T\phi(\w_t,\q)-\sum_{t=1}^T\phi(\w_t,\q_t)\right] 	\leq 3\sqrt{\frac{T}{2}}+\frac{2m}{k}+2\sqrt{\frac{mT}{k\ln m}}+3\sqrt{\frac{mT\ln m}{k}}+\frac{m \ln m}{k}. 
\end{align*}

\subsection{Proof of Theorem~\ref{thm:10}}
The proof is almost identical to that of Theorem~\ref{thm:4}. We just need to replace $\Delta_m$ with $\SS_{m,k}$ in  (\ref{eqn:decom:error}) and (\ref{eqn:decom:exp:error}), and then substitute the conclusions of Theorems~\ref{thm:2} and \ref{thm:9}.

\subsection{Proof of Theorem~\ref{thm:alg7}} 
Similar to the proof of Theorem~\ref{thm:alg2_anytime} in Section~\ref{sec:thm:anytime:online}, we decompose the optimization error in the $t$-th round as
\begin{equation} \label{eqn:decom:error_anytime}
	\begin{split}
&\epsilon_{\phi}'(\wb_t, \qb_t)  \overset{\eqref{eqn:anytime_output}}{=}  \max_{\q\in \SS_{m,k}}  \phi\left(\sum_{j=1}^t\frac{\eta^w_j\w_j}{\sum_{k=1}^t\eta^w_k},\q\right)- \min_{\w\in \W}  \phi\left(\w,\sum_{j=1}^t\frac{\eta^q_j\q_j}{\sum_{k=1}^t\eta^q_k}\right)\\
		\leq 	 & \left(\sum_{j=1}^t\eta_{j}\right)^{-1} \left(\max_{\q\in \SS_{m,k}}\sum_{j=1}^t \eta_{j} \left[\phi(\w_j,\q)-\phi(\w_j,\q_j)\right]\right) \\
		&+\left(\sum_{j=1}^t\eta_{j}\right)^{-1} \left(\max_{\w\in \W} \sum_{j=1}^t \eta_{j}\left[\phi(\w_j,\q_j)- \phi(\w,\q_j)\right]\right)\\
		= & O_1+O_2',
	\end{split}
\end{equation}
where $O_1$ is defined in \eqref{eqn:alg2_anytime_O1O2}  and
\[
O_2'=\left(\sum_{j=1}^t\eta^q_j\right)^{-1} \left(\max_{\q\in \SS_{m,k}}\sum_{j=1}^t \eta^q_j \left[\phi(\w_j,\q)-\phi(\w_j,\q_j)\right]\right).
\]

Note that the 1st player is identical to the one in Section~\ref{sec:online:anytime}, so we can directly use Theorem~\ref{thm:alg2_anytime_w} to bound $O_1$. For the 2nd player, due to the difference in the domain, we need to reanalyze and have proven the same upper bounds for $O_2'$  as in Theorem~\ref{thm:alg2_anytime_q}.

\begin{theorem} \label{thm:alg7_anytime_q} 
Under Assumption \ref{ass:2},  we have
\[
\E\big[O_2'\big]\leq \frac{1}{2\left(\sqrt{t+1}-1\right)}\left(\left(3+\ln t\right)\sqrt{m\ln m}+6\sqrt{\frac{m}{\ln m}}+4\sqrt{\frac{1+\ln t}{2}}\right), \ \forall  t \in \zn.
\]
Furthermore, with probability at least $1-\delta$,  we have
\[
O_2'
\leq
\frac{1}{2\left(\sqrt{t+1}-1\right)} \left(\left(3+\ln t\right)\sqrt{m\ln m}+\left(2\sqrt{\frac{m}{\ln m}}+\sqrt{\frac{1+\ln t}{2}}\right)\ln\frac{3}{\delta}+\sqrt{\frac{1+\ln t}{2}}\right)
\]
for each $t \in \zn$.
\end{theorem}
By combining Theorems~\ref{thm:alg2_anytime_w} and \ref{thm:alg7_anytime_q}, we obtain Theorem~\ref{thm:alg7} and the upper bounds are exactly the same as those in  Theorem~\ref{thm:alg2_anytime}.

\subsection{Proof of Theorem~\ref{thm:alg7_anytime_q}}
We need to specifically adjust the proof of Theorem~\ref{thm:alg2_anytime_q} in Section~\ref{sec:online:weighted} based on the fact that the domain is the capped simplex $\SS_{m,k}$.

First, we modify (\ref{eqn:alg2_anytime_q_1}) as
\begin{equation}\label{eqn:alg7_anytime_q_1}
\max_{\q\in \SS_{m,k}}\sum_{j=1}^t \eta^q_j \phi\left(\w_j,\q\right)-\sum_{j=1}^t\eta^q_j\phi\left(\w_j,\q_j\right) = \sum_{j=1}^t \eta^q_j\inner{\q_j}{\s_j} - \frac1k\sum_{i\in \I_t^*} \left(\sum_{j=1}^t \eta^q_js_{j,i}\right)
\end{equation}
where $\I_t^*=\argmax_{\I\in \mathcal{B}_{m,k}} \sum_{i\in\I} \big[\sum_{j=1}^t \eta^q_jR_i(\w_j)\big]$. Based on the property of online mirror descent with local norms and the fact that the diameter of $\SS_{m,k}$ is upper bounded by $\ln \frac{m}{k}$, (\ref{eqn:alg2_anytime_q_2}) becomes 
\begin{equation}   \label{eqn:alg7_anytime_q_2}
\begin{split}
	\sum_{j=1}^t \eta^q_j\inner{\q_j}{\tildesb_j} - \sum_{j=1}^t \eta^q_j \left(\frac1k\sum_{i\in \I_t^*}   \tildes_{j,i} \right) \leq  \ln  \frac{m}{k} + \frac12 \sum_{j=1}^t (\eta^q_j)^2 \sum_{i=1}^m \tildes_{j,i}.
\end{split}
\end{equation}
By using (\ref{eqn:alg7_anytime_q_2}) in the derivation of (\ref{eqn:alg2_anytime_q_4}), we obtain
\begin{equation}    \label{eqn:alg7_anytime_q_4}
\sum_{j=1}^t  \eta^q_j \hats_{j,i_t} \leq \sum_{j=1}^t \eta^q_j \left(\frac1k\sum_{i\in \I_t^*}   \tildes_{j,i} \right)+  \sum_{j=1}^t \left( \frac{(\eta^q_j)^2}{2} +\gamma_j \eta^q_j \right) \sum_{i=1}^m \tildes_{j,i} +  \ln  \frac{m}{k}.
\end{equation}

From \eqref{eqn:alg7_anytime_q_1}, we have
\begin{equation}\label{eqn:alg7_anytime_q_decom_ABC}
	\begin{split}
		& \max_{\q\in \SS_{m,k}}\sum_{j=1}^t \eta^q_j \phi\left(\w_j,\q\right)-\sum_{j=1}^t\eta^q_j\phi\left(\w_j,\q_j\right)  \\
		= & \sum_{j=1}^t \eta^q_j\inner{\q_j}{\s_j} - \sum_{j=1}^t \eta^q_j\hats_{j,i_j} + \sum_{j=1}^t \eta^q_j\hats_{j,i_j} - \frac1k\sum_{i\in \I_t^*} \left(\sum_{j=1}^t \eta^q_js_{j,i}\right)\\
		\overset{\eqref{eqn:alg7_anytime_q_4}}{\leq} & \underbrace{\frac1k  \sum_{i\in \I_t^*} \sum_{j=1}^t \eta^q_j\big(\tildes_{j,i}- s_{j,i} \big)}_{:=A_t}   + \underbrace{\sum_{j=1}^t \left( \frac{(\eta^q_j)^2}{2} + \gamma_j\eta^q_j\right)  \sum_{i=1}^m \tildes_{j,i}}_{:=B_t}+ \underbrace{\sum_{j=1}^t \eta^q_j\big( \inner{\q_j}{\s_j} - \hats_{j,i_j} \big)}_{:=C_t} + \ln  \frac{m}{k}.
	\end{split}
\end{equation}
Next, we bound three terms $A_t$, $B_t$ and $C_t$, respectively. 

Note that (\ref{eqn:alg2_anytime_q_res_A}) in Section~\ref{sec:online:weighted} holds for any possible value of $k_t^* \in [m]$. As a result, with probability at least $1-\delta$, we have
\begin{equation}\label{eqn:alg7_anytime_q_res_A}
\frac1k \sum_{i\in \I_t^*} \sum_{j=1}^t \eta^q_j\big(\tildes_{j,i}- s_{j,i} \big) \leq  \max_{i \in [m]} \sum_{j=1}^t \eta^q_j\big(\tildes_{j,i}- s_{j,i} \big)\leq\ln \frac{m}{\delta}.
\end{equation}
To bound $B_t$ and $C_t$, we can directly use the inequalities in (\ref{eqn:alg2_anytime_q_res_B}) and (\ref{eqn:alg2_anytime_q_res_C}). Substituting (\ref{eqn:alg7_anytime_q_res_A}), (\ref{eqn:alg2_anytime_q_res_B}) and (\ref{eqn:alg2_anytime_q_res_C}) into (\ref{eqn:alg7_anytime_q_decom_ABC}), and taking the union bound,  with probability at least $1-\delta$, we have 
\begin{equation}\label{eqn:alg7_anytime_q_decom_ABC:sum}
	\begin{split}
		& \max_{\q\in \SS_{m,k}}\sum_{j=1}^t \eta^q_j \phi\left(\w_j,\q\right)-\sum_{j=1}^t\eta^q_j\phi\left(\w_j,\q_j\right) \\
		\leq & \ln \frac{m}{k} + \ln m +  m\sum_{j=1}^t (\eta^q_j)^2+2 \ln \frac{3}{\delta} + \sqrt{\frac{1}{2}\sum_{j=1}^t(\eta^q_j)^2}\left(1+\ln\frac{3}{\delta}\right) \\
\leq &2 \ln m+  m\sum_{j=1}^t (\eta^q_j)^2+2 \ln \frac{3}{\delta} + \sqrt{\frac{1}{2}\sum_{j=1}^t(\eta^q_j)^2}\left(1+\ln\frac{3}{\delta}\right) .
	\end{split}
\end{equation}
Note that the final bound in (\ref{eqn:alg7_anytime_q_decom_ABC:sum}) is exactly the same as that in  (\ref{eqn:alg2_anytime_q_decom_ABC:sum}), and therefore we can reach the same conclusion as  Theorem~\ref{thm:alg2_anytime_q}.
\section{Experiments}
We present experiments to evaluate the effectiveness of the proposed algorithms. 

\begin{table}[htbp]
	\centering
	\begin{tabular}{lll}
		\toprule
		Algorithms &  Notation & Highlights \\
		\midrule
		Alg.~1 of \citet{Gouop_DRO} & SMD(1) & SMD with 1 sample per round \\
		\midrule
		Alg.~\ref{alg:1} & SMD($m$) & SMD with $m$ samples per round \\
		\midrule
		Anytime extension of Alg.~\ref{alg:1} & SMD($m$)$_\mathrm{a}$ & SMD($m$) with time-varying step sizes \\
		\midrule
		Alg.~\ref{alg:2} & Online(1) & \makecell[l]{Online learning method with $1$ sample \\  per round}\\
		\midrule
		Anytime extension of Alg.~\ref{alg:2} & Online(1)$_\mathrm{a}$ & Online(1) with time-varying step sizes \\
		\midrule
		Alg.~\ref{alg:3} & SMD$_\mathrm{r}$ & SMD with random sampling \\
		\midrule
		Alg.~\ref{alg:4} & SMPA$_\mathrm{m}$ & SMPA with mini-batches \\
		\midrule
		Alg.~\ref{alg:5} & AT$_k$RO$(m)$ &  SMD($m$) for AT$_k$RO \\
		\midrule
		Anytime extension of Alg.~\ref{alg:5} & AT$_k$RO$(m)_\mathrm{a}$ & \makecell[l]{SMD($m$) with time-varying step sizes \\  for AT$_k$RO} \\
		\midrule
		Alg.~\ref{alg:6} & AT$_k$RO$(k)$ & \makecell[l]{Online learning method with  $k$ samples\\ per round for AT$_k$RO }\\
		\midrule
		Alg.~\ref{alg:7}  & AT$_k$RO$(1)_\mathrm{a}$ & \makecell[l]{Anytime online method with  $1$ sample  \\ per round for AT$_k$RO }\\
		\bottomrule
	\end{tabular}
	\tabcaption{Notation for Algorithms.}
	\label{tab:notation}
\end{table}%
\begin{figure}[t]
	\begin{center}
		\subfigure[The synthetic data set]{
			\label{fig:balance_synth_risk} %% label for second subfigure
			\includegraphics[width=0.4\textwidth]{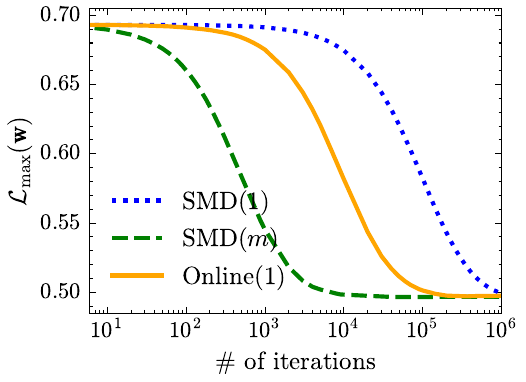}}
		\quad
		\subfigure[The Adult data set]{
			\label{fig:balance_adult_risk} %% label for second subfigure
			\includegraphics[width=0.4\textwidth]{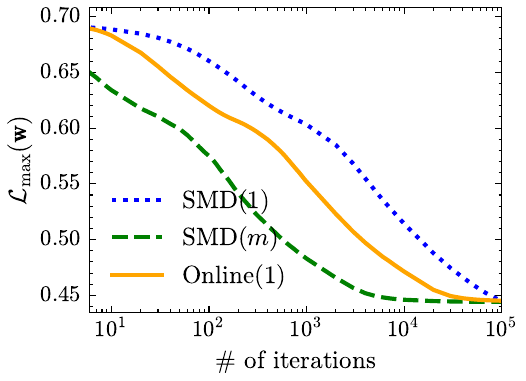}}\\
		\caption{Balanced settings: maximum risk versus the number of iterations.}
		\label{fig:balance_risk}
		\subfigure[The synthetic data set]{
			\label{fig:balance_synth_sample} %% label for second subfigure
			\includegraphics[width=0.4\textwidth]{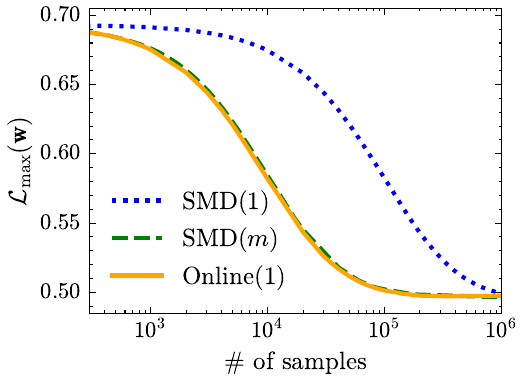}}
		\quad
		\subfigure[The Adult data set]{
			\label{fig:balance_adult_sample} %% label for second subfigure
			\includegraphics[width=0.4\textwidth]{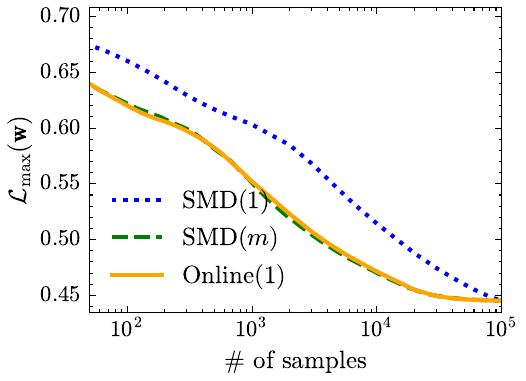}}\\
		\caption{Balanced settings: maximum risk versus the number of samples.}
		\label{fig:balance_sample}\vspace{-2ex}
	\end{center}
\end{figure}

\subsection{Data Sets and Experimental Settings} \label{sec:exp:setting}
Following the setup in previous work \citep{NIPS2016_4588e674,DRO:Online:Game}, we use both synthetic and real-world data sets. 

First, we create a synthetic data set with $m=20$ groups, each associated with a true classifier $\w_i^* \in \R^{1000}$. The set $\{\w_i^*\}_{i \in [m]}$ is constructed  as follows: we start with an arbitrary vector $\w_0$ on the unit sphere; then, we randomly choose $m$ points on a sphere of radius $d$ centered at $\w_0$; these points are projected onto the unit sphere to form $\{\w_i^*\}_{i \in [m]}$.  For distribution $\P_i$, the sample $(\x,y)$ is generated by sampling  $\x$ from the standard normal distribution $\N(0, I)$ and setting $y = \sgn(\x^{\top} \w_i^*)$ with probability $0.9$, or to its inverse with probability $0.1$. We set $d=0.5$ in this data set.

To simulate heterogeneous distributions, we specifically construct another synthetic data set, which contains $m=20$ distributions. The classifiers $\w_i^*$s are generated in the same way as described above. For a sample $\x \sim \N(0, I)$, the distribution $\P_i$ outputs $y = \sgn(\x^{\top} \w_i^*)$ with probability  $p_i$ and $y = -\sgn(\x^{\top} \w_i^*)$ with  probability  $1-p_i$. We choose $\P_1$ as the outlier distribution and set $p_1=0.6$, while the remaining $p_i$ values are uniformly chosen from the range $0.85$ to $0.95$. Additionally, we set $d=0.2$ to ensure that $\{\w_i^*\}_{i \in [m]}$ are close, emphasizing that the heterogeneity is primarily due to noise.

We also use the Adult data set \citep{misc_adult_2}, which includes attributes such as age, gender, race, and educational background of $48,842$ individuals. The objective is to determine whether an individual's income exceeds $50,000$ USD or not.  We set up $m=6$ groups based on the race and gender attributes, where each group represents a combination of \{black, white, others\} with \{female, male\}.  

We set $\ell(\cdot;\cdot)$ to be the logistic loss and utilize different methods to train a linear model. Table~\ref{tab:notation} lists the notation  for the algorithms referenced in this section. When we need to estimate the risk $R_i(\cdot)$, we draw a substantial number of samples from $\P_i$, and use the empirical average to approximate the expectation.

\subsection{GDRO on Balanced Data}
For experiments on the first synthetic data set, we will generate the random sample on the fly, according to the protocol in Section~\ref{sec:exp:setting}. For those on the Adult data set, we will randomly select samples from each group with replacement. In other words, $\P_i$ is defined as the empirical distribution over the data in the $i$-th group. 

In the experiments, we compare SMD(1) with our algorithms SMD($m$) and Online($1$). Fig.~\ref{fig:balance_risk} plots the maximum risk $\L_{\max}(\w)$, with respect to the number of iterations.  We observe that SMD($m$) is faster than Online($1$), which in turn outperforms SMD($1$). This observation is consistent with our theories, since their convergence rates are $O(\sqrt{(\log m)/T})$, $O(\sqrt{m (\log m)/T})$, and $O(m\sqrt{(\log m)/T})$, respectively. Next, we plot $\L_{\max}(\w)$ against the number of samples consumed by each algorithm in Fig.~\ref{fig:balance_sample}. As can be seen, the curves of SMD($m$) and Online($1$) are very close, indicating that they share the same sample complexity, i.e., $O(m (\log m)/\epsilon^2)$. On the other hand, SMD($1$) needs more samples to reach a target precision, which aligns with its higher sample complexity, i.e., $O(m^2 (\log m)/\epsilon^2)$.

\begin{figure}[t]
	\begin{center}
		\subfigure[Risk on $\P_1$]{
			\label{fig:imbalance_synth_1} %% label for second subfigure
			\includegraphics[width=0.31\textwidth]{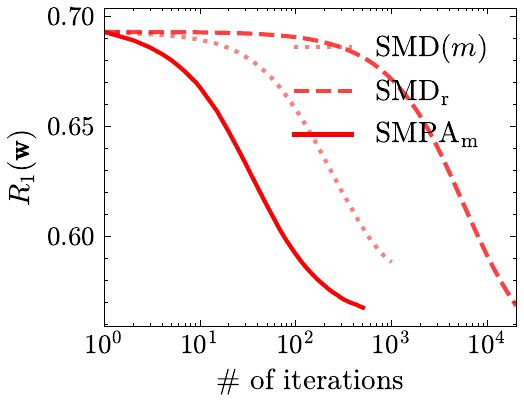}}%
		\subfigure[Risk on $\P_4$]{
			\label{fig:imbalance_synth_4} %% label for second subfigure
			\includegraphics[width=0.31\textwidth]{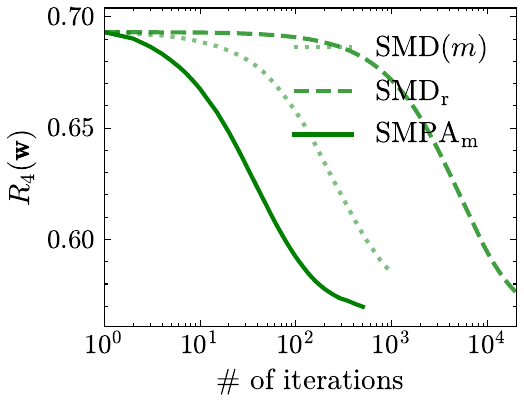}}%
		\subfigure[Risk on $\P_8$]{
			\label{fig:imbalance_synth_8} %% label for second subfigure
			\includegraphics[width=0.31\textwidth]{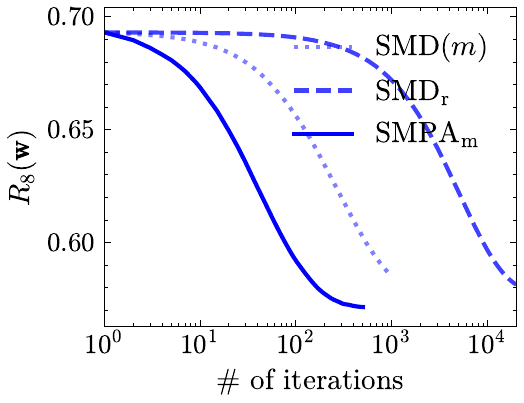}}\\%
		\subfigure[Risk on $\P_{12}$]{
			\label{fig:imbalance_synth_12} %% label for second subfigure
			\includegraphics[width=0.31\textwidth]{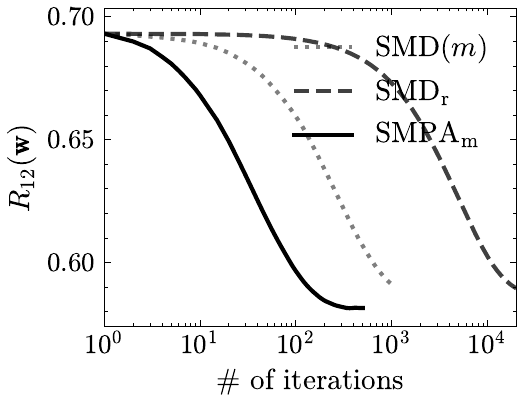}}%
		\subfigure[Risk on $\P_{16}$]{
			\label{fig:imbalance_synth_16} %% label for second subfigure
			\includegraphics[width=0.31\textwidth]{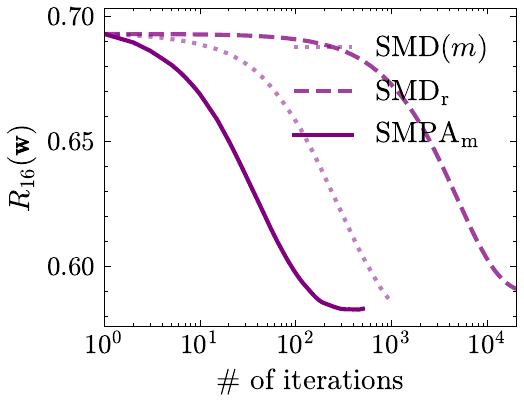}}%
		\subfigure[Risk on $\P_{20}$]{
			\label{fig:imbalance_synth_20} %% label for second subfigure
			\includegraphics[width=0.31\textwidth]{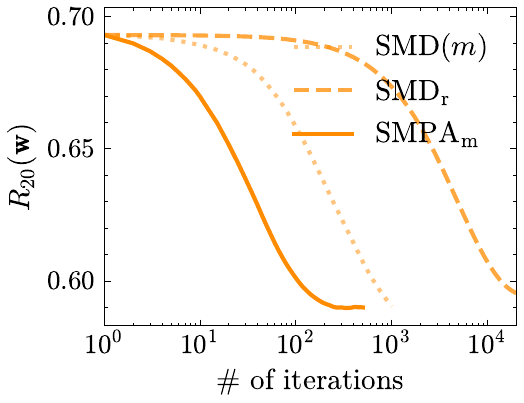}}\\%
		\caption{Imbalanced settings with the synthetic data set: individual risk versus the number of iterations.}
		\label{fig:imbalance_synth}\vspace{-2ex}
	\end{center}
\end{figure}

\subsection{Weighted GDRO on Imbalanced Data}
%For experiments on the first synthetic data set, we set the number of samples as $n_i = 1000\times (21-i)$, and generate each sample as before. 
For experiments on the first synthetic data set, we set the sample size for each group $i$ as $n_i=1000 \times (21-i)$. For those on the Adult data set, we first select $364$ samples randomly from each group,  reserving them for later use in estimating the risk of each group. Then, we visit the remaining samples in each group \emph{once} to simulate the imbalanced setting, where the numbers of samples in $6$ groups are $26656$, $11519$, $1780$, $1720$, $999$, and $364$. In this way, $\P_i$ corresponds to the (unknown) underlying distribution from which the samples in the $i$-th group are drawn.

\begin{figure}[t]
	\begin{center}
		\subfigure[Risk on $\P_1$]{
			\label{fig:imbalance_adult_1} %% label for second subfigure
			\includegraphics[width=0.31\textwidth]{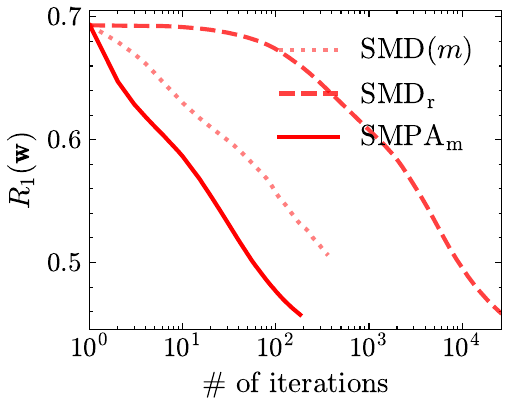}}%
		\subfigure[Risk on $\P_2$]{
			\label{fig:imbalance_adult_2} %% label for second subfigure
			\includegraphics[width=0.31\textwidth]{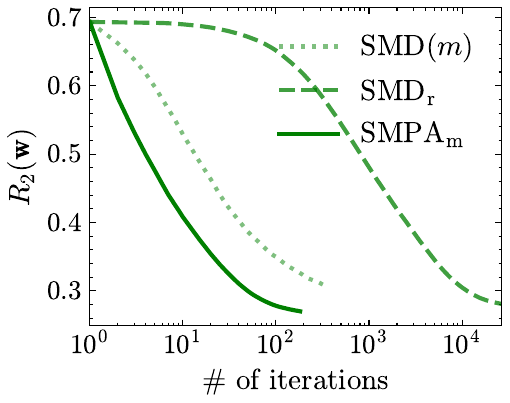}}%
		\subfigure[Risk on $\P_3$]{
			\label{fig:imbalance_adult_3} %% label for second subfigure
			\includegraphics[width=0.31\textwidth]{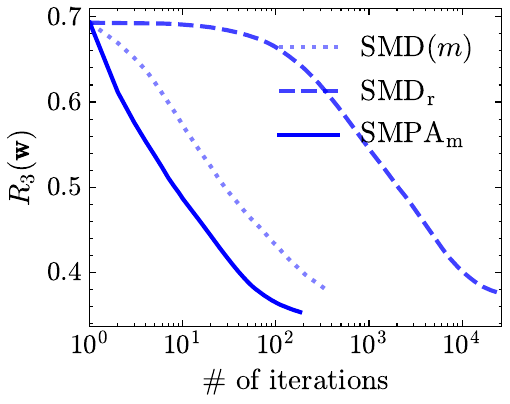}}\\%
		\subfigure[Risk on $\P_4$]{
			\label{fig:imbalance_adult_4} %% label for second subfigure
			\includegraphics[width=0.31\textwidth]{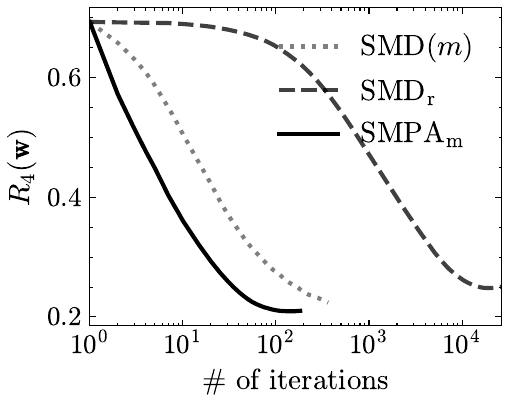}}%
		\subfigure[Risk on $\P_5$]{
			\label{fig:imbalance_adult_5} %% label for second subfigure
			\includegraphics[width=0.31\textwidth]{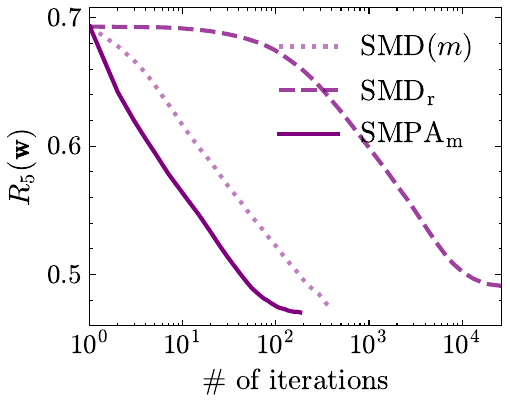}}%
		\subfigure[Risk on $\P_6$]{
			\label{fig:imbalance_adult_6} %% label for second subfigure
			\includegraphics[width=0.31\textwidth]{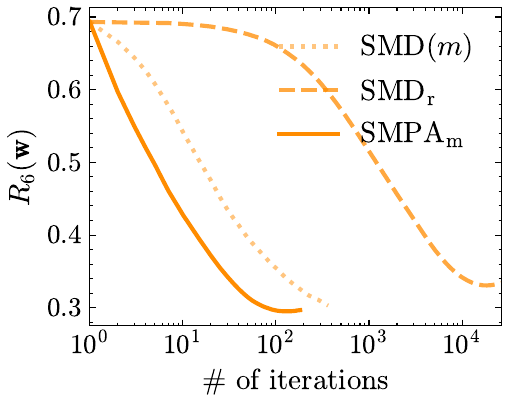}}\\%
		\caption{Imbalanced settings with the Adult data set: individual risk versus the number of iterations.}
		\label{fig:imbalance_adult}\vspace{-2ex}
	\end{center} 
\end{figure}
On imbalanced data, we will compare SMD$_\mathrm{r}$ and SMPA$_\mathrm{m}$ with the baseline SMD($m$), and examine how the risk on each individual distribution decreases with respect to the number of iterations. Recall that the total number of iterations of SMD$_\mathrm{r}$, SMPA$_\mathrm{m}$ and SMD($m$) are  $n_1$  $n_m/2$, and $n_m$, respectively. We present the experimental results on the synthetic and the Adult data sets in Fig.~\ref{fig:imbalance_synth} and Fig.~\ref{fig:imbalance_adult}, respectively. First, we observe that our SMPA$_\mathrm{m}$ is faster than both SMD$_\mathrm{r}$  and SMD($m$) across all distributions, and finally attains the lowest risk in most cases. This behavior aligns with our Theorem \ref{thm:7}, which reveals that SMPA$_\mathrm{m}$ achieves a nearly optimal rate of $O((\log m)/\sqrt{n_i})$ for all distributions $\P_i$, after $n_m/2$ iterations. We also note that on distribution $\P_1$, the distribution with the most samples, although SMD$_\mathrm{r}$ converges slowly, its final risk is the lowest, as illustrated in Fig.~\ref{fig:imbalance_synth_1} and Fig.~\ref{fig:imbalance_adult_1}. This phenomenon is again in accordance with our Theorem \ref{thm:5}, which shows that the risk of SMD$_\mathrm{r}$ on $\P_1$ reduces at a nearly optimal $O(\sqrt{(\log m)/n_1})$ rate, after $n_1$ iterations. From Fig.~\ref{fig:imbalance_synth_20} and Fig.~\ref{fig:imbalance_adult_6}, we can see that the final risk of SMD($m$) on the last distribution $\P_m$ matches that of SMPA$_\mathrm{m}$. This outcome is anticipated, as they exhibit similar convergence rates of $O(\sqrt{(\log m)/n_m})$ and $O((\log m)/\sqrt{n_m})$, respectively.

\begin{figure}[t]
	\begin{center}
		\subfigure[$\L_{k}(\w)$ versus the number of iterations]{
			\label{fig:heter_risk} %% label for second subfigure
			\includegraphics[width=0.4\textwidth]{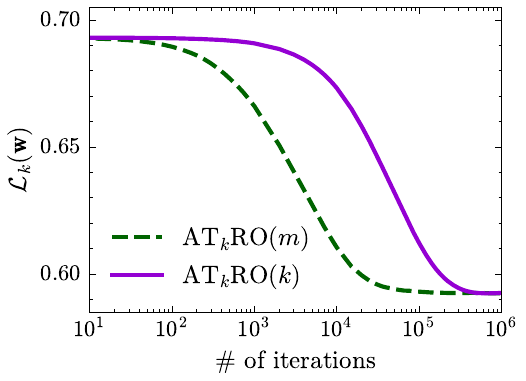}}%
		\quad
		\subfigure[$\L_{k}(\w)$ versus the number of samples]{
			\label{fig:heter_sample} %% label for second subfigure
			\includegraphics[width=0.4\textwidth]{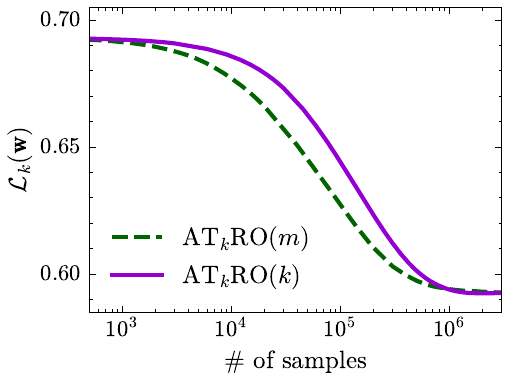}}\\
		\caption{Heterogeneous settings with the synthetic data set.}
		\label{fig:heter}
	\end{center}\vspace{-2ex}
	\begin{center}
		\subfigure[Risk on $\P_1$]{
			\label{fig:heter_1} %% label for second subfigure
			\includegraphics[width=0.31\textwidth]{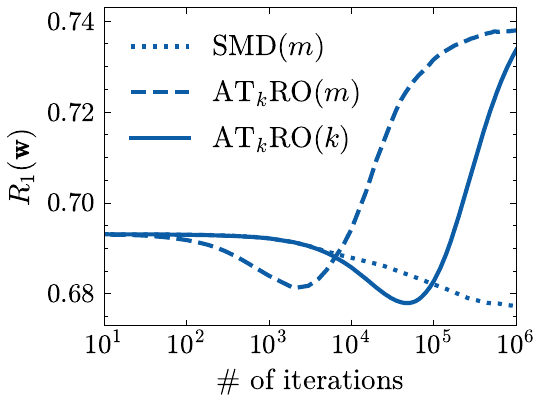}}%
		\subfigure[Risk on $\P_2$]{
			\label{fig:heter_2} %% label for second subfigure
			\includegraphics[width=0.31\textwidth]{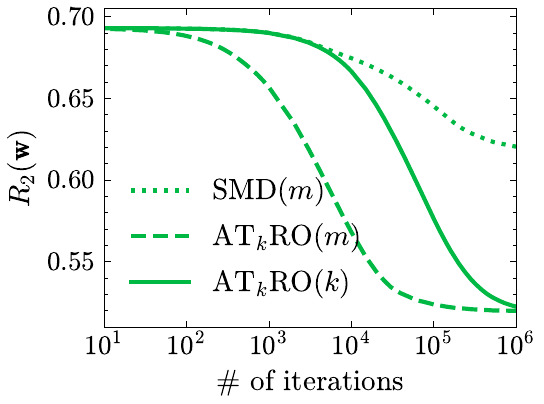}}
		\subfigure[Risk on $\P_5$]{
			\label{fig:heter_5} %% label for second subfigure
			\includegraphics[width=0.31\textwidth]{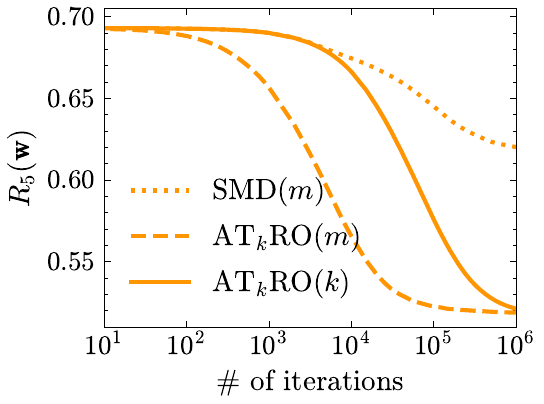}}\\
				\subfigure[Risk on $\P_{10}$]{
			\label{fig:heter_10} %% label for second subfigure
			\includegraphics[width=0.31\textwidth]{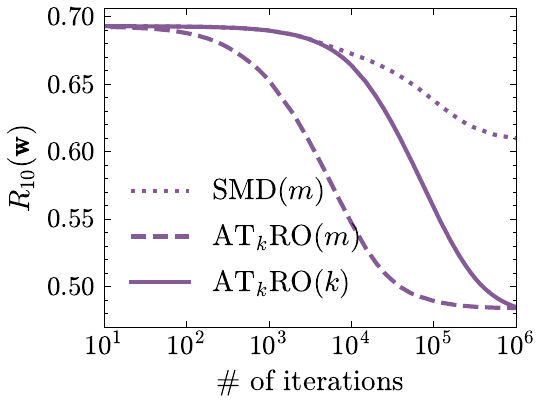}}%
		\subfigure[Risk on $\P_{15}$]{
			\label{fig:heter_15} %% label for second subfigure
			\includegraphics[width=0.31\textwidth]{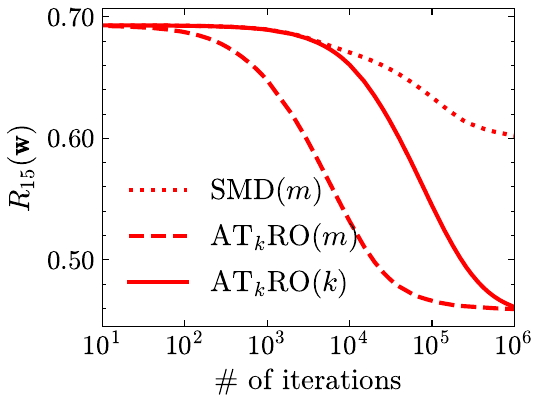}}
		\subfigure[Risk on $\P_{20}$]{
			\label{fig:heter_20} %% label for second subfigure
			\includegraphics[width=0.31\textwidth]{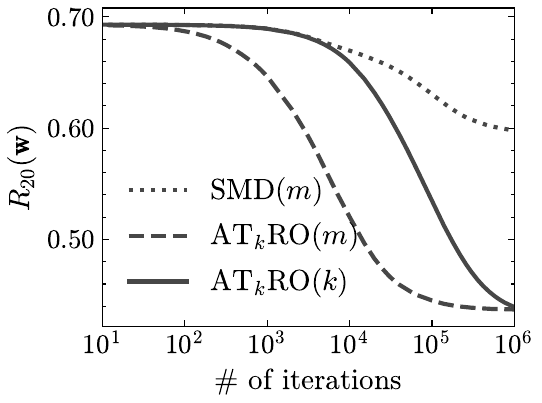}}\\
		\caption{Heterogeneous settings with the synthetic data set: individual risk versus the number of iterations.}
		\label{fig:heter_distributions}
	\end{center}\vspace{-2ex}
\end{figure}
\subsection{AT$_k$RO on Heterogeneous Distributions}
For experiments on heterogeneous distributions, we use the second synthetic data set described in Section~\ref{sec:exp:setting}. We first compare our two algorithms AT$_k$RO($m$) and AT$_k$RO($k$), where $k=3$, and plot the changes of the average top-$k$ risk $\L_{k}(\w)$ in Fig.~\ref{fig:heter}.  From Theorems~\ref{thm:8} and \ref{thm:10}, we know that their convergence rates are $O(\sqrt{(\log (m/k))/T})$ and $O(\sqrt{m(\log m)/(kT)})$, respectively, and their sample complexities are  $O((m \log (m/k))/\epsilon^2)$ and $O(m (\log m)/\epsilon^2)$, respectively. Fig.~\ref{fig:heter_risk} indicates that AT$_k$RO($m$) indeed converges faster than AT$_k$RO($k$), and Fig.~\ref{fig:heter_sample} shows that AT$_k$RO($m$) requires slightly fewer samples than AT$_k$RO($k$). 

Additionally, to demonstrate the advantages of AT$_k$RO, we examine the performance of directly applying the SMD($m$) algorithm, which is designed for GDRO, to the synthetic data set. Fig.~\ref{fig:heter_distributions} presents the changes in risk across a subset of distributions for SMD($m$), AT$_k$RO($m$), and AT$_k$RO($k$). We observe that SMD($m$) concentrates entirely on $\P_1$ and achieves the lowest final risk on the outlier distribution $\P_1$, approximately $0.061$ lower than
AT$_k$RO($m$) and $0.056$ lower than AT$_k$RO($k$). However, for the remaining $19$ distributions $\{\P_2,\ldots,\P_{20}\}$, the risk of SMD($m$) is approximately $0.12$ higher on average than those of the other two algorithms. Therefore, we conclude that AT$_k$RO can mitigate the impact of the outlier distribution and deliver a more balanced model in heterogeneous distributions compared to GDRO.

\subsection{Anytime Capability}
To demonstrate the benefits of the anytime capability, we compare SMD($m$) and Online(1) with their anytime extensions SMD($m$)$_\mathrm{a}$ and Online(1)$_\mathrm{a}$ on the Adult data set under balanced settings, and AT$_k$RO($m$) and AT$_k$RO$(k)$ with AT$_k$RO$(m)_\mathrm{a}$ and AT$_k$RO$(1)_\mathrm{a}$ on the second synthetic data set, where $k=3$.

\begin{figure}[t]
	\begin{center}
		\subfigure[The Adult data set]{
			%\label{fig:balance_adult_anytime_1_risk} %% label for second subfigure
			\includegraphics[width=0.23\textwidth]{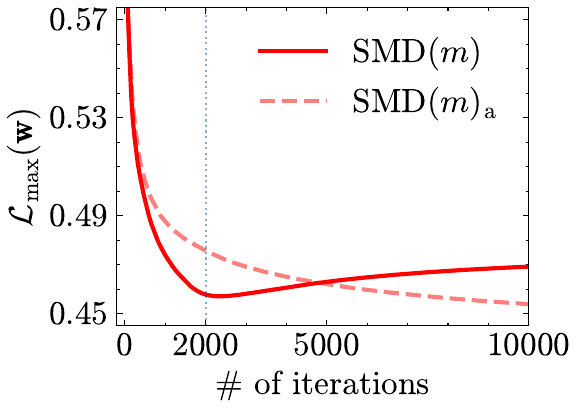}
		%\subfigure[The Adult data set]{
			%\label{fig:balance_adult_anytime_2_risk}
			\label{fig:balance_adult_anytime_risk} %% label for second subfigure
			\includegraphics[width=0.23\textwidth]{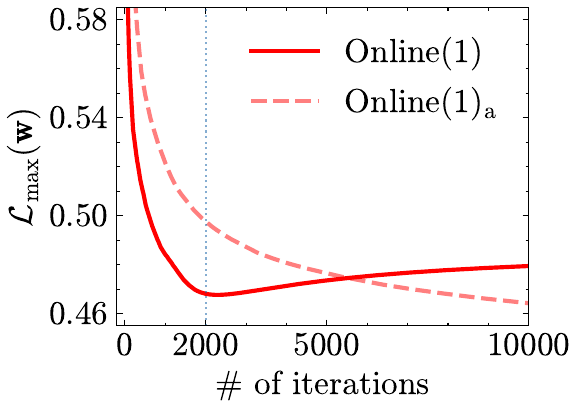}}
		\subfigure[The synthetic data set]{
			%\label{fig:ATkRO_anytime_risk} %% label for second subfigure
			\includegraphics[width=0.23\textwidth]{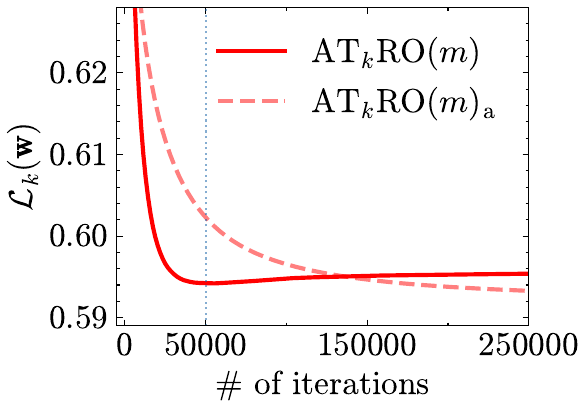}
		%\subfigure[Synth$_A$ data set]{
			\label{fig:ATkRO_anytime_risk} %% label for second subfigure
			\includegraphics[width=0.23\textwidth]{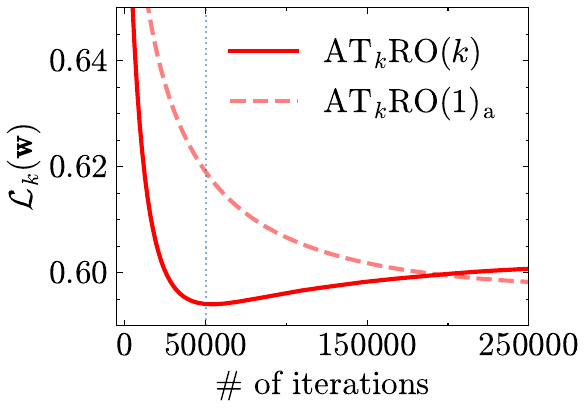}}
		\caption{The performance of different methods versus the number of iterations. Blue dashed lines indicate the predetermined $T$ for non-anytime algorithms.}
		\label{fig:experiment_anytime_set}
	\end{center}
\end{figure}

We assign a preset value of $T=2000$ for SMD($m$) and Online(1), and $T=50000$ for AT$_k$RO($m$) and AT$_k$RO($k$).  When the actual number of iterations exceeds the preset number $T$, we continue running the four algorithms with the initial parameters. As illustrated in Fig.~\ref{fig:experiment_anytime_set}, non-anytime algorithms initially reduce the objective  (the maximum risk or the average top-$k$ risk) more rapidly than anytime algorithms before reaching the predetermined $T$, where they achieve minimal  values. However, as the number of iterations increases, their curves  plateau or even increase due to sub-optimal parameters.  In contrast, the anytime extensions, with time-varying step sizes, consistently reduce their targets over time, eventually falling below the risk attained by the corresponding non-anytime algorithms.

%\section{Conclusion and Future Work}
\section{Conclusion}
For the GDRO problem, we develop two SA approaches based on SMD and non-oblivious MAB, which  consume $m$ and $1$ sample per round, respectively, and both achieve a nearly optimal sample complexity of $O(m (\log m)/\epsilon^2)$.  Then, we consider two special scenarios: imbalanced data and heterogeneous distributions. In the first scenario, we formulate a weighted GDRO problem and propose two methods by incorporating non-uniform sampling into SMD and using mini-batches with SMPA, respectively. These methods yield distribution-dependent convergence rates, and in particular, the latter one attains nearly optimal rates for multiple distributions simultaneously. In the second scenario, we formulate an AT$_k$RO problem and propose two algorithms: one using SMD with $m$ samples per round, obtaining an $O(m (\log (m/k))/\epsilon^2)$ sample complexity, and the other combining SMD with non-oblivious combinatorial semi-bandits, using $k$ samples per round and achieving an $O(m (\log m)/\epsilon^2)$ sample complexity. For both GDRO and AT$_k$RO, we have also developed SA algorithms with anytime capabilities.

%For future work, there are three interesting directions to explore: (i) extending the techniques of non-oblivious online learning to weighted GDRO to reduce the number of samples required in each iteration, thereby lowering the implementation threshold; (ii) adapting our SA algorithms to problems with similar formulations, such as collaborative PAC learning \citep{NIPS2017_186a157b} and federated learning~\citep{pmlr-v97-mohri19a}, aiming to broaden the applicability of our methods across diverse domains and settings; (iii) applying our algorithms to practical applications, such as optimizing the training of language models \citep{DRO:DoReMi}, with the goal of improving the performance and robustness of models in real-world scenarios.
%These avenues aim not only to facilitate the development of learning theory but also to provide effective and robust solutions for complex optimization challenges across various applications.

%\section*{Acknowledgements}
%\acks{Acknowledgements}
%All acknowledgements go at the end of the paper before appendices and references. Moreover, you are required to declare funding (financial activities supporting the submitted work) and competing interests (related financial activities outside the submitted work).

\appendix
\section{Supporting Lemmas} \label{appendix:Lemmas}
\subsection{Proof of Lemma~\ref{lemma:capped_simplex_opt}} \label{sec:Equivalence}
We first define $\hat{\q}$ as 
\begin{equation} \label{q_OMD_1}
\nabla\nu_q\left(\hat{\q}\right)=\nabla\nu_q(\q_0)-\eta \g.
\end{equation}
Then, we have
\begin{align*}
&\argmin_{\q \in \SS_{m,k}} B_q(\q,\hat{\q}) =\argmin_{\q \in \SS_{m,k}} \big\{\nu_q(\q)-\nu_q\left(\hat{\q}\right)-\langle\nabla\nu_q\left(\hat{\q}\right),\q-\hat{\q}\rangle\big\} \\
=&\argmin_{\q \in \SS_{m,k}} \big\{\nu_q(\q)-\langle\nabla\nu_q\left(\hat{\q}\right),\q\rangle\big\}
\overset{\eqref{q_OMD_1}}{=}\argmin_{\q \in \SS_{m,k}} \big\{\nu_q(\q)-\langle\nabla\nu_q(\q_0)-\eta \g,\q\rangle\big\} \\
=&\argmin_{\q \in \SS_{m,k}} \big\{ \eta \langle \g, \q -\q_0\rangle + B_c(\q,\q_0) \big\}=\eqref{q_OMD_0}.
\end{align*}

Recall that $B_q(\cdot,\cdot)$ is defined in terms of the neg-entropy, i.e., $\nu_q(\q)=\sum_{i=1}^{m}q_i\ln q_i$, we have $[\nabla \nu_q(\q)]_i=1+\ln q_i$. Therefore, the $i$-th component of $\hat{\q}$ can be computed as
\[
\hat{q}_{i}=\exp\big([\nabla\nu_q(\q_{0})]_i-\eta g_i-1\big)=\exp\left(\ln q_{0,i}-\eta g_i\right)=q_{0,i} e^{-\eta g_i}.
\]

\subsection{Proof of Lemma~\ref{lemma:high-prob-martingale}}
\label{sec:proof-concentration}
The proof follows the argument of~\citet[Proof of Lemma~1]{NIPS2015_e5a4d6bf}, and we generalize it to the setting with stochastic rewards. First, observe that for any $i \in [m]$ and $t \in [T]$,
\begin{align}
    \tilde{\xi}_{t,i} = {}& \frac{\hat{\xi}_{t,i}}{p_{t,i} + \gamma_t} \cdot \indicator{i_t = i} \nonumber \\
    \leq {}& \frac{\hat{\xi}_{t,i}}{p_{t,i} + \gamma_t \hat{\xi}_{t,i}} \cdot\indicator{i_t = i} \tag{$\hat{\xi}_{t,i} \in [0,1]$}\\
    = {}& \frac{1}{2\gamma_t} \frac{2\gamma_t \cdot \hat{\xi}_{t,i}/p_{t,i}}{1 + \gamma_t \cdot \hat{\xi}_{t,i}/p_{t,i}} \cdot \indicator{i_t = i}\nonumber \\
    \leq {}& \frac{1}{\beta_t} \log\left(1 + \beta_t \bar{\xi}_{t,i}\right) \label{eq:concentration-1}
\end{align}
where the last step is due to the inequality $\frac{z}{1 + z/2} \leq \log(1+z)$ for $z \geq 0$ and we introduce the notations $\beta_t = 2\gamma_t$ and $\bar{\xi}_{t,i} = (\hat{\xi}_{t,i}/p_{t,i}) \cdot \indicator{i_t = i}$ to simplify the presentation.

Define the notation $\tilde{\lambda}_t = \sum_{i=1}^m \alpha_{t,i} \tilde{\xi}_{t,i}$ and $\lambda_t = \sum_{i=1}^m \alpha_{t,i} \xi_{t,i}$. Then, we have
\begin{align}
    \E_{t-1}\left[\exp(\tilde{\lambda}_t)\right] = {}& \E_{t-1}\left[\exp\Big(\sum_{i=1}^m \alpha_{t,i} \tilde{\xi}_{t,i}\Big)\right] \nonumber \\
    \overset{\eqref{eq:concentration-1}}{\leq} {}& \E_{t-1}\left[\exp\left(\sum_{i=1}^m  \frac{\alpha_{t,i}}{\beta_t} \log\Big(1 + \beta_t \bar{\xi}_{t,i}\Big)\right)\right]\nonumber \\
    \leq {}& \E_{t-1}\left[\exp\left(\sum_{i=1}^m  \log\Big(1 + \alpha_{t,i} \bar{\xi}_{t,i}\Big)\right)\right] \tag*{($\frac{\alpha_{t,i}}{\beta_t} \leq 1$ by assumption)} \\
         ={}& \E_{t-1}\left[\Pi_{i=1}^m \big(1 + \alpha_{t,i} \bar{\xi}_{t,i}\big)\right] = \E_{t-1}\left[1 + \sum_{i=1}^m \alpha_{t,i} \bar{\xi}_{t,i}\right] \nonumber 
    \end{align}
 \begin{align}  
    ={}& 1 + \sum_{i=1}^m \alpha_{t,i} \xi_{t,i}     \leq \exp\left(\sum_{i=1}^m \alpha_{t,i} \xi_{t,i} \right)  = \exp(\lambda_t) \label{eq:concentration-2}
\end{align}
where the second inequality is by the inequality $x \log(1+y) \leq \log(1+ xy)$ that holds for all $y \geq -1$ and $x \in [0,1]$,  the equality $\E_{t-1}\left[\Pi_{i=1}^m \big(1 + \alpha_{t,i} \bar{\xi}_{t,i}\big)\right] = \E_{t-1}\left[1 + \sum_{i=1}^m \alpha_{t,i} \bar{\xi}_{t,i}\right]$ follows from the fact that $\bar{\xi}_{t,i} \cdot \bar{\xi}_{t,j} = 0$ holds whenever $i \neq j$, and the last line is due to $\E_{t-1}[\bar{\xi}_{t,i}] = \E_{t-1}[(\hat{\xi}_{t,i}/p_{t,i}) \cdot \indicator{i_t = i}] = \xi_{t,i}$ and the inequality $1+ z \leq e^z$ for all $z \in \R$.

As a result, from~\eqref{eq:concentration-2} we conclude that the process $Z_t = \exp\big(\sum_{s=1}^t (\tilde{\lambda}_s - \lambda_s)\big)$ is a supermartingale. Indeed, $\E_{t-1}[Z_t] = \E_{t-1}\big[\exp\big(\sum_{s=1}^{t-1} (\tilde{\lambda}_s - \lambda_s)\big) \cdot \exp(\tilde{\lambda}_t - \lambda_t)\big] \leq Z_{t-1}$. Thus, we have $\E[Z_T] \leq \E[Z_{T-1} \leq \ldots \leq \E[Z_{0}] = 1$. By Markov's inequality,
\[
    \Pr\left[\sum_{t=1}^T (\tilde{\lambda}_t - \lambda_t) > \epsilon\right] \leq \E\left[\exp\left( \sum_{t=1}^T (\tilde{\lambda}_t - \lambda_t) \right)\right] \cdot \exp(-\epsilon) \leq \exp(-\epsilon)
\]
holds for any $\epsilon >0$. By setting $\exp(-\epsilon) = \delta$ and solving the value, we complete the proof for~\eqref{eq:high-prob-margingale}. And the inequality~\eqref{eq:high-prob-margingale2} for the scenario $\gamma_t = \gamma$ can be immediately obtained by setting $\alpha_{t,i} = 2\gamma \cdot \indicator{i = j}$ and taking the union bound over all $j\in [m]$.

\subsection{Proof of Lemma~\ref{lem:1}}
From the definition of norms in (\ref{eqn:norm:new}), we have
\[
\begin{split}
&\| F([\w;\q]) -F([\w';\q'])\|_*^2 \\
= & \left \|\left[\sum_{i=1}^m q_i p_i \nabla R_i(\w) - \sum_{i=1}^m q_i' p_i \nabla R_i(\w') ; \right.\right. \\
  & \left.\left. \big[ p_1 R_1(\w')-p_1 R_1(\w), \ldots, p_m R_m(\w')-p_m R_m(\w) \big]^\top\right] \right \|_*^2\\
=& 2 \Dw^2 \left\|\sum_{i=1}^m q_i p_i \nabla R_i(\w) - \sum_{i=1}^m q_i' p_i \nabla R_i(\w') \right\|_{w,*}^2 \\
 &+  2 \left\|\big[ p_1 R_1(\w')-p_1 R_1(\w), \ldots, p_m R_m(\w')-p_m R_m(\w) \big]^\top \right\|_\infty^2   \ln m \\
= & 2 \Dw^2 \left\|\sum_{i=1}^m q_i p_i \nabla R_i(\w) -\sum_{i=1}^m q_i' p_i \nabla R_i(\w) +\sum_{i=1}^m q_i' p_i \nabla R_i(\w) - \sum_{i=1}^m q_i' p_i \nabla R_i(\w') \right\|_{w,*}^2\\
 &+  2 \left\|\big[ p_1 R_1(\w')-p_1 R_1(\w), \ldots, p_m R_m(\w')-p_m R_m(\w) \big]^\top \right\|_\infty^2   \ln m \\
\leq & \underbrace{4 \Dw^2 \left\|\sum_{i=1}^m q_i p_i \nabla R_i(\w) -\sum_{i=1}^m q_i' p_i \nabla R_i(\w)\right\|_{w,*}^2}_{:=A} \\
& + \underbrace{4 \Dw^2 \left\|\sum_{i=1}^m q_i' p_i \nabla R_i(\w) - \sum_{i=1}^m q_i' p_i \nabla R_i(\w') \right\|_{w,*}^2}_{:=B} + \underbrace{2  \max_{i\in[m]} \left|p_i \big[R_i(\w)-R_i(\w')\big] \right|^2 \ln m}_{:=C}.
\end{split}
\]

To bound term $A$, we have
\[
\begin{split}
& 4 \Dw^2 \left\|\sum_{i=1}^m q_i p_i \nabla R_i(\w) -\sum_{i=1}^m q_i' p_i \nabla R_i(\w)\right\|_{w,*}^2  \\
\leq & 4 \Dw^2 \left( \sum_{i=1}^m |q_i-q_i'| \|p_i \nabla R_i(\w)\|_{w,*}\right)^2 \overset{\text{(\ref{eqn:R:Lipschitz:1})}}{\leq} 4 \Dw^2 \left( \sum_{i=1}^m |q_i-q_i'| p_i G \right)^2 \leq 4 \Dw^2  G^2 p_{\max}^2 \|\q -\q'\|_1^2.
\end{split}
\]
where $p_{\max}$ is defined in (\ref{eqn:mirror:parameters}). To bound $B$, we have
\[
\begin{split}
& 4 \Dw^2 \left\|\sum_{i=1}^m q_i' p_i \nabla R_i(\w) - \sum_{i=1}^m q_i' p_i \nabla R_i(\w') \right\|_{w,*}^2\\
\leq & 4 \Dw^2  \left( \sum_{i=1}^m q_i' p_i \left\|\nabla R_i(\w)- \nabla R_i(\w')\right\|_{w,*} \right)^2 \overset{\text{(\ref{eqn:smooth:R})}}{\leq} 4 \Dw^2  \left( \sum_{i=1}^m q_i' p_i L \|\w-\w'\|_w \right)^2 \\
\leq &  4 \Dw^2 L^2 p_{\max}^2 \|\w-\w'\|_w^2  \left( \sum_{i=1}^m q_i'  \right)^2 =4 \Dw^2 L^2  p_{\max}^2 \|\w-\w'\|_w^2 .
\end{split}
\]
To bound $C$, we have
\[
\begin{split}
& 2  \max_{i\in[m]} \left|p_i \big[R_i(\w)-R_i(\w')\big] \right|^2 \ln m \\
\overset{\text{(\ref{eqn:R:Lipschitz:2})}}{\leq}& 2  \max_{i\in[m]} \left|p_i G \|\w-\w'\|_w\right|^2 \ln m \leq 2 G^2 p_{\max}^2 \|\w-\w'\|_w^2 \ln m.
\end{split}
\]

Putting everything together, we have
\[
\begin{split}
&\| F([\w;\q]) -F([\w';\q'])\|_*^2 \leq (4 \Dw^2 L^2  +  2 G^2 \ln m) p_{\max}^2 \|\w-\w'\|_w^2 +  4 \Dw^2  G^2 p_{\max}^2 \|\q -\q'\|_1^2  \\
\leq & p_{\max}^2  (  8 \Dw^4 L^2  +   8 \Dw^2  G^2  \ln m ) \left(  \frac{1}{2 \Dw^2} \|\w-\w'\|_{w}^2 + \frac{1}{2 \ln m} \|\q-\q'\|_1^2 \right) \\
=& p_{\max}^2  (  8 \Dw^4 L^2  +   8 \Dw^2  G^2  \ln m ) \big\|[\w-\w';\q-\q'] \big\|^2
\end{split}
\]
which implies
\[
\begin{split}
\| F([\w;\q]) -F([\w';\q'])\|_* 
\leq & p_{\max} \sqrt{8 \Dw^4 L^2  +   8 \Dw^2  G^2  \ln m} \big\|[\w-\w';\q-\q'] \big\| \\
\leq & \Lt \big\|[\w-\w';\q-\q'] \big\|
\end{split}
\]
where $\Lt$ is defined in (\ref{eqn:mirror:parameters}).

\subsection{Proof of Lemma~\ref{lem:2}}
The light tail condition, required by \citet{Nemirovski:SMP}, is essentially the sub-Gaussian condition. To this end, we introduce the following sub-gaussian properties \citep[Proposition 2.5.2]{HDP:Vershynin}.
\begin{prop}[Sub-gaussian properties] \label{pro:subgaussian}
Let $X$ be a random variable. Then the following properties are equivalent; the parameters $K_i > 0$ appearing in these properties differ
from each other by at most an absolute constant factor.
\begin{compactenum}[(i)]
\item The tails of $X$ satisfy
\[
\Pr[|X| \geq t] \leq 2 \exp(-t^2/K_1^2), \ \forall t \geq 0.
\]
\item The moments of $X$ satisfy
\[
\|X\|_{L_p} = \left( \E |X|^p\right)^{1/p} \leq K_2 \sqrt{p}, \ \forall p \geq 1.
\]
\item The moment generating function (MGF) of $X^2$ satisfies
\[
\E \big[\exp(\lambda^2 X^2) \big] \leq \exp(K_3^2 \lambda^2), \ \forall \lambda \textrm{ such that }|\lambda| \leq 1/K_3.
\]
\item The MGF of  $X^2$ is bounded at some point, namely
\[
\E \big[ \exp(X^2/K_4^2) \big]\leq 2.
\]
%Moreover, if $\E [X] = 0$ then properties (i)--(iv) are also equivalent to the following property.
%\item The MGF of X satisfies
%\[
%\E \big[\exp(\lambda X)\big] \leq \exp(K_5^2 \lambda^2), \ \forall \lambda \in \R.
%\]
\end{compactenum}
\end{prop}
From the above proposition, we observe that the exact value of those constant $K_1,\ldots,K_5$ is not important, and it is very tedious to calculate them. So, in the following, we only focus on the order of those constants. To simplify presentations, we use $c$ to denote an absolute constant that is independent of all the essential parameters, and its value may change from line to line.

Since
\[
\begin{split}
&\| F([\w;\q]) -\g([\w;\q])\|_*^2 \\
 =& 2 \Dw^2 \|\nabla_\w \varphi(\w,\q)-\g_w(\w,\q)\|_{w,*}^2 +  2 \|\nabla_\q \varphi(\w,\q)-\g_q(\w,\q)\|_\infty^2  \ln m,
\end{split}
\]
we proceed to analyze the behavior of $\|\nabla_\w \varphi(\w,\q)-\g_w(\w,\q)\|_{w,*}^2$ and $\|\nabla_\q \varphi(\w,\q)-\g_q(\w,\q)\|_\infty^2$. To this end, we have the following lemma.

\begin{lemma} \label{lem:erro}
We have
\begin{equation} \label{eqn:variance:two:norm}
\begin{split}
&\E  \bigg[\exp \left(  \frac{1}{c  \kappa G^2 \omega_{\max}} \|\nabla_\w \varphi(\w,\q)-\g_w(\w,\q)\|_{w,*}^2  \right)\bigg] \leq  2,\\
&\E \left[\exp\left( \frac{1}{c \omega_{\max}  \ln m} \|\nabla_\q \varphi(\w,\q)-\g_q(\w,\q)\|_\infty^2\right) \right] \leq 2
\end{split}
\end{equation}
where $\omega_{\max}$ is defined in (\ref{eqn:mirror:parameters}) and $c>0$ is an absolute constant.
\end{lemma}
From Lemma~\ref{lem:erro}, we have
\[
\begin{split}
& \E \left[\exp \left( \frac{1}{2 c \kappa \Dw^2  G^2 \omega_{\max} + 2 c \omega_{\max}  \ln^2 m } \| F([\w;\q]) -\g([\w,\q])\|_*^2 \right) \right] \\
=& \E \left[\exp \left( \frac{2 \Dw^2 }{2 c \kappa \Dw^2  G^2 \omega_{\max} + 2 c \omega_{\max}  \ln^2 m} \|\nabla_\w \varphi(\w,\q)-\g_w(\w,\q)\|_{w,*}^2 \right. \right.\\
& \left. \left.+ \frac{ 2 \ln m}{  2 c \kappa \Dw^2  G^2 \omega_{\max} + 2 c \omega_{\max}  \ln^2 m}  \|\nabla_\q \varphi(\w,\q)-\g_q(\w,\q)\|_\infty^2   \right)\right]\\
=& \E \left[\exp \left( \frac{\kappa \Dw^2    G^2 }{ \kappa \Dw^2  G^2  +    \ln^2 m} \frac{\|\nabla_\w\varphi(\w,\q)-\g_w(\w,\q)\|_{w,*}^2}{c  \kappa G^2 \omega_{\max}}  \right. \right.\\
& \left. \left.+ \frac{ \ln^2 m}{  \kappa \Dw^2  G^2  +   \ln^2 m }  \frac{\|\nabla_\q \varphi(\w,\q)-\g_q(\w,\q)\|_\infty^2}{c \omega_{\max}  \ln m}    \right)\right]\\
\end{split}
\]
\[
\begin{split}
\leq &  \frac{ \kappa \Dw^2    G^2 }{ \kappa \Dw^2  G^2  +    \ln^2 m} \E  \left[\exp \left(  \frac{\|\nabla_\w\varphi(\w,\q)-\g_w(\w,\q)\|_{w,*}^2}{c  \kappa G^2 \omega_{\max}}     \right)\right] \\
&+ \frac{ \ln^2 m}{  \kappa \Dw^2  G^2  +   \ln^2 m }\E \left[ \exp \left( \frac{\|\nabla_\q \varphi(\w,\q)-\g_q(\w,\q)\|_\infty^2}{c \omega_{\max}  \ln m}    \right)\right]\\
\overset{\text{(\ref{eqn:variance:two:norm})}}{\leq} & \frac{ \kappa \Dw^2    G^2 }{ \kappa \Dw^2  G^2  +    \ln^2 m} 2 + \frac{ \ln^2 m}{ \kappa \Dw^2  G^2  +   \ln^2 m } 2 =2
\end{split}
\]
where the first inequality follows from Jensen's inequality.

\subsection{Proof of Lemma~\ref{lemma:high-prob-martingale-2}}
The proof is built upon that of Corollary 1 of  \citet{NIPS2015_e5a4d6bf}. 
Let $\beta=2\gamma$ and $\xi_{t,i}^{\prime}=\frac{\hat{\xi}_{t,i}}{kp_{t,i}}\mathbb{I}[i\in \I_t]$. First, we have
\begin{equation} \label{eqn:cor1:1}
\begin{split}
\tilde{\xi}_{t, i}=&\frac{\hat{\xi}_{t,i}}{kp_{t,i}+\gamma} \ind[i\in \I_t] \\
	\leq &\frac{\hat{\xi}_{t,i}}{kp_{t,i}+\gamma \hat{\xi}_{t,i}} \ind[i\in \I_t]=\frac{1}{2 \gamma} \cdot \frac{2 \gamma \hat{\xi}_{t,i} / kp_{t,i}}{1+\gamma \hat{\xi}_{t,i} / kp_{t, i}} \ind[i\in \I_t]  \leq \frac{1}{\beta} \cdot \log \left(1+\beta \xi_{t,i}^{\prime}\right)
\end{split}
\end{equation}
where the first inequality follows from $\hat{\xi}_{t, i} \in[0,1]$ and last inequality from the elementary inequality $\frac{z}{1+z / 2} \leq$ $\log (1+z)$ that holds for all $z \geq 0$. Second, from the property of DepRound, we have
\begin{equation} \label{eqn:cor1:2}
\E_{t-1}[\xi_{t,i}^{\prime}] =\E_{t-1}\left[\frac{\xi_{t,i}}{kp_{t,i}}\mathbb{I}[i\in \I_t]\right] \overset{\text{(\ref{DepRound:prop})}}{=} \xi_{t,i}.
\end{equation}
Then, we have
\[
\E_{t-1}\left[ \exp(\beta \tilde{\xi}_{t, i}) \right] \overset{\text{(\ref{eqn:cor1:1})}}{\leq} \E_{t-1}\left[ 1+\beta \xi_{t,i}^{\prime} \right] \overset{\text{(\ref{eqn:cor1:2})}}{=} 1 + \beta \xi_{t,i} \leq \exp(\beta \xi_{t,i}).
\]
Then, by repeating the subsequent analysis from Corollary 1 of \citet{NIPS2015_e5a4d6bf}, we can derive this lemma.

\subsection{Proof of Lemma~\ref{lem:erro}}
To analyze $\|\nabla_\w \varphi(\w,\q)-\g_w(\w,\q)\|_{w,*}^2$, we first consider the approximation error caused by samples from $\P_i$:
\[
\left\|\frac{n_m}{n_i} \sum_{j=1}^{n_i/n_m} \nabla \ell(\w;\z^{(i,j)})- \nabla R_i(\w) \right  \|_{w,*} = \left\|\frac{n_m}{n_i} \sum_{j=1}^{n_i/n_m} \left[ \nabla \ell(\w;\z^{(i,j)})- \nabla R_i(\w) \right] \right  \|_{w,*}.
\]
Under the regularity condition of $\|\cdot\|_{w,*}$ in Assumption~\ref{ass:5}, we have, for any $\gamma \geq 0$,
\begin{equation}\label{eqn:con:norm:1}
\left \|\frac{n_m}{n_i} \sum_{j=1}^{n_i/n_m} \left[ \nabla \ell(\w;\z^{(i,j)})- \nabla R_i(\w) \right] \right\|_{w,*} \geq 2G ( \sqrt{2\kappa} + \sqrt{2} \gamma ) \sqrt{\frac{n_m}{n_i}} \leq \exp(-\gamma^2/2)
\end{equation}
which is a directly consequence of the concentration inequality of vector norms \citep[Theorem 2.1.(iii)]{Vector:Martingales} and (\ref{eqn:error:RGrad}).  Then, we introduce the following lemma to simplify (\ref{eqn:con:norm:1}).
\begin{lemma} \label{lem:sug:gau}
Suppose we have
\[
\Pr\left[ X \geq \alpha + \gamma \right] \leq \exp({-} \gamma^2/2), \ \forall \gamma>0
\]
where $X$ is nonnegative. Then, we have
\[
\Pr\left[ X \geq \gamma \right]  \leq 2 \exp\big({-} \gamma^2/\max(6\alpha^2,8)\big), \ \forall \gamma>0.
\]
\end{lemma}
From  (\ref{eqn:con:norm:1}) and Lemma~\ref{lem:sug:gau}, we have
\[
\begin{split}
& \Pr\left[\frac{1}{2\sqrt{2} G} \sqrt{\frac{n_i}{n_m}} \left \|\frac{n_m}{n_i} \sum_{j=1}^{n_i/n_m} \left[ \nabla \ell(\w;\z^{(i,j)})- \nabla R_i(\w) \right] \right\|_{w,*} \geq \gamma \right] \\
\leq & 2 \exp\big({-} \gamma^2/\max(6\kappa,8)\big) \leq  2 \exp\big({-} \gamma^2/(8\kappa)\big), \ \forall \gamma>0
\end{split}
\]
which satisfies the Proposition~\ref{pro:subgaussian}.(i). From the equivalence between Proposition~\ref{pro:subgaussian}.(i) and Proposition~\ref{pro:subgaussian}.(iv), we have
\[
\E \left[\exp\left(\left.\left \|\frac{n_m}{n_i} \sum_{j=1}^{n_i/n_m} \left[ \nabla \ell(\w;\z^{(i,j)})- \nabla R_i(\w) \right] \right\|_{w,*}^2 \right/ \frac{c \kappa G^2 n_m}{n_i}\right) \right] \leq 2.
\]
Inserting the scaling factor $p_i$, we have
\begin{equation}\label{eqn:con:norm:2}
\E \left[\exp\left(\left.\left \|p_i \frac{n_m}{n_i} \sum_{j=1}^{n_i/n_m} \left[ \nabla \ell(\w;\z^{(i,j)})- \nabla R_i(\w) \right] \right\|_{w,*}^2 \right/  \frac{c  \kappa G^2 p_i^2 n_m}{n_i}\right) \right]\leq 2.
\end{equation}

To simplify the notation, we define
\[
\u_i=p_i \frac{n_m}{n_i} \sum_{j=1}^{n_i/n_m} \left[ \nabla \ell(\w;\z^{(i,j)})- \nabla R_i(\w) \right], \textrm{ and }\omega_{\max}=  \max_{i \in [m]}  \frac{p_i^2 n_m}{n_i}.
\]
By Jensen's inequality, we have
\[
\begin{split}
&\E \left[\exp \left( \frac{1}{c  \kappa G^2 \omega_{\max}} \|\nabla_\w \varphi(\w,\q)-\g_w(\w,\q)\|_{w,*}^2 \right)\right]\\
=& \E \left[ \exp \left( \left. \left\|\sum_{i=1}^m q_{i} \u_i \right\|_{w,*}^2 \right/ \big[c  \kappa G^2 \omega_{\max}\big]\right)\right] \\
\leq &  \sum_{i=1}^m q_{i} \E \left[ \exp \left( \left. \left\| \u_i \right\|_{w,*}^2 \right/ \big[c  \kappa G^2 \omega_{\max}\big]\right)\right] \overset{\text{(\ref{eqn:con:norm:2})}}{\leq} \sum_{i=1}^m q_{i} 2 = 2.
\end{split}
\]
where we use the fact that $\|\cdot\|_{w,*}$, $(\cdot)^2$ and $\exp(\cdot)$ are convex, and the last two functions are increasing in $\R_+$.

To analyze $\|\nabla_\q \varphi(\w,\q)-\g_q(\w,\q)\|_\infty^2$, we consider the approximation error related to $\P_i$:
\[
\left|\frac{n_m}{n_i} \sum_{j=1}^{n_i/n_m}  \ell(\w;\z^{(i,j)})-  R_i(\w)  \right| = \left|\frac{n_m}{n_i} \sum_{j=1}^{n_i/n_m} \left[ \ell(\w;\z^{(i,j)})-  R_i(\w) \right]  \right|.
\]
Note that the absolute value $|\cdot|$ is $1$-regular \citep{Vector:Martingales}. Following (\ref{eqn:error:R}) and  the derivation of (\ref{eqn:con:norm:2}), we have
\begin{equation}\label{eqn:con:norm:3}
\E\left[ \exp\left(\left.\left |p_i \frac{n_m}{n_i} \sum_{j=1}^{n_i/n_m} \left[  \ell(\w;\z^{(i,j)})-  R_i(\w) \right] \right|^2  \right/  \frac{c   p_i^2 n_m} {n_i}\right) \right]\leq 2.
\end{equation}
To prove that $\|\nabla_\q \varphi(\w,\q)-\g_q(\w,\q)\|_\infty^2$ is also sub-Gaussian, we need to analyze the effect of the infinity norm. To this end, we develop the following lemma.
\begin{lemma} \label{lem:max:Gaus} Suppose
\begin{equation} \label{eqn:lem:max:1}
\E \left[\exp\left(|X_j|^2 / K_j^2\right) \right]\leq 2, \ \forall j \in [m].
\end{equation}
Then,
\[
\E \left[\exp\left(\left. \max_{j\in[m]} |X_j|^2\right/ \big[c K_{\max}^2 \ln m\big]\right)\right] \leq 2.
\]
where $c>0$ is an absolute constant, and $K_{\max}=\max_{j \in [m]} K_j$.
\end{lemma}
From (\ref{eqn:con:norm:3}) and Lemma~\ref{lem:max:Gaus}, we have
\[
\E \left[\exp\left(\frac{1}{c \omega_{\max}  \ln m} \|\nabla_\q \varphi(\w,\q)-\g_q(\w,\q)\|_\infty^2 \right)\right] \leq 2.
\]

\subsection{Proof of Lemma~\ref{lem:sug:gau}}
When $\gamma \in [0,2 \alpha]$, we have
\[
\Pr\left[ X \geq \gamma \right] \leq 1 \leq 2 \exp(-2/3) \leq  2 \exp({-} \gamma^2/6\alpha^2).
\]
When $\gamma \geq 2\alpha$, we have
\[
\Pr\left[ X \geq \gamma \right] = \Pr\left[ X \geq \alpha + \gamma-\alpha \right] \leq \exp({-} (\gamma-\alpha)^2/2) \leq  \exp({-} \gamma^2/8)
\]
where we use the fact $\gamma-\alpha \geq \frac{\gamma}{2}$. Thus, we always have
\[
\Pr\left[ X \geq \gamma \right]  \leq 2 \exp\big({-}\gamma^2/\max(6\alpha^2,8)\big), \ \forall \gamma>0.
\]

\subsection{Proof of Lemma~\ref{lem:max:Gaus}}
From (\ref{eqn:lem:max:1}), and the equivalence between Proposition~\ref{pro:subgaussian}.(i) and Proposition~\ref{pro:subgaussian}.(iv), we have
\[
\Pr\big[|X_j| \geq t\big] \leq 2 \exp\big({-}t^2/c K_j^2\big), \ \forall t \geq 0, \forall j \in [m].
\]
As a result,
\[
\begin{split}
& \Pr\left[\max_{j\in[m]} |X_j| \geq t\right] =\Pr\big[ \exists j, |X_j| \geq t\big] \leq \sum_{j=1}^m \Pr\big[|X_j| \geq t\big]\leq 2 \sum_{j=1}^m \exp\big({-}t^2/c K_j^2\big) \\
\leq & 2 m \exp\big({-}t^2/c K_{\max}^2\big) = \exp \big({-}t^2/c K_{\max}^2 + \ln [2m] \big).
\end{split}
\]
Choosing $t=\sqrt{c K_{\max}^2 (\ln [2m] + \gamma^2/2)}$, we have
\[
\Pr\left[\max_{j\in[m]} |X_j| \geq \sqrt{c K_{\max}^2 (\ln [2m] + \gamma^2/2)}\right] \leq  \exp \left(-\gamma^2/2 \right).
\]
Thus
\[
\begin{split}
&\Pr\left[\max_{j\in[m]} |X_j| \geq \sqrt{c K_{\max}^2} \left( \sqrt{\ln [2m]} + \gamma/\sqrt{2} \right)\right] \leq \exp\left(-\gamma^2/2 \right)\\
 \Leftrightarrow &\Pr\left[ \sqrt{\frac{2}{c K_{\max}^2}} \max_{j\in[m]} |X_j| \geq   \sqrt{2 \ln [2m]} + \gamma \right] \leq \exp\left(-\gamma^2/2 \right).
\end{split}
\]
By Lemma~\ref{lem:sug:gau}, we have
\[
\Pr\left[ \sqrt{\frac{2}{c K_{\max}^2}} \max_{j\in[m]} |X_j| \geq \gamma \right]  \leq 2 \exp\Big({-} \gamma^2/\max\big(12 \cdot \ln [2m],8\big)\Big), \ \forall \gamma>0.
\]
From the equivalence between Proposition~\ref{pro:subgaussian}.(i) and Proposition~\ref{pro:subgaussian}.(iv), we have
\[
\E \left[ \exp\left(\left. \max_{j\in[m]} |X_j|^2 \right/ \big[c K_{\max}^2 \ln m\big]\right) \right] \leq 2.
\]
\section{Details of Example~\ref{example:GDRO&ATk}} \label{sec:prof:example}
According to our constructions, the $i$-th risk function is given by
\[
R_i(\w)=\E_{\z \sim \operatorname{Ber}(\mu_i,1)}\big[(\w-\z)^2\big]=\w^2-2\mu_i\w+\mu_i.
\]

We first derive the objective of GDRO, i.e., $\L_{\text{max}}(\w)$  in \eqref{eqn:Max_risk}.  When $\w\in[0,0.5]$, $\L_{\text{max}}(\w)=\w^2-2\w+1$ with corresponding $\mu_i=1$, and when $\w\in [0.5,1]$, $\L_{\text{max}}(\w)=\w^2-\w+0.5$ with corresponding $\mu_i=0.5$. Then, it can be easily verified from Fig.~\ref{fig:Example_objective} that  $\w^*_G=\argmin_{\w\in\W} \L_{\text{max}}(\w)=0.5$. 

Then, we analyze the objective of AT$_5$RO, i.e., $\L_5(\w)$ in \eqref{eqn:ATkR}. Denote $\I^*(\w)=\argmax_{\I\in\B_{m,5}}$ $ \left\{\frac{1}{5}\sum_{i\in \I} R_i(\w)\right\}$. We discuss the following two situations:
    \begin{enumerate}
        \item When $\w\in[0,0.5]$, we have $\{\mu_i\}_{i\in\I^*(\w)}=\{0.96,\cdots,1\}$, and $\L_{5}(\w)= \w^2-1.96\w+0.98$. In this case, $0.5$ attains the minimum objective $\L_5(0.5)=0.25$;
        \item When $\w\in(0.5,1]$, we have $\{\mu_i\}_{i\in\I^*(\w)}=\{0.5,0.86,\cdots,0.89\}$, and $\L_{5}(\w)= \w^2-1.6\w+0.8$. In this case, $0.8$ attains the minimum objective $\L_5(0.8)=0.16$.
    \end{enumerate}
In summary, $\w^*_A=\argmin_{\w\in\W} \L_{5}(\w)=0.8$.
\section{Supporting Algorithms} \label{appendix:Algorithms}
\subsection{Projection onto the Capped Simplex} \label{appendix:section:Projection}
\begin{algorithm}[H]
	\caption{Neg-entropy Bregman projection onto the capped simplex}
	\label{alg:projection}
	\textbf{Input:} size $k$, and non-negative vector $\p\in\mathbb{R}^m$
	\begin{algorithmic}[1]
		\IF{$\max_{i\in[m]}(p_i)\leq 1/k$ and $\sum_{i=1}^mp_i=1$}
		\RETURN $\p$
		\ENDIF
		\STATE Partially sort $\p$ to $\p^{\prime}$ s.t. $p^{\prime}_m\geq\cdots\geq p^{\prime}_{m-k+1}\geq p^{\prime}_i \ \forall i\in[m-k]$ and record mapping $\M$, i.e., $\M(\p)=\p^{\prime}$.
		\STATE Set $p^{\prime}_{m+1}=+\infty$
		\FOR{$i=m$ to $m-k+1$}
		\STATE $c=\left(1-\frac{m-i}{k}\right)/\left(\|\p^{\prime}\|_1-\sum_{j=i+1}^{m}p^{\prime}_j\right)$
		\IF{$p^{\prime}_{i}c< 1/k \leq p^{\prime}_{i+1}c$}
			\STATE $$p^{\prime}_{j}=\begin{cases} 1/k \  &j\geq i+1 \\ cp^{\prime}_j  \  & j\leq i  \end{cases}$$
			\RETURN $\M^{-1}(\p^{\prime})$
		\ENDIF
		\ENDFOR
		\RETURN $1/k \cdot \indicator{\p\neq\mathbf{0}_m}$
	\end{algorithmic}
\end{algorithm}

\subsection{Sampling Rule: DepRound} \label{appendix:section:DepRound}
The original DepRound algorithm \citep{DepRound} takes an input vector  $\p\in\R^m$ that satisfies $\mathbf{0} \leq\p\leq\mathbf{1} $ and $\p^{\top}\mathbf{1}=k$. Here, we modify it to require  $\p\in\SS_{m,k}$.
\begin{algorithm}[H]
	\caption{DepRound}
	\label{alg:DepRound}
	{\bf Input}: size $k$, and probability vector $\p\in \SS_{m,k}\subset\mathbb{R}^m$
	\begin{algorithmic}[1]
 		\STATE $\p = k \cdot \p$
		\WHILE{$\exists i\in [m]$ s.t. $p_i\in(0,1)$}
		\STATE Choose any $i,j\in[m]$ with $i\neq j$ and $p_i\in(0,1),p_j\in(0,1)$
		\STATE Set $\alpha=\min\{1-p_i,p_j\}$ and $\beta=\min\{p_i,1-p_j\}$
		\STATE Update 
		$$(p_i,p_j)=\begin{cases}(p_i+\alpha,p_j-\alpha)\text{ with probability }\frac{\beta}{\alpha+\beta}\\(p_i-\beta,p_j+\beta)\text{ with probability }\frac{\alpha}{\alpha+\beta}\end{cases}$$
		\ENDWHILE
		\RETURN $\I=\{i\in[n]\mid p_i=1\}$
	\end{algorithmic}
\end{algorithm}

\vskip 0.2in
%\bibliography{E:/MyPaper/ref}
\bibliography{ref}

\end{document}